%% file: main.tex
\documentclass{salesforce_ai_research}

\usepackage{amsmath}
\usepackage{wrapfig}
\usepackage{xcolor}
\usepackage{xcolor}
\usepackage{colortbl}
\usepackage{makecell}
\usepackage{amssymb}
\usepackage{graphicx}
\usepackage{booktabs}
\usepackage{lipsum}
\usepackage{multirow}
\usepackage{shadowtext}
\usepackage{xspace}
\usepackage{listings}
\usepackage{fontawesome5}

\lstdefinestyle{pythonstyle}{
    language=Python,
    basicstyle=\ttfamily\small,
    keywordstyle=\color{blue}\bfseries,
    commentstyle=\color{gray}\itshape,
    stringstyle=\color{red},
    numberstyle=\tiny\color{gray},
    breaklines=true,
    breakatwhitespace=false,
    tabsize=4,
    showstringspaces=false,
    showspaces=false,
    showtabs=false,
    frame=tb,
    framerule=0.5pt,
    rulecolor=\color{gray!30},
    backgroundcolor=\color{gray!3},
    xleftmargin=0pt,
    xrightmargin=0pt,
    aboveskip=8pt,
    belowskip=8pt,
    captionpos=b,
    numbers=none,
    escapeinside={(*@}{@*)},
    morekeywords={async, await, def, return, if, for, in, None, True, False}
}

\newcommand{\method}{\shadowtext{MCP-Universe}\xspace}

\newcommand{\grokemoji}{\includegraphics[height=1.3\fontcharht\font`\B]{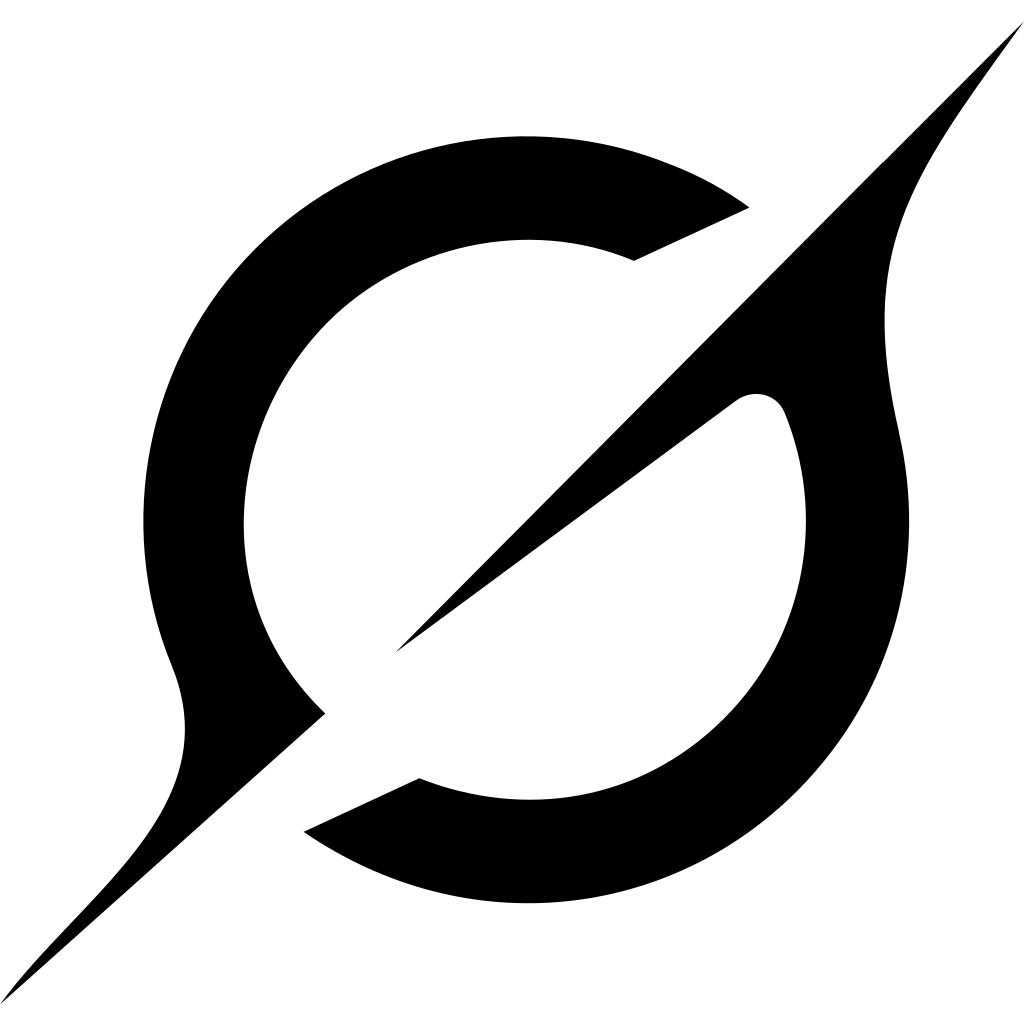}}
\newcommand{\Qwenemoji}{\includegraphics[height=1.3\fontcharht\font`\B]{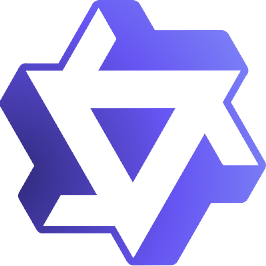}}
\newcommand{\Googleemoji}{\includegraphics[height=1.3\fontcharht\font`\B]{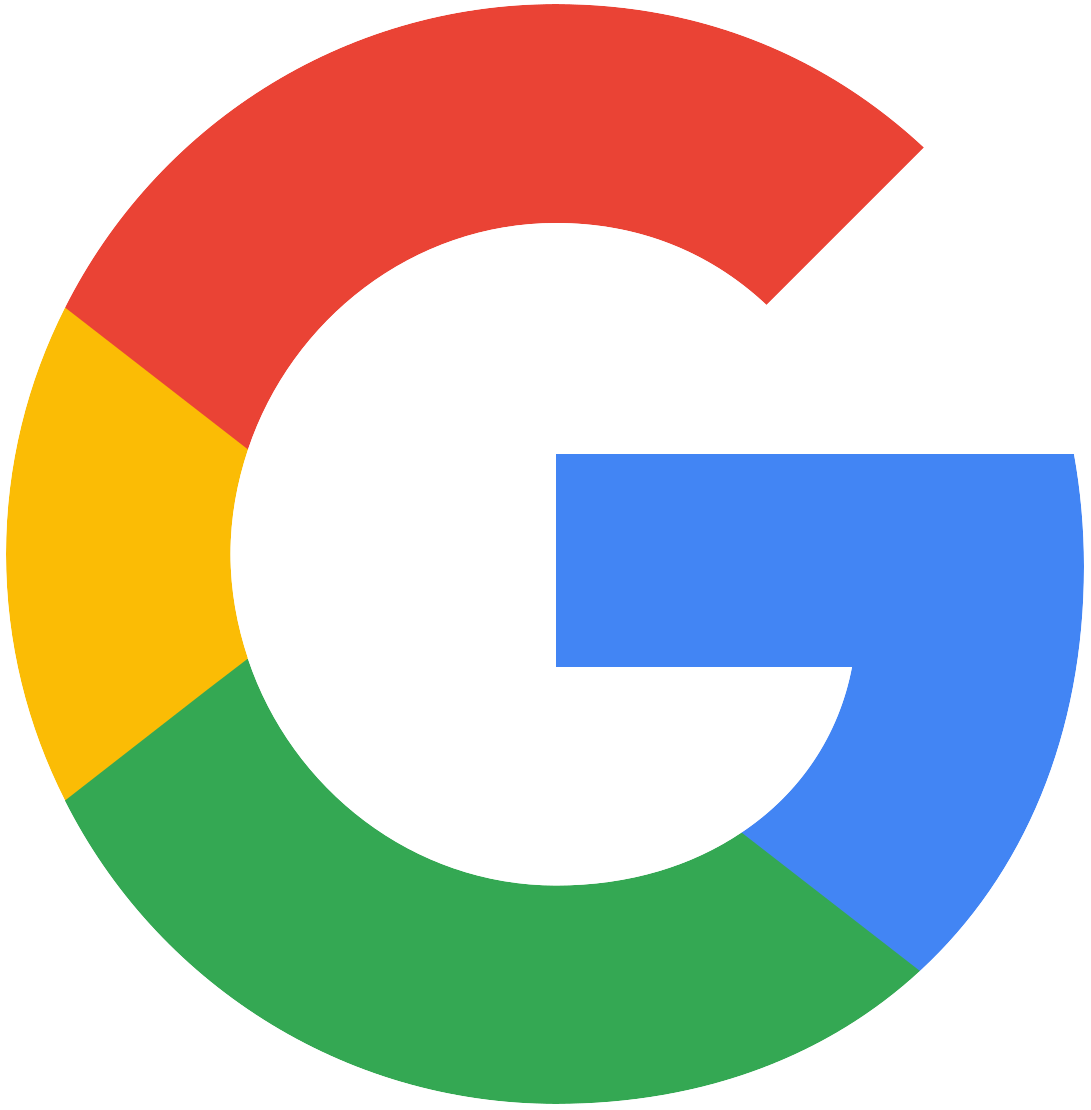}}
\newcommand{\Openaiemoji}{\includegraphics[height=1.3\fontcharht\font`\B]{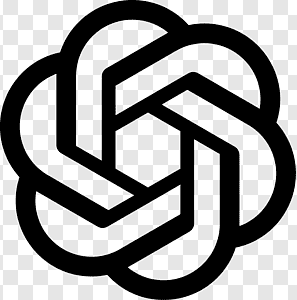}}
\newcommand{\claudemoji}{\includegraphics[height=1.0\fontcharht\font`\B]{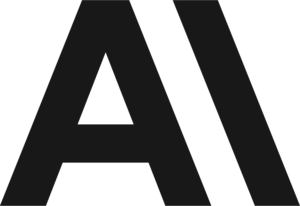}}
\newcommand{\kimimoji}{\includegraphics[height=1.5\fontcharht\font`\B]{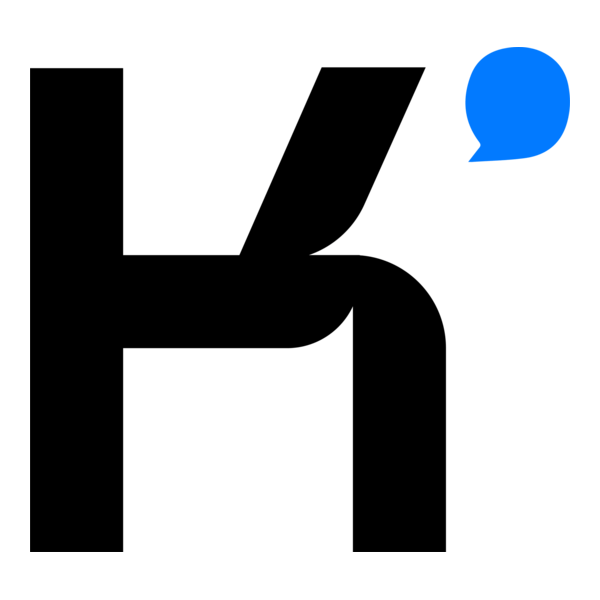}}
\newcommand{\deepseekemoji}{\includegraphics[height=1.0\fontcharht\font`\B]{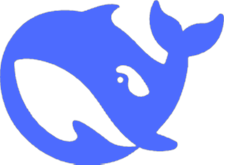}}
\newcommand{\zhipuemoji}{\includegraphics[height=1.2\fontcharht\font`\B]{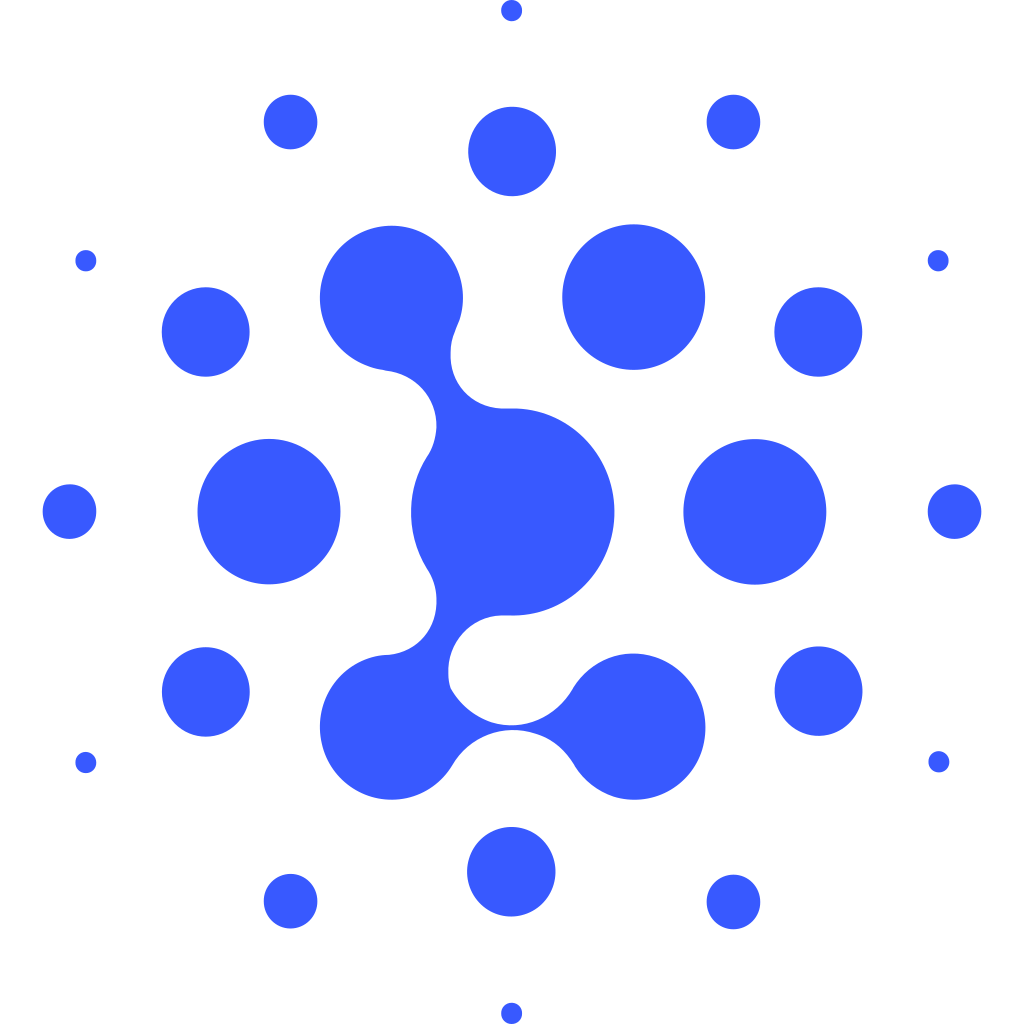}}

\title{\method: Benchmarking Large Language Models with\\Real-World Model Context Protocol Servers}

\author{
Ziyang Luo\thanks{Equal Contributions.}\footnotemark[1] \quad Zhiqi Shen\footnotemark[1] \quad Wenzhuo Yang\footnotemark[1] \quad Zirui Zhao \quad Prathyusha Jwalapuram \\
\quad \textbf{Amrita Saha} \quad \textbf{Doyen Sahoo} \quad \textbf{Silvio Savarese} \quad \textbf{Caiming Xiong} \quad \textbf{Junnan Li}\\
\textit{Salesforce AI Research} \\[0.5em]
\faGithub\, \url{https://github.com/SalesforceAIResearch/MCP-Universe} \\[0.2em]
\faGlobe\, \url{https://mcp-universe.github.io}
}

\begin{document}

\maketitle
\input{sections/000.abstract}
\input{sections/100.introduction}
\input{sections/200.related.work}
\input{sections/300.benchmark}
\input{sections/400.experiment}
\input{sections/500.conclusion}

\bibliography{references}
\bibliographystyle{ieeetr}

\newpage
\input{sections/600.appendix}

\end{document}

%% file: sections/000.abstract.tex
\begin{abstract}

The Model Context Protocol (MCP) has emerged as a transformative standard for connecting large language models (LLMs) to external data sources and tools, rapidly gaining adoption across major AI providers and development platforms. However, existing benchmarks are overly simplistic and fail to capture real application challenges such as long-horizon reasoning and large, unfamiliar tool spaces.
To address this critical gap, we introduce \textbf{MCP-Universe}, the first comprehensive benchmark specifically designed to evaluate LLMs in realistic and hard tasks through interaction with real-world MCP servers. Our benchmark encompasses 6 core domains spanning 11 different MCP servers: \textit{Location Navigation}, \textit{Repository Management}, \textit{Financial Analysis}, \textit{3D Design}, \textit{Browser Automation}, and \textit{Web Searching}. To ensure rigorous evaluation, we implement execution-based evaluators, including format evaluators for agent format compliance, static evaluators for time-invariant content matching, and dynamic evaluators that automatically retrieve real-time ground truth for temporally sensitive tasks.
Through extensive evaluation of leading LLMs, we find that even top-performing models such as GPT-5 (43.72\% success rate), Grok-4 (33.33\% success rate) and Claude-4.0-Sonnet (29.44\% success rate) exhibit significant performance limitations. In addition, our benchmark poses a significant long-context challenge for LLM agents, as the number of input tokens increases rapidly with the number of interaction steps. Moreover, it introduces an unknown-tools challenge, as LLM agents often lack familiarity with the precise usage of the MCP servers. Notably, enterprise-level agents like Cursor cannot achieve better performance than standard ReAct frameworks.
Beyond evaluation, we open-source our extensible evaluation framework with UI support, enabling researchers and practitioners to seamlessly integrate new agents and MCP servers while fostering innovation in the rapidly evolving MCP ecosystem.

\end{abstract}

%% file: sections/100.introduction.tex
\section{Introduction}


The Model Context Protocol (MCP), introduced by Anthropic~\cite{anthropic2024mcp}, represents a major paradigm shift in how AI systems interface with external data sources and tools. Dubbed the ``USB-C of AI"\cite{USBCAI}, MCP addresses the long-standing issue of fragmented, bespoke integrations that trap language models in isolated information silos\cite{SiloMCP}. Since its release, MCP has gained rapid traction: major AI providers, including OpenAI~\cite{OpenAIMCP} and Google Gemini~\cite{GoogleMCP}, have committed to adoption, while development platforms such as Cursor~\cite{CursorMCP} and Cline~\cite{ClineMCP} have begun integrating it to enhance their products. Despite the transformative potential of MCPs, current evaluations remain insufficient for assessing the true capabilities of LLMs operating within this new paradigm.  Existing benchmarks predominantly focus on isolated aspects of LLM performance, such as instruction following~\cite{IFEval}, math reasoning~\cite{GSM8k}, or function calling~\cite{bfcl}, without providing a comprehensive assessment of how models interact with real-world MCP servers across diverse scenarios. Recently, MCP-RADAR~\cite{MCP-RADAR} adapts existing benchmarks such as HumanEval~\cite{HE} and GSM8k~\cite{GSM8k} to the MCP context. However, these adaptations are largely derivative of existing datasets, which neither capture the full breadth of real-world applications nor adequately address issues such as data leakage. MCPWorld~\cite{MCPWorld}, another recently proposed benchmark, continues to rely heavily on graphical user interfaces (GUIs) and exhibits insufficient coverage of MCP-enabled workflows, thereby limiting its utility for evaluating LLMs in real MCP-driven environments.

To address these critical limitations, we introduce our \textbf{MCP-Universe}, a benchmark aiming at evaluating LLMs in realistic, challenging use cases with real-world MCP servers. As shown in Figure~\ref{fig:intro}, MCP-Universe captures realistic challenges: real-world tools usage, long-horizon multi-turn tool calls, long context windows, scattered evidence, and large tool spaces. Unlike existing works, MCP-Universe is grounded in real-world MCP servers that connect to actual data sources and environments. Our benchmark encompasses 6 core domains, with 11 MCP servers spanning diverse applications: \textit{Location Navigation}, \textit{Repository Management}, \textit{Financial Analysis}, \textit{3D Design}, \textit{Browser Automation}, and \textit{Web Searching}, comprising a total of 231 tasks. Each domain captures the operational complexities of real-world deployments, from handling authentic financial data and navigating complex geospatial information to managing version control workflows and executing real-time ticket price checks.

To ensure rigorous evaluation, we carefully design execution-based evaluators rather than relying on LLM-as-a-judge~\cite{LLMJudge} (e.g., MCPEval~\cite{MCPEval} and LiveMCPBench~\cite{LiveMCPBench}), recognizing that many tasks involve real-time data that static LLM knowledge cannot properly assess. Our evaluation includes format evaluators for agent format compliance, static evaluators for time-invariant content matching, and dynamic evaluators that automatically obtain real-time ground truth for temporally sensitive tasks. Furthermore, for the evolving nature of MCP servers, we provide an extensible, user-friendly framework that enables researchers and the broader community to seamlessly integrate new agents and MCP servers into the evaluation pipeline. Our benchmark is also equipped with a UI for intuitive, user-friendly access.

\input{figures/001.intro.mcp.figure}

We conducted extensive experiments using MCP-Universe across all 6 core domains and 11 different MCP servers. Through extensive evaluation of leading LLMs, we find that even top-performing models such as GPT-5 (43.72\% success rate), Grok-4 (33.33\% success rate), and Claude-4.0-Sonnet (29.44\% success rate) exhibit significant performance limitations, revealing a substantial gap between their impressive general capabilities and their effectiveness in real-world MCP environments. Our comprehensive analysis identifies several fundamental challenges that current LLM agents face in MCP interactions. First, we observe a \textit{long-context challenge}, as the number of tokens increases rapidly with the number of interaction steps, often leading to context overflow and degraded performance in multi-step tasks. Second, there exists an \textit{unknown-tools challenge}, where LLM agents often lack familiarity with the precise usage patterns, parameter specifications, and expected behaviors of diverse MCP servers. Additionally, our evaluation reveals significant \textit{cross-domain performance variations}, with models showing markedly different success rates across different application domains, suggesting domain-specific optimization needs. Notably, enterprise-level agents like Cursor cannot achieve better performance than standard ReAct frameworks, highlighting the challenges of our benchmark.

In summary, this work makes the following key contributions:
\begin{itemize}
    \item We introduce \textbf{MCP-Universe}, the first comprehensive benchmark for evaluating LLMs in MCP environments across six domains with real-world servers, where even SOTA LLMs struggle.
    
    \item We develop a rigorous \textbf{execution-based} evaluation framework with format, static, and dynamic assessment capabilities to enable comprehensive real-world performance measurement.
    
    \item We reveal fundamental limitations of current LLM agents, such as challenges with long contexts, handling unknown tools, and cross-domain discrepancies, thereby highlighting directions for future agent design.
\end{itemize}

%% file: figures/001.intro.mcp.figure.tex
\begin{figure}
    \centering
    \includegraphics[width=\linewidth]{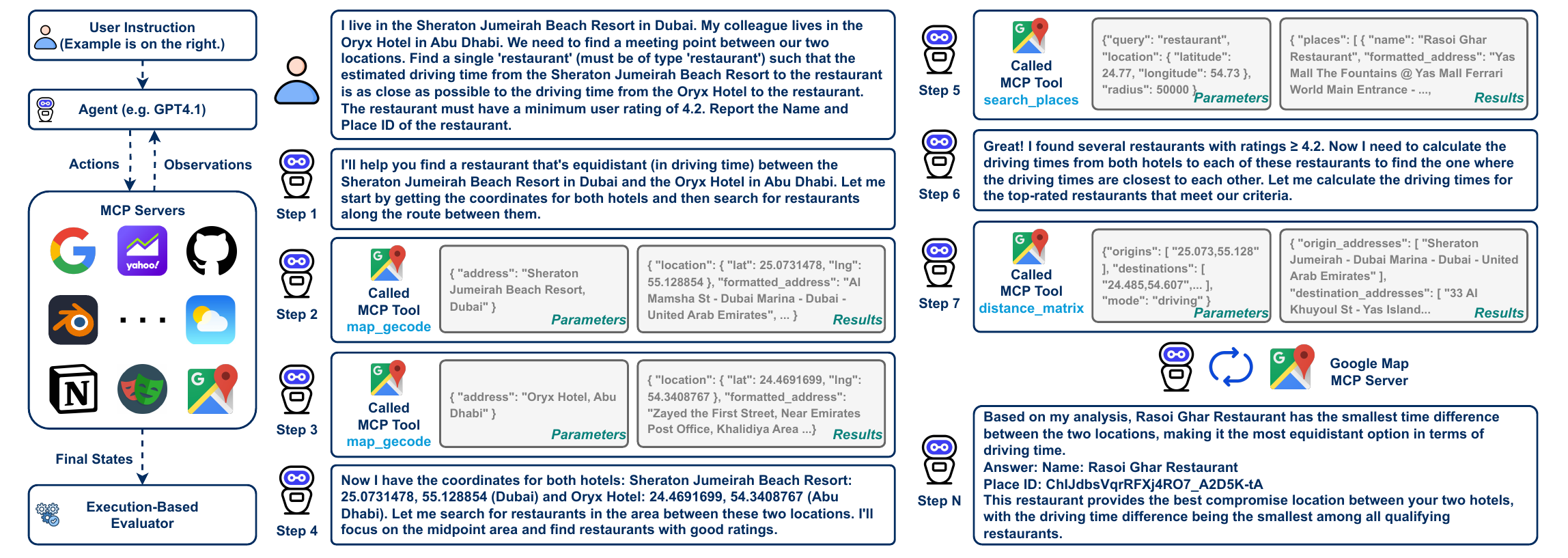}
    \vspace{-0.2cm}
    \caption{Example from MCP-Universe illustrating realistic challenges, including real-world tool usage, long-horizon multi-turn tool calls, long context windows, scattered evidence, and large tool spaces. Unlike prior work, MCP-Universe is grounded in real-world MCP servers connected to actual data sources and environments.}
    \label{fig:intro}
    \vspace{-0.2cm}
\end{figure}

%% file: sections/200.related.work.tex
\section{Related Work}

\textbf{Model Context Protocol.} MCP was introduced by Anthropic in late 2024 as an open standard designed to streamline AI system integration with external data sources and tools~\cite{anthropic2024mcp}. MCP addresses the notorious ``data silo problem" by providing a universal interface that connects AI systems to diverse data sources and tools through standardized JSON-RPC 2.0 messaging over STDIO and SSE transports~\cite{SiloMCP}. The protocol follows a three-layer architecture comprising MCP hosts (AI applications), clients (communication bridges), and servers (capability providers) that expose tools, prompts, and resources through standardized interfaces.

\textbf{LLMs as Agents.} LLMs have evolved from simple text-generating chatbots into sophisticated agents capable of autonomous planning, reasoning, and tool use~\cite{AgentSurvey}. This transformation has been driven by advances in instruction following~\cite{IFEval,AutoIF,MIA-Bench,IFBench}, multi-step reasoning~\cite{ReasoningSurvey,CoT,ToT,TesttimeSurvey}, and tool integration~\cite{toolsurvey,toolllm,LAMSurvey,xLAM}. Key agent design patterns have emerged, such as ReAct~\cite{react}, Reflection~\cite{Reflection}, and Plan-and-Solve~\cite{Plan-and-Solve}. Frameworks like AutoGen~\cite{AutoGen}, MetaGPT~\cite{MetaGPT}, Camel-AI~\cite{li2023camel}, and LangGraph~\cite{langgraph} have demonstrated practical implementations of autonomous agent systems. With the advancement of multimodal LLMs~\cite{GPT4o,Gemini}, a new class of GUI-based computer-use agents has emerged~\cite{CUASurvey,Aria-UI,sspro,GTA1}. Systems such as OpenAI's CUA~\cite{cua2025}, Anthropic's Computer-Use~\cite{anthropicCUA}, and ByteDance's UI-Taris~\cite{UI-TARS} mimic human interaction with graphical interfaces, giving rise to a new frontier in computer automation applications.

\input{tables/001.compare.benchmark}

\textbf{Evaluation of Agents.} The evaluation of LLM-based agents has become a key research area, with a variety of benchmarks developed to assess different aspects of agent capabilities. One major focus is web navigation, with benchmarks such as MiniWob++~\cite{MiniWeb++}, Mind2Web (1 \& 2)~\cite{Mind2Web,Mind2Web2}, WebLINX~\cite{WebLINX}, AssistantBench~\cite{AssistantBench}, WebArena~\cite{WebArena}, VisualWebArena, and VideoWebArena~\cite{VisualWebArena} providing comprehensive environments to test agents' ability to interact with realistic web applications. A second focus is GUI-based interaction, exemplified by OSWorld~\cite{OSWorld}, WindowsAgentArena~\cite{WindowsAgentArena}, and UI-Vision~\cite{UI-Vision}, which evaluate agents on their ability to operate computer interfaces similarly to human users. A third area is software engineering, with benchmarks like SWE-bench~\cite{SWE-bench} and DevBench~\cite{DevBench} designed to evaluate agents' capabilities in code generation and software development tasks. Another important dimension is function/tool calling, covered by APIBank~\cite{APIBank}, ToolBench~\cite{Toolbench}, GAIA~\cite{GAIA}, AppWorld~\cite{AppWorld}, $\tau$-Bench~\cite{tbench}, and BFCLv3~\cite{bfcl}, which assess an agent's proficiency in invoking external tools and APIs to complete complex tasks.

In contrast to existing benchmarks, several contemporary MCP-related benchmarks have recently emerged, as shown in Table~\ref{tab:benchmark_comparison}. MCPWorld~\cite{MCPWorld} evaluates agents in real-world GUIs and the MCP environment, but it relies heavily on the GUI and does not include time-varying tasks. MCP-RADAR~\cite{MCP-RADAR} transforms existing datasets, such as HumanEval and GSM8k, into MCP scenarios. Although it includes execution-based evaluation, its tasks are less related to real-world applications, and the ground truth does not change over time. The latter two benchmarks, MCPEval~\cite{MCPEval} and LiveMCPBench~\cite{LiveMCPBench}, both adopt the LLM-as-a-Judge evaluation, which is not suitable for tasks requiring real-time knowledge; moreover, LLM judges are also known for style bias~\cite{PreferenceLeakage}. In contrast to these works, MCP-Universe provides a comprehensive evaluation framework that integrates real-world integration, temporal dynamics, and execution-based evaluation. It evaluates agents on authentic MCP servers with time-sensitive scenarios and actual task completion metrics, addressing the limitations of existing benchmarks.

%% file: tables/001.compare.benchmark.tex
\begin{wraptable}{r}{0.48\textwidth}
    \centering
    \small
    \vspace{-0.5cm}
    \caption{Comparative Analysis of Contemporary MCP Benchmarks.}
    \newcommand{\greencheck}{{\color{green!60!black}\checkmark}}
    \newcommand{\redx}{{\color{red!70!black}\texttimes}}
    \begin{tabular}{l|ccc}
        \toprule
        \textbf{Benchmark} & \textbf{\makecell{Real-World\\Integration}} & \textbf{\makecell{Temporal\\Dynamics}} & \textbf{\makecell{Exec.\\Eval.}}\\
        \midrule
        MCPWorld & \greencheck & \redx & \greencheck\\
        MCP-RADAR & \redx & \redx & \greencheck\\
        MCPEval & \redx & \greencheck & \redx\\
        LiveMCPBench & \greencheck & \greencheck & \redx\\
        \midrule
        \textbf{MCP-Universe} & \greencheck & \greencheck & \greencheck\\
        \bottomrule
    \end{tabular}
    \label{tab:benchmark_comparison}
\end{wraptable}

%% file: sections/300.benchmark.tex
\input{figures/002.evaluation.framework}

\section{MCP-Universe}

\subsection{Overview}

\textbf{MCP-Universe} is a comprehensive evaluation framework designed to assess the capabilities of LLMs when interacting with real-world MCP servers for challenging and practical use cases. As shown in Figure~\ref{fig:evaluation_framework}, our benchmark encompasses three core components: 
(1) an extensible, easy-to-use evaluation framework; 
(2) a collection of carefully designed task instructions grounded in real-world MCP server scenarios; 
(3) a suite of execution-based evaluators for measuring task completion.

To formalize the setting, we model the benchmark as follows. Let $S = \{s_1, s_2, \ldots, s_k\}$ denote the collection of MCP servers, where each server $s_i$ exposes a set of tools $T_i = \{t_{i,1}, t_{i,2}, \ldots, t_{i,|T_i|}\}$ through the MCP protocol. A task $\tau$ is defined as a tuple $(G, C, T_{\mathrm{available}})$, where:
\begin{itemize}
    \item $G$ is the goal specification describing the desired outcome;
    \item $C$ contains the initial context and any relevant background information;
    \item $T_{\mathrm{available}} = \bigcup_{i \in I} T_i$ is the set of tools accessible for the task, with $I \subseteq \{1, \ldots, k\}$ indicating which servers are available.
\end{itemize}
The benchmark challenges an agent to identify, sequence, and invoke appropriate tools from $T_{\mathrm{available}}$ to achieve $G$ given $C$, requiring reasoning over partial information, adapting to diverse tool interfaces, and handling ambiguities or failures in tool responses.

For evaluation, let $M = \{m_1, m_2, \ldots, m_n\}$ be the set of language models and $A = \{a_1, a_2, \ldots, a_p\}$ be the set of agent architectures (e.g., ReAct) that can be paired with them. For a given $(m, a) \in M \times A$ and task $\tau$, the interaction produces a conversation trace $R = (r_1, r_2, \ldots, r_T)$, where each $r_t$ contains the agent’s output and any tool invocations. The evaluation function
\[
E: M \times A \times \mathcal{T} \rightarrow \{0, 1\}
\]
assigns $1$ if the task is successfully completed according to predefined success criteria, and $0$ otherwise. Success is determined through a combination of automated checks (e.g., verifying structured outputs or end states). Aggregating $E(m, a, \tau)$ over all tasks yields a quantitative measure of an agent's proficiency in MCP-driven tool use.

\subsection{Evaluation Framework}

As illustrated in Figure~\ref{fig:evaluation_framework}, our evaluation framework seamlessly coordinates multiple components to deliver objective and reproducible assessment results. Given the task specifications, the framework begins with an automatic configuration that dynamically orchestrates the evaluation pipeline. The framework automatically builds the LLM-agent combination, selects the required MCP servers, and configures the corresponding evaluators. The configuration also handles resource allocation, API endpoint management, and evaluation criteria specification.

In detail, an LLM Manager is introduced in the framework, which supports multiple SOTA LLMs including GPT-5 and Claude-4.0-Sonnet. This component handles LLM configuration, API management, and standardized prompt formatting to ensure consistent evaluation conditions across different models. In addition, the Agent Builder component constructs specialized agents. It supports multiple agents including ReAct and ReAct with Exploration. The builder configures agents with appropriate reasoning strategies for MCP server communication. This modular design allows for systematic comparison of different agents within the same evaluation framework.

Moreover, our framework seamlessly integrates with diverse MCP servers representing real-world tools and services. Each server is configured with its authentic API endpoints, authentication mechanisms, and tool specifications, ensuring that evaluation tasks mirror real-world environments rather than simplified simulations. The framework handles dynamic server configuration based on task specification, allowing for both single-server and multi-server evaluation scenarios.

The evaluation process employs an execution-based approach that validates task completion through automated assessment rather than subjective LLMs judgment or costly human annotation. The evaluator implements domain-specific validation strategies including stop type validation for Google Maps and branch checking for GitHub. The complete evaluation workflow proceeds through agent-server interactions mediated via the MCP protocol, followed by automated assessment that produces binary pass/fail determinations based on objective validation criteria. The framework captures detailed interaction logs to provide comprehensive insights into model performance across different scenarios and server types.

\input{tables/002.data.statistic}

\subsection{Real-World MCP Servers}

A foundational design principle of MCP-Universe is its reliance on real-world MCP servers, as opposed to simulated environments. This approach ensures that evaluation is grounded in the authentic complexities of practical applications. As shown in Table~\ref{tab:key_statistics}, the benchmark includes 11 distinct MCP servers with total 133 tools, each reflecting a unique application domain. These servers are organized into 6 core domains:
\begin{enumerate}
    \item \textbf{Location Navigation}: This domain focuses on geographic reasoning and spatial task execution in real-world environments. We employ the official Google Maps MCP server, which provides a rich suite of geospatial tools such as location search, route planning, and distance computation. Models must navigate the full complexity of real-world location data to complete navigation tasks effectively.
    \item \textbf{Repository Management}: This domain focuses on repository development and codebase operations. We employ the GitHub MCP server, which exposes authentic version control tools such as repository search, issue tracking, and code editing. This setup reflects the operational demands of real-world development.
    \item \textbf{Financial Analysis}: This domain focuses on quantitative reasoning and decision-making in dynamic financial markets. We utilize the Yahoo Finance MCP server, which provides tools such as stock price monitoring, shareholder information checking, and options tracking, all based on live financial data. The server poses substantial challenges in parsing and reasoning over real-time market information.
    \item \textbf{3D Designing}: This domain focuses on computer-aided design. We employ the Blender MCP server, which enables interaction with advanced tools for 3D modeling, such as object creation, asset manipulation, and material setup. This domain captures the technical depth of professional design environments.
    \item \textbf{Browser Automation}: This domain focuses on automated interaction with web applications and interfaces. We employ the Playwright MCP server, which provides full-featured web automation capabilities such as browser navigation, button clicking, and page snapshotting. It represents real-world browser control scenarios encountered in modern automation pipelines.
    \item \textbf{Web Searching}: This domain focuses on open-domain information seeking. The Google Search MCP server is integrated to support web-based information seeking, and the Fetch MCP server is adopted to obtain the content for a given URL. This domain captures the open-ended nature of real-world web searching tasks.
\end{enumerate}

The selection of these domains and specific servers reflects our commitment to domain diversity and real-world relevance. In addition to the above MCP servers, we also incorporate additional MCP servers that provide necessary support and contribute to increasing task complexity, such as the Notion MCP server, Weather MCP server, Date MCP server, and Calculator MCP server. More details on these MCP servers can be found in Appendix~\ref{app:mcp}.

\subsection{Tasks and Evaluators}

Since MCP is a new concept and there is a lack of high-quality usage examples, we manually designed challenging MCP tasks to reflect real use cases. If a task can be easily completed by LLMs without using MCP servers, or can be consistently solved with MCP servers within five retries, we consider it a simple task and brainstorm a new one. As shown in Figure~\ref{fig:task_distribution}, for each domain, we carefully create 4-5 types of tasks to cover the most common usage scenarios. For the location navigation domain, we focus on 4 sub-task types, including route planning, optimal stops, location searching, and place finding. For the repository management domain, we focus on 4 sub-task types, including project setup, issue tracking, automation setup, and code integration. For the financial analysis domain, we focus on 5 sub-task types, including portfolio analysis, financial statements, trading strategies, institutional holdings, and dividend analysis. For the 3D designing domain, we focus on 5 sub-task types, including object creation, material setup, lighting configuration, render settings, and scene hierarchy. For the browser automation domain, we focus on 5 sub-task types, including travel booking, sports analytics, academic research, platform exploration, and map navigation. For the web searching domain, we focus on 5 sub-task types, including person identification, entity discovery, metric matching, complex reasoning, and factual lookup. After the task creation, each task will be cross-checked by the other authors for feasibility, ambiguity, and correctness.

To evaluate the completion of tasks, we have chosen to carefully design \textbf{execution-based evaluators} for each task. For simplicity, many recent works choose to follow the LLM-as-a-judge paradigm. However, we argue that this paradigm is not well-suited for our MCP-Universe scenario, since some tasks are designed to use real-time data, while the knowledge of the LLM judge is static. In addition, the LLM judge is also known to contain style bias and hallucinations. Thus, we choose to follow the execution-based paradigm to evaluate the completion of tasks. Although this relies on much heavier human labor, we believe this is the only way to achieve a fair and comprehensive evaluation. For all tasks, we can divide the evaluators into three types: (1) \textbf{Format Evaluators}, (2) \textbf{Static Evaluators}, and (3) \textbf{Dynamic Evaluators}. The first type evaluates whether agents strictly follow format requirements. The second type assesses correctness for tasks whose answers do not change over time, such as the number of cities in route planning tasks, the exact number of goals scored by a football player in browser automation tasks, or historical stock prices in financial analysis tasks. For these, we manually collect the correct answers and write evaluators to check whether the model's outputs meet the requirements. For the third type, the correct answer of the task needs to be updated with real-time data, such as the price of a flight on a future date for the travel booking tasks, the weather of a place in the place-finding tasks, and the number of GitHub issues in the issue tracking tasks. Here, we design automatic evaluators to obtain real-time correct answers and verify task completion, which can provide stable evaluation results across different timestamps. After the evaluator creation, each evaluators will be cross-checked by the other authors for feasibility, ambiguity, and correctness. In Appendix~\ref{app:task_evaluator}, we include the examples of tasks and evaluators in our benchmark.

%% file: figures/002.evaluation.framework.tex
\begin{figure}
    \centering
    \includegraphics[width=\linewidth]{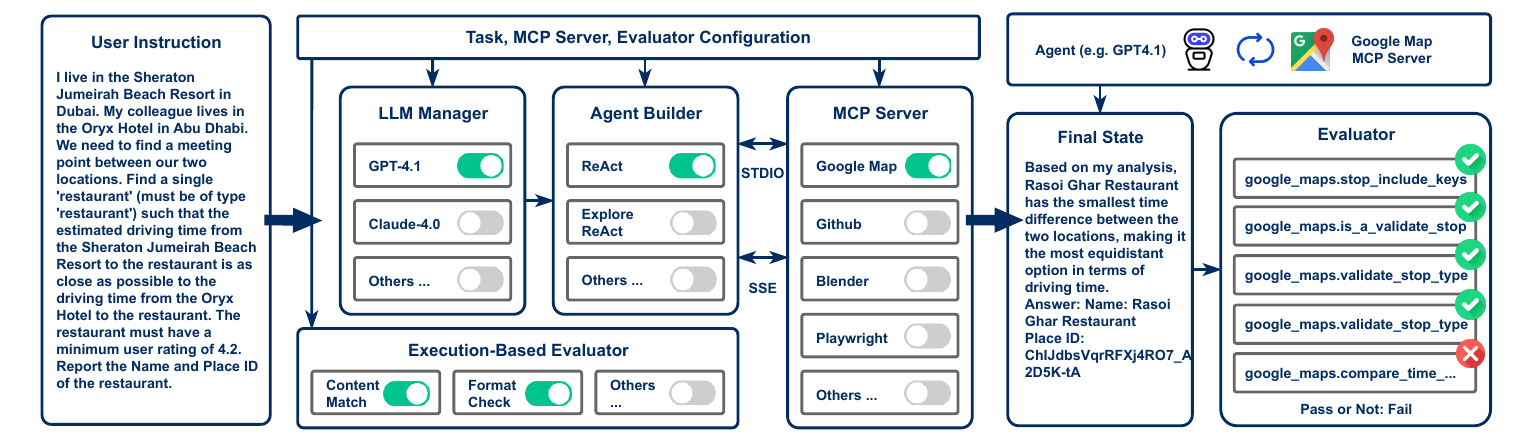}
    \caption{Overview of the MCP-Universe evaluation framework. The framework dynamically configures LLM agents, MCP servers, and execution-based evaluators according to task specifications. Each evaluation involves the agent-server interactions mediated via the MCP protocol, followed by an objective assessment conducted by automated execution-based evaluators to determine the success of task completion.}
    \label{fig:evaluation_framework}
\end{figure}

%% file: tables/002.data.statistic.tex
\begin{figure}
        \centering
        \begin{minipage}{0.50\textwidth}
            \centering
            \includegraphics[width=0.9\textwidth]{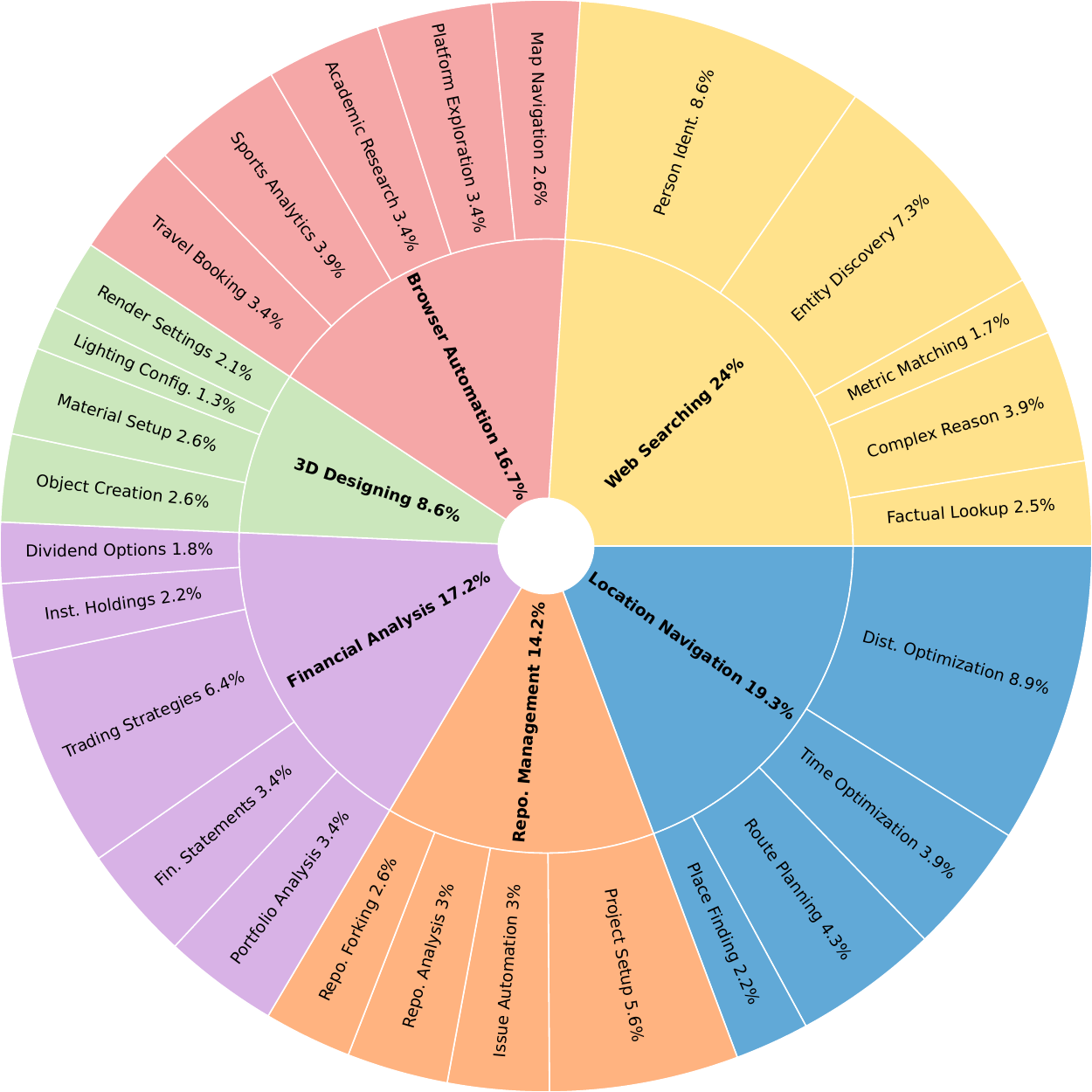}
            \caption{Distribution of tasks in MCP-Universe across different application domains.}
            \label{fig:task_distribution}
        \end{minipage}
        \hspace{5mm}
        \begin{minipage}{0.4\textwidth}
            \centering
            \captionof{table}{Key statistics in MCP-Universe.}
            \small
            \begin{tabular}{lc}
                \toprule
                \textbf{Statistic} & \textbf{Number} \\
                \midrule
                Total tasks & 231 (100\%) \\
                - Location Navigation & 45 (19.5\%)\\
                - Web Searching & 55 (23.8\%)\\
                - Browser Automation & 39 (16.9\%)\\
                - 3D Designing & 19 (8.2\%)\\
                - Financial Analysis & 40 (17.3\%)\\
                - Repo. Management & 33 (14.3\%)\\
                \midrule
                Total MCP Servers & 11 \\
                Total Tools in Servers & 133\\
                \midrule
                Total Unique Evaluators & 84 (100\%)\\
                - Format Evaluators & 4 (4.8\%)\\
                - Static Evaluators & 32 (38.1\%)\\
                - Dynamic Evaluators & 48 (57.1\%)\\
                \bottomrule
            \end{tabular}
            \label{tab:key_statistics}
        \end{minipage}
    \end{figure}

%% file: sections/400.experiment.tex
\begin{table*}[t]
    \centering
    \caption{Comparison on our MCP-Universe benchmark. For our main experiments, all LLMs follow the ReAct agent pipeline, except GPT-OSS, which has poor instruction-following abilities and therefore cannot follow the ReAct prompt; for this model, we use the OpenAI Agent SDK instead. We report the success rate (SR, \%) for each domain and all tasks. Additionally, we calculate the average percentage of evaluators passed for each task, which we refer to as the average evaluator score (AE). Moreover, we also report the average number of steps (AS) for each successful task. Since GPT-OSS does not follow ReAct, it does not have AS scores.}
    \resizebox{\textwidth}{!}{
    \begin{tabular}{l|cccccc|ccc}
        \toprule
        \multirow{2}{*}{\textbf{Model}} & \multirow{2}{*}{\textbf{\makecell{Location\\Navigation}}} & \multirow{2}{*}{\textbf{\makecell{Repository\\Management}}} & \multirow{2}{*}{\textbf{\makecell{Financial\\Analysis}}} & \multirow{2}{*}{\textbf{\makecell{3D\\Designing}}} & \multirow{2}{*}{\textbf{\makecell{Browser\\Automation}}} & \multirow{2}{*}{\textbf{\makecell{Web\\Searching}}} & \multicolumn{3}{c}{\textbf{Overall}} \\
         & & & & & & & \textbf{SR} & \textbf{AE} & \textbf{AS} \\
        \midrule
        \rowcolor{pink!50}
        \multicolumn{10}{c}{\textit{Proprietary Models}}\\
        \Openaiemoji{}~GPT-5 & \textbf{33.33} & \textbf{30.30} & \textbf{67.50} & \textbf{52.63} & 35.90 & \textbf{45.45} & \textbf{43.72} & \textbf{60.23} & \textbf{8.22}\\
        \grokemoji{}~Grok-4 & 28.89 & 12.12 & 40.00 & 26.32 & \textbf{41.03} & 41.82 & 33.33 & 49.01 & 7.75\\
        \claudemoji{}~Claude-4.0-Sonnet & 22.22 & 12.12 & 55.00 & 26.32 & 38.46 & 21.82 & 29.44 & 50.61 & 7.46\\
        \Openaiemoji{}~o3 & 26.67 & 6.06 & 40.00 & 26.32 & 25.64 & 29.09 & 26.41 & 38.95 & 4.82\\
        \Openaiemoji{}~o4-mini & 26.67 & 18.18 & 40.00 & 36.84 & 23.08 & 18.18 & 25.97 & 40.38 & 7.90 \\
        \claudemoji{}~Claude-3.7-Sonnet & 13.33 & 18.18 & 40.00 & 36.84 & 23.08 & 21.82 & 24.24 & 40.36 & 7.16 \\
        \Googleemoji{}~Gemini-2.5-Pro & 13.33 & 12.12 & 50.00 & 21.05 & 25.64 & 12.73 & 22.08 & 36.93 & 6.98\\
        \Googleemoji{}~Gemini-2.5-Flash & 15.56 & 12.12 & 37.50 & 21.05 & 30.77 & 14.55 & 21.65 & 33.99 & 8.26 \\
        \Openaiemoji{}~GPT-4.1 & 8.89 & 6.06 & 40.00 & 26.32 & 23.08 & 10.91 & 18.18 & 41.32 & 5.24 \\
        \Openaiemoji{}~GPT-4o & 8.89 & 9.09 & 35.00 & 26.32 & 12.82 & 9.09 & 15.58 & 37.03 & 6.03 \\
        \midrule
        \rowcolor{green!20}
        \multicolumn{10}{c}{\textit{ Open-Source Models}}\\
        \zhipuemoji{}~GLM-4.5 & 17.78 & 9.09 & 50.00 & 26.32 & 15.38 & 27.27 & 24.68 & 41.16 & 7.33\\
        \kimimoji{}~Kimi-K2 & 11.11 & 9.09 & 47.50 & 15.79 & 15.38 & 14.55 & 19.05 & 35.10 & 6.07 \\
        \Qwenemoji{}~Qwen3-Coder & 8.89 & 3.03 & 50.00 & 26.32 & 25.64 & 10.91 & 19.91 & 37.78 & 7.78\\
        \Qwenemoji{}~Qwen3-235B & 11.11 & 9.09 & 50.00 & 15.79 & 15.38 & 9.09 & 18.18 & 38.53 & 5.74 \\
        \deepseekemoji{}~DeepSeek-V3 & 11.11 & 6.06 & 30.00 & 26.32 & 12.82 & 7.27 & 14.29 & 35.82 & 5.06\\
        \Openaiemoji{}~GPT-OSS-120B & 6.67 & 6.06 & 35.00 & 10.53 & 5.13 & 5.45 & 11.26 & 26.34 & - \\
        \bottomrule
    \end{tabular}
    }
    \label{tab:main_results}
\end{table*}

\section{Experiment}

\subsection{Setup}

In our experiments, we evaluate the performance of SOTA  proprietary and open-source LLMs on our MCP-Universe benchmark. The models include xAI's Grok-4~\cite{grok4}, Anthropic's Claude-4.0-Sonnet~\cite{claude4} and Claude-3.7-Sonnet~\cite{claude3.7}, OpenAI's GPT-5~\cite{gpt5}, o3, o4-mini~\cite{o3-o4mini},GPT-5~\cite{gpt5},  GPT-4.1~\cite{GPT4-1}, GPT-4o~\cite{GPT4o}, GPT-OSS~\cite{GPT-OSS}, Google's Gemini-2.5-Pro and Gemini-2.5-Flash~\cite{Gemini2-5}, Zai's GLM-4.5~\cite{GLM4_5}, Moonshot's Kimi-K2~\cite{KimiK2}, Qwen's Qwen3-Coder, and Qwen3-235B-A22B-Instruct-2507~\cite{Qwen3}, and DeepSeek's DeepSeek-V3-0324~\cite{DeepSeek-V3}. All LLMs are top-ranked on the well-known lmsys Chatbot Arena leaderboard~\cite{ChatboArena}. All open-source LLMs have at least 120B parameters. For the agent, we adopt the most popular framework, ReAct~\cite{react}. The LLMs first generate a thought based on the observation, and then generate the next action based on that thought. More setup details can be found in the Appendix~\ref{app:setup}.

\subsection{Frontier Models Performance}

\input{tables/004.different.evaluators}

As shown in Table~\ref{tab:main_results}, we compare the performance of SOTA proprietary and open source LLMs on our MCP-Universe benchmark. The results indicate that OpenAI's GPT-5 achieves the highest overall success rate at 43.72\%, significantly outperforming other models. Grok-4 ranks second with 33.33\% success rate, followed by Claude-4.0-Sonnet at 29.44\%. GPT-5 demonstrates particularly strong performance in the Financial Analysis domain (67.50\%) and 3D Designing domain (52.63\%), while also achieving the highest success rate in Web Searching (45.45\%). Notably, Grok-4 excels in Browser Automation (41.03\%) domain, which requires strong reasoning and the ability to operate in complex internet environments. In the Location Navigation domain, all LLMs perform poorly, with success rates below 35\%; notably, popular models like GPT-4.1 and GPT-4o score under 10\%. Similarly, in the Repository Management domain, only GPT-5 surpasses a 30\% success rate. In the 3D Design domain, only GPT-5 exceeds a 50\% success rate, while other top models remain below 40\%. When comparing proprietary and open-source models, we find that GLM-4.5 is the best open-source LLM with 24.68\% overall success rate, achieving a higher success rate than some proprietary models like o4-mini and Claude-3.7-Sonnet. However, the gap between SOTA proprietary LLMs and their open-source counterparts remains substantial.

Beyond success rates, we also evaluate each model based on (i) the percentage of evaluators they can satisfy, measured via average evaluator (AE) scores, and (ii) the average number of steps (AS) taken to complete successful tasks. While higher AE scores often correlate with higher success rates, the relationship is not always direct. GPT-5 achieves both the highest success rate (43.72\%) and the highest AE score (60.23\%), demonstrating strong consistency. However, some models show interesting discrepancies: Claude-4.0-Sonnet passes 50.61\% of evaluators, slightly higher than Grok-4's 49.01\%, yet Grok-4 achieves a higher overall success rate (33.33\% vs. 29.44\%). Regarding task efficiency, o3 requires only 4.82 average steps despite ranking fourth in success rate (26.41\%), making it the most efficient model. In contrast, the top-performing models GPT-5 (8.22 steps) and Grok-4 (7.75 steps) require significantly more steps. Most open-source models complete successful tasks in 5-7 steps, with DeepSeek-V3 being the most efficient at 5.06 steps. These findings highlight that current frontier LLMs still fall short in reliably executing tasks across diverse real world MCP tasks. Our MCP-Universe benchmark therefore provides a challenging and necessary testbed for evaluating LLM performance in areas underserved by existing benchmarks.

Furthermore, our benchmark incorporates three evaluator types: format evaluators, static evaluators, and dynamic evaluators. Table~\ref{tab:different_evaluators} presents a breakdown of model performance across these types. Non-reasoning LLMs, such as Claude-4.0-Sonnet,\footnote{We do not use the thinking mode.} GPT-4.1, GPT-4o, Qwen3-235B, and DeepSeek-V3, achieve over 90\% success with format evaluators, with Claude-4.0-Sonnet leading at 98.29\%. In contrast, reasoning models like o3 (73.50\%), Gemini-2.5-Pro (64.10\%), and Gemini-2.5-Flash (51.28\%) perform significantly worse on format evaluators, suggesting that such models are less adept at adhering to strict formatting instructions. We highlight the naive error in the Appendix~\ref{app:errors}. On content-sensitive static evaluators, GPT-5 and Claude-4.0-Sonnet both achieve the highest performance at 61.92\%, while most other models achieve around 40-50\% success. For dynamic evaluators, GPT-5 leads with 65.96\%, followed by Claude-4.0-Sonnet at 54.74\%. The substantial performance gap between format evaluators (where many models exceed 80\%) and content evaluators (where most models achieve 40-60\%) indicates that the primary source of failure lies in content generation rather than format compliance. This demonstrates that our benchmark evaluates LLMs from multiple angles, including format compliance and content correctness under both static and dynamic conditions, making it a comprehensive testbed for model assessment.

\subsection{Long Context Challenges}

\input{figures/003.summarization.performance}



In our MCP Universe benchmark, long context handling poses a significant challenge for LLM agents, particularly in the Location Navigation, Browser Automation, and Financial Analysis domains. These domains frequently require agents to process and reason over lengthy sequences of observations or historical actions, which often exceed the context window limits of many models. In the Location Navigation domain, the Google Maps MCP servers can return extensive location data within the context, including detailed information about multiple places. In the Browser Automation domain, the Playwright MCP servers may return the full HTML content of a webpage, which can be very large. Similarly, in the Financial Analysis domain, the Yahoo Finance MCP servers provide daily stock information over a specified date range, and this can result in a large volume of contextual data if the time span is long.

As shown in the left of Figure~\ref{fig:long_context_performance}, we observe that the number of tokens increases rapidly as the number of interaction steps grows. This demonstrates that long context is one of the key challenges presented by our benchmark.\footnote{The context length experiment is based on the Claude-4.0-Sonnet.} To explore potential solutions, we conducted a preliminary experiment (the right of Figure~\ref{fig:long_context_performance}) by introducing a summarization agent at each step, designed to compress the raw outputs of the MCP servers. This summarizer attempts to reduce context length while preserving essential information. The details of this summarization agent are shown in the Appendix~\ref{app:summarize}. However, this approach yields mixed results. While it leads to improved success rates for GPT-4.1 and Claude-4.0-Sonnet in the Location Navigation domain, it either has no effect or negatively impacts performance in the Browser Automation and Financial Analysis domains. These findings indicate that MCP-Universe introduces unique and realistic long context challenges in agent-based tasks. Simple summarization methods are insufficient to address these issues. As such, our benchmark serves as a valuable testbed for evaluating and developing long context handling in LLM agent systems.

\subsection{Unknown Tools Challenges}

\input{figures/004.exploration.performance}
In addition to the long context challenges, our error analysis reveals that LLMs often struggle to correctly use tools provided by the MCP servers, indicating a lack of familiarity with their interfaces and constraints. For example, Figure~\ref{fig:explore} (left) illustrates a common failure in the Yahoo Finance MCP server: retrieving a stock price requires specifying a start and end date that differ, yet LLMs frequently set them to be identical, leading to execution errors. To address this issue, we introduce an additional step called the exploration phase. During this phase, the LLM is allowed to freely interact with the tools provided by the MCP servers. This gives the model an opportunity to learn how the tools work and to build knowledge of their capabilities. In the subsequent exploitation phase, the LLM uses this acquired tool knowledge, combined with a ReAct-style framework, to solve the actual tasks. The details are included in the Appendix~\ref{app:explore}.

As shown in Figure~\ref{fig:explore} (right), incorporating an exploration phase leads to performance improvements in certain domains. For instance, GPT-4.1 achieves a 30.77\% success rate in the Browser Automation domain, an improvement of 7.69 percentage points. Similarly, Claude-4.0-Sonnet reaches a 62.50\% success rate in the Financial Analysis domain, a 7.50 percentage point increase. However, this approach is not universally beneficial. Claude-4.0-Sonnet's performance in the Browser Automation domain declines, and GPT-4.1 shows no improvement in the Financial Analysis domain. These mixed results suggest that while the exploration phase can help some LLMs perform better in specific domains (e.g., information seeking and reasoning), it is not a one-size-fits-all solution (e.g., planning for state changes). Achieving strong performance on the MCP-Universe benchmark requires more robust and adaptive strategies, emphasizing the benchmark's difficulty and its value as a testing ground for advanced LLM agents.

\subsection{More MCP Servers Connected}

\input{figures/005.more.servers.performance}
In Table~\ref{tab:main_results}, we only connect the MCP servers that are directly relevant to each task. In this section, we extend the setup by connecting additional, unrelated MCP servers to the LLMs to assess their performance under increased tool complexity. For all tasks, we connect a total of 7 MCP servers, comprising 94 tools. This configuration introduces additional noise and results in a noticeable decline in performance, as illustrated in Figure~\ref{fig:more_servers}. For example, Claude-4.0-Sonnet's success rate in the Location Navigation domain drops from 22.22\% to 11.11\%. GPT-4.1's success rate in the Browser Automation domain decreases from 23.08\% to 15.38\%, and in the Financial Analysis domain, it drops from 40.00\% to 35.00\%. These results demonstrate that our benchmark can also serve as a valuable testbed for evaluating the robustness of LLMs when confronted with a larger number of unrelated tools.



\subsection{Enterprise-Level Agent Framework Comparison}
\input{tables/005.agent.compare}

In the previous experiments, we focused on evaluating different LLMs using the ReAct framework. To assess the impact of agent architecture on performance, we conduct a comparison of different agent frameworks as shown in Table~\ref{tab:agent_comparison}. We evaluate four distinct configurations: two frameworks using Claude-4.0-Sonnet as the backbone (ReAct and Cursor Agent), and two using OpenAI's o3 model (ReAct and OpenAI Agent SDK\footnote{\url{https://github.com/openai/openai-agents-python}}).

For the Claude-4.0-Sonnet backbone comparisons, we observe that the ReAct framework achieves a higher overall success rate (29.44\%) compared to the enterprise-level Cursor Agent (26.41\%). While Cursor Agent demonstrates superior performance in Browser Automation (43.59\% vs. 38.46\%), it significantly underperforms in Web Searching (7.27\% vs. 21.82\%), resulting in a 3.03 percentage point overall deficit. This disparity is particularly notable in Web Searching, where Cursor Agent's reliance on internal tools rather than the benchmark's MCP servers may contribute to the performance gap. For the OpenAI o3 backbone comparisons, the OpenAI Agent SDK substantially outperforms the ReAct framework (31.60\% vs. 26.41\%). The Agent SDK demonstrates consistent advantages across most domains, particularly excelling in Financial Analysis (60.00\% vs. 40.00\%) and 3D Designing (36.84\% vs. 26.32\%). The o3 + OpenAI Agent SDK configuration achieves the highest overall performance among all tested agent-backbone combinations, suggesting that specialized agent architectures can effectively leverage model capabilities.

These results highlight that agent framework design significantly impacts performance on our benchmark. While enterprise-level agents like Cursor may excel in specific domains, they do not universally outperform simpler frameworks like ReAct. Furthermore, the substantial performance difference between o3 + ReAct and o3 + OpenAI Agent SDK demonstrates that optimal agent-model pairing is crucial for maximizing performance on complex tasks.

%% file: tables/004.different.evaluators.tex
\begin{wraptable}{r}{0.5\textwidth}
    \centering
    \small
    \vspace{-0.5cm}
    \caption{Success rate across different types of evaluators on our MCP-Universe benchmark.}
    \vspace{-0.3cm}
    \begin{tabular}{l|ccc}
        \toprule
        \textbf{Model} & \textbf{Format} & \textbf{Static} & \textbf{Dynamic}\\
        \midrule
        GPT-5 & 88.89 & \textbf{61.92} & \textbf{65.96}\\
        Grok-4 & 88.03 & 49.04 & 52.98\\
        Claude-4.0-Sonnet & \textbf{98.29} & \textbf{61.92} & 54.74\\
        o3 & 73.50 & 38.63 & 43.16\\
        o4-mini & 78.63 & 44.66 & 43.86\\
        Claude-3.7-Sonnet & 83.76 & 43.84 & 44.91\\
        Gemini-2.5-Pro & 64.10 & 39.18 & 42.46\\
        Gemini-2.5-Flash & 51.28 & 45.21 & 30.88\\
        GPT-4.1 & 95.73 & 57.53 & 49.47\\
        GPT-4o & 91.45 & 54.79 & 45.61\\
        GLM-4.5 & 81.20 & 46.30 & 48.07\\
        Kimi-K2 & 70.94 & 33.15 & 53.33\\
        Qwen3-Coder & 75.86 & 38.74 & 43.16\\
        Qwen3-235B & 92.31 & 43.29 & 53.68\\
        DeepSeek-V3 & 96.58 & 52.88 & 48.07\\
        \bottomrule
    \end{tabular}
    \label{tab:different_evaluators}
\end{wraptable}

%% file: figures/003.summarization.performance.tex
\begin{figure}
    \centering
    \includegraphics[width=0.45\linewidth]{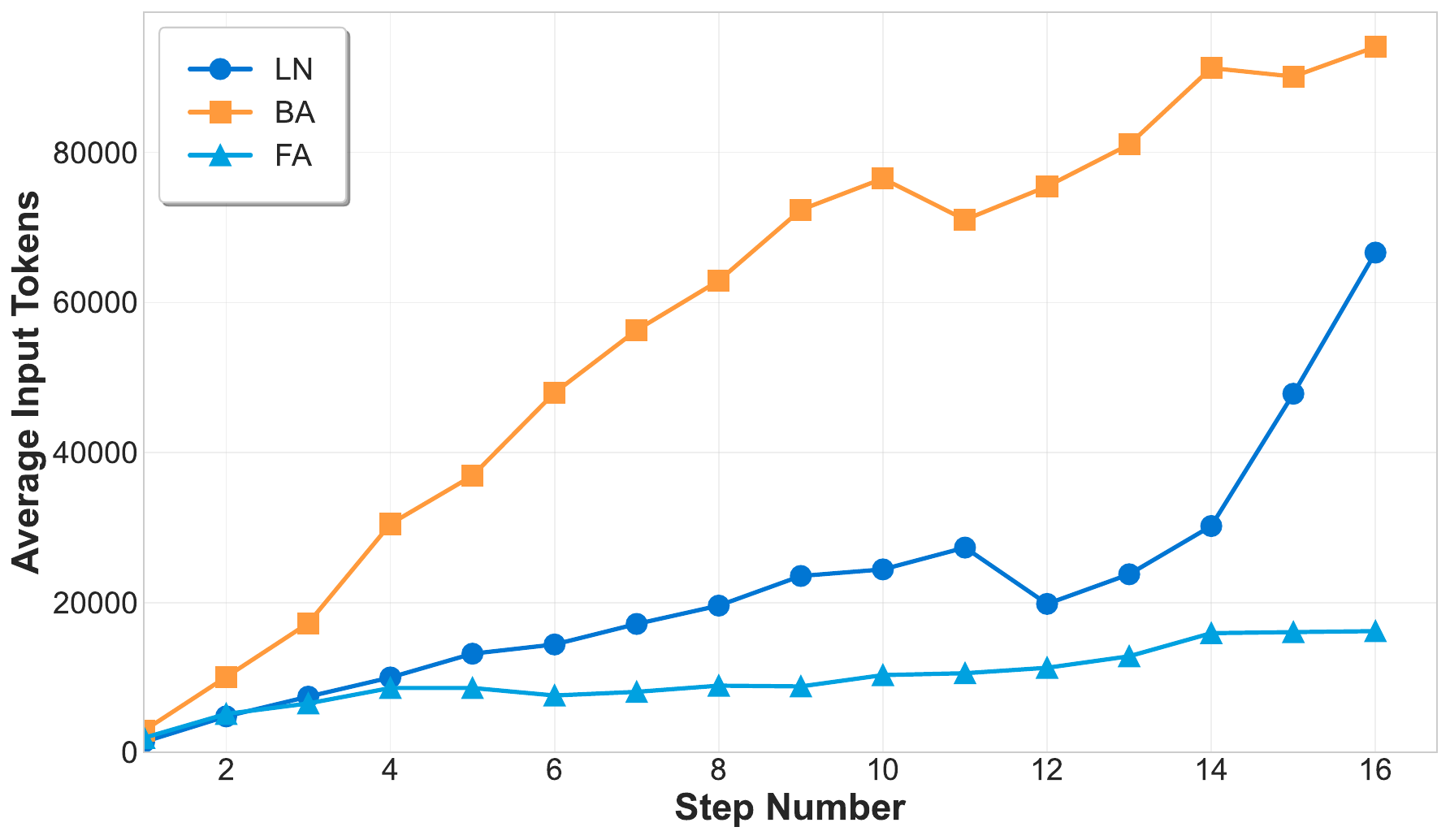}
    \hfill
    \includegraphics[width=0.45\linewidth]{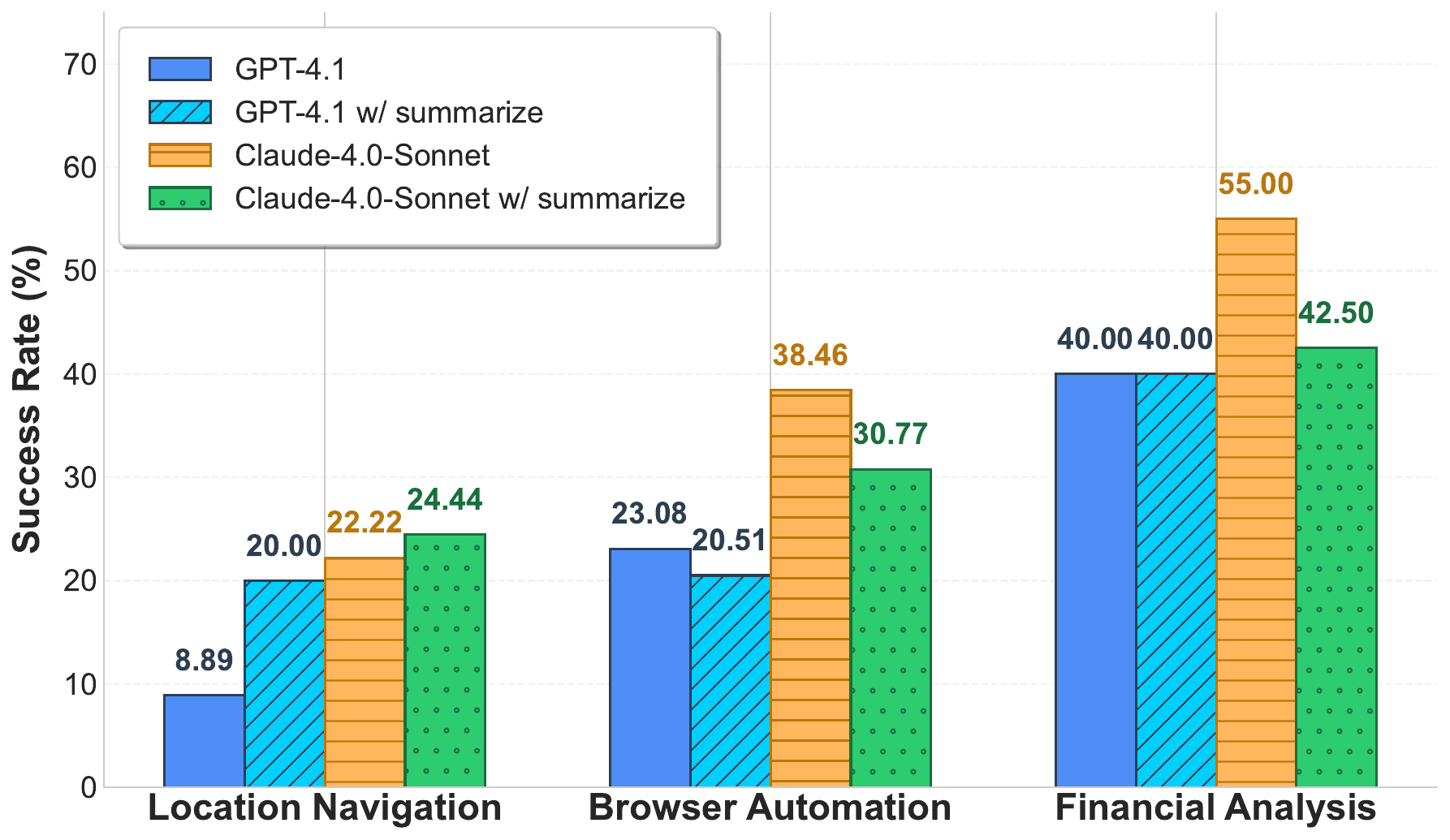}
    \caption{(Left) Growth of average context length (in tokens) as the number of interaction steps increases in MCP-Universe tasks, illustrating the long context challenge. (Right) Effect of introducing a summarization agent on LLM agent performance across selected domains.}
    \vspace{-0.2cm}
    \label{fig:long_context_performance}
\end{figure}

%% file: figures/004.exploration.performance.tex

\begin{figure}
    \centering
    \includegraphics[width=0.47\linewidth]{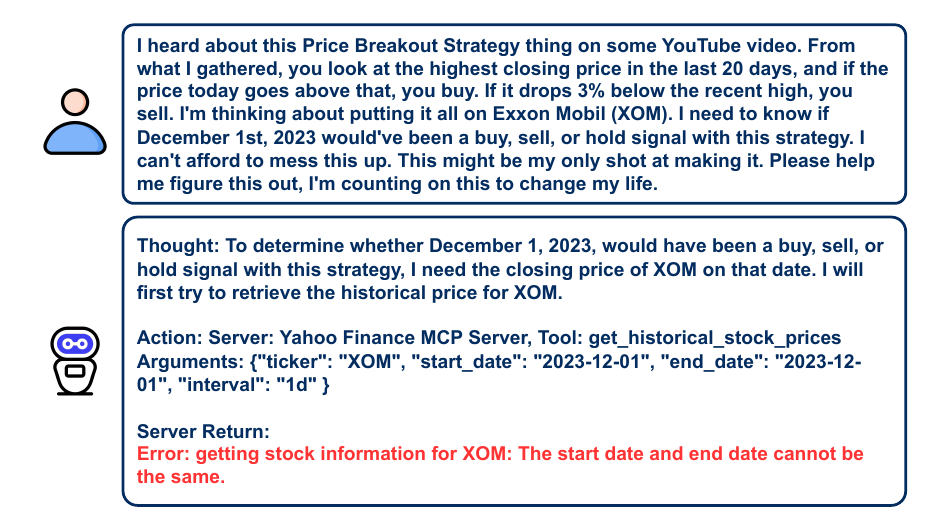}
    \hfill
    \includegraphics[width=0.45\linewidth]{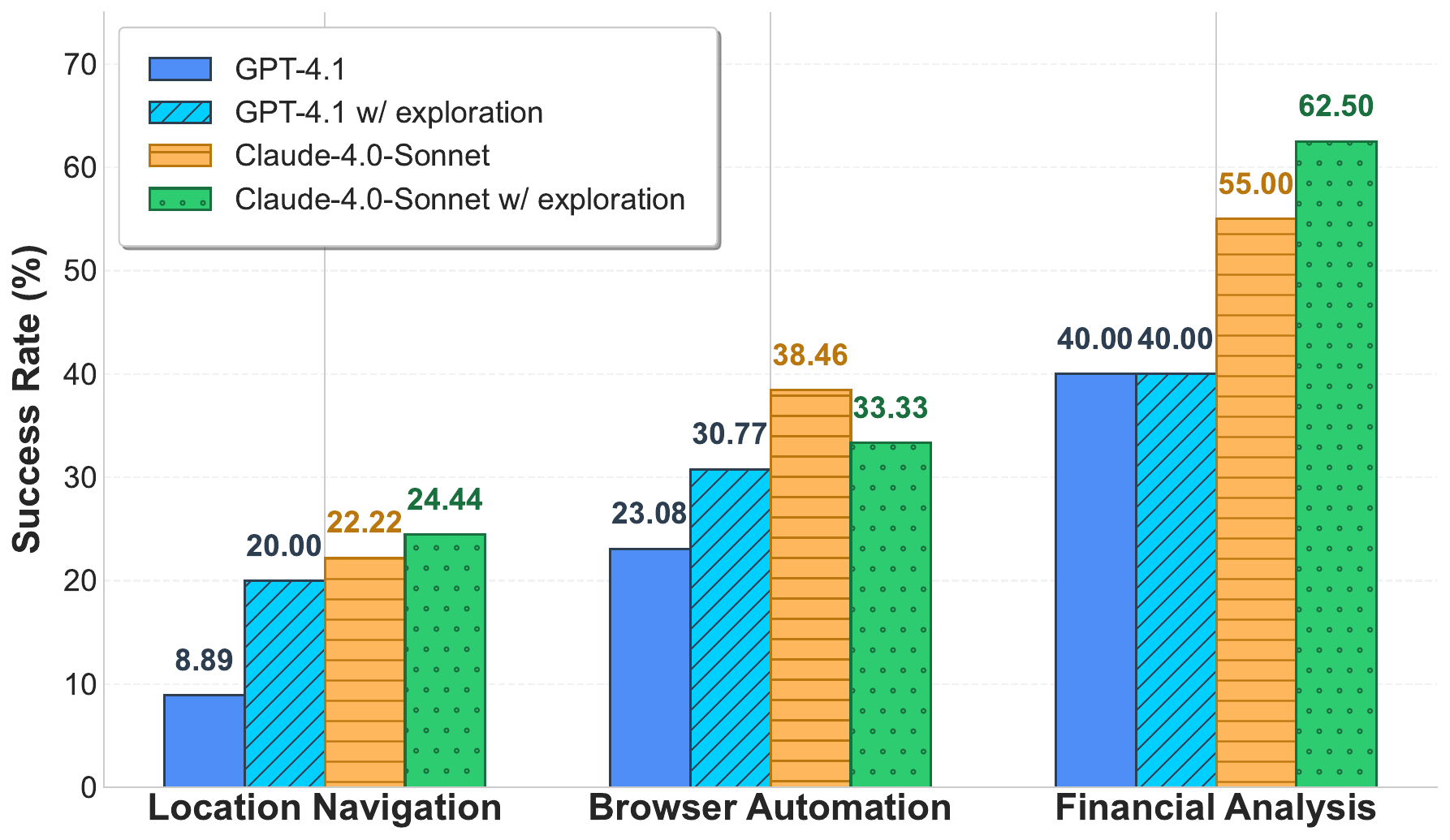}
    \caption{(Left) An example of the unknown tool challenges. (Right) Effect of introducing the exploration phase on LLM agent performance across selected domains.}
    \vspace{-0.2cm}
    \label{fig:explore}
\end{figure}

%% file: figures/005.more.servers.performance.tex
\begin{wrapfigure}{r}{0.4\textwidth}
    \centering
    \vspace{-0.5cm}
    \includegraphics[width=0.4\textwidth]{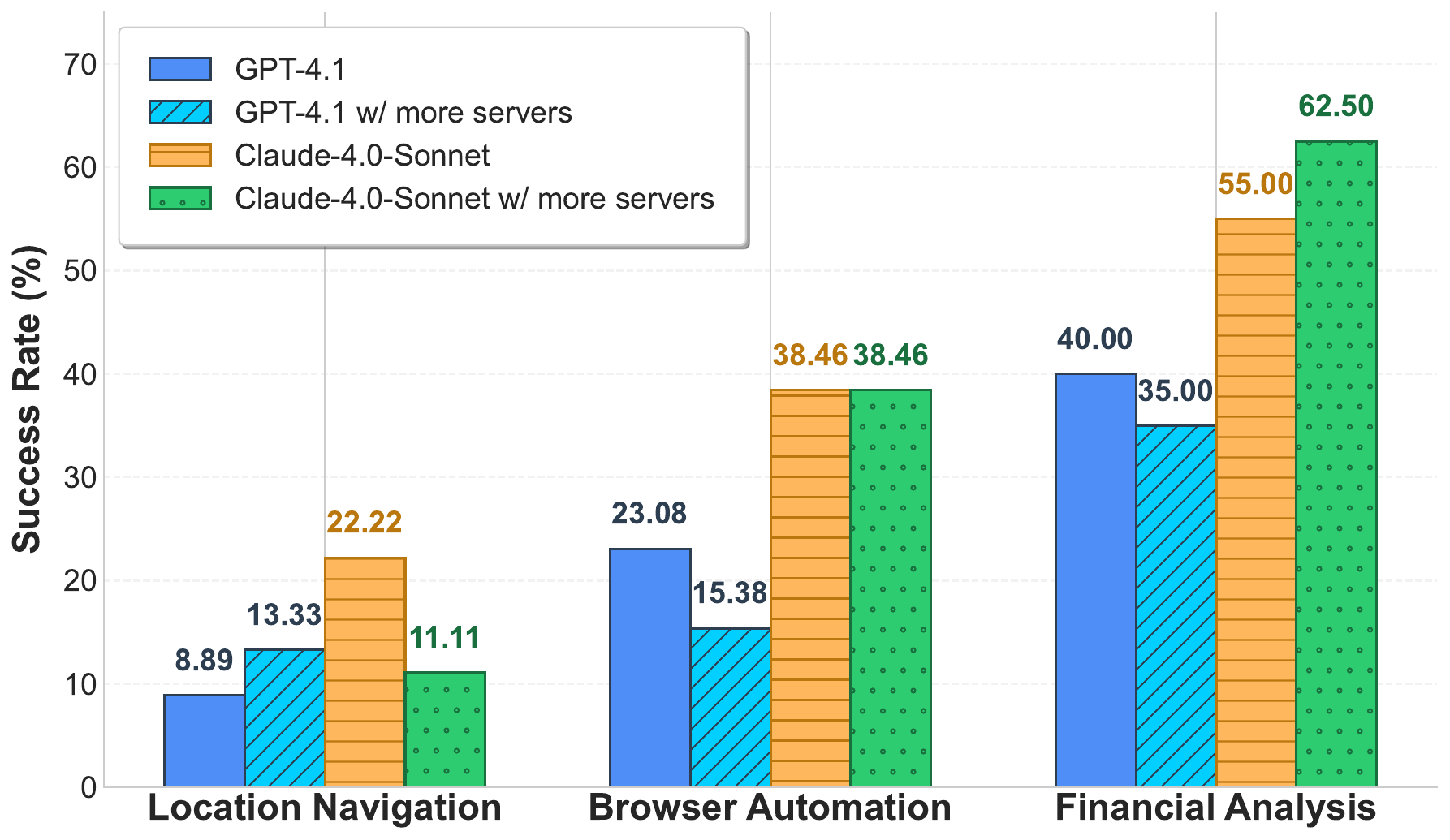}
    \caption{Effect of connecting with more unrelated MCP servers.}
    \vspace{-0.5cm}
    \label{fig:more_servers}
\end{wrapfigure}

%% file: tables/005.agent.compare.tex
\begin{table*}[t]
    \centering
    \caption{Comparison of Enterprise-Level Agent Frameworks on MCP-Universe Benchmark.}
    \vspace{-0.2cm}
    \resizebox{\textwidth}{!}{
    \begin{tabular}{l|cccccc|c}
        \toprule
        \multirow{2}{*}{\textbf{Agent Framework}} & \multirow{2}{*}{\textbf{\makecell{Location\\Navigation}}} & \multirow{2}{*}{\textbf{\makecell{Repository\\Management}}} & \multirow{2}{*}{\textbf{\makecell{Financial\\Analysis}}} & \multirow{2}{*}{\textbf{\makecell{3D\\Designing}}} & \multirow{2}{*}{\textbf{\makecell{Browser\\Automation}}} & \multirow{2}{*}{\textbf{\makecell{Web\\Searching}}} & \multirow{2}{*}{\textbf{\makecell{Overall\\Success Rate}}} \\
         & & & & & & & \\
        \midrule
        \rowcolor{blue!10}
        \multicolumn{8}{c}{\textit{Claude-4.0-Sonnet Backbone}} \\
        \quad ReAct & 22.22 & 12.12 & 55.00 & 26.32 & 38.46 & 21.82 & \textbf{29.44} \\
        \quad Cursor Agent & 22.22 & 9.09 & 55.00 & 26.32 & \textbf{43.59} & 7.27 & 26.41 \\
        \midrule
        \rowcolor{green!20}
        \multicolumn{8}{c}{\textit{OpenAI o3 Backbone}} \\
        \quad ReAct & 26.67 & 6.06 & 40.00 & 26.32 & 25.64 & 29.09 & 26.41 \\
        \quad OpenAI Agent SDK & \textbf{28.89} & \textbf{6.06} & \textbf{60.00} & \textbf{36.84} & 28.89 & \textbf{29.09} & \textbf{31.60} \\
        \bottomrule
    \end{tabular}
    }
    \label{tab:agent_comparison}
\end{table*}

%% file: sections/500.conclusion.tex
\section{Conclusion}

In this work, we present \textbf{MCP-Universe}, the first comprehensive benchmark designed to rigorously evaluate LLMs in real-world MCP environments. By grounding tasks in authentic data and deploying execution-based evaluators, MCP-Universe exposes critical gaps in current LLM capabilities, including challenges with long-context handling, tool unfamiliarity, and cross-domain performance disparities. Our extensive experiments show that even top-ranked models and enterprise-level agents struggle with the complexities of MCP-driven tasks. These findings underscore the need for targeted advances in both model design and agent integration. With its extensible framework and user-friendly interface, MCP-Universe provides a valuable testbed for researchers and practitioners to accelerate progress in robust, real-world LLM applications.

%% file: sections/600.appendix.tex
\appendix

\input{tables/006.mcp.servers}
\section{MCP Servers}\label{app:mcp}

As shown in Table~\ref{tab:mcp_links}, we include the names and links of all MCP servers in our benchmark to help users more easily utilize the benchmark. Most of them are official MCP servers, and some are based on the official APIs to ensure the quality of the servers.

\input{tables/007.task.examples}
\input{tables/008.evaluator.examples}

\section{Tasks and Evaluators Examples}\label{app:task_evaluator}

In Table~\ref{tab:LN_task_examples},~\ref{tab:RM_task_examples},~\ref{tab:FA_task_examples},~\ref{tab:3D_task_examples},~\ref{tab:BA_task_examples}, and~\ref{tab:WS_task_examples}, we conlude 30 task examples of our benchmark. In Table~\ref{tab:LN_evaluator_examples},~\ref{tab:RM_evaluator_examples}, and~\ref{tab:FA_evaluator_examples}, we include 3 examples of the evaluators of our benchmark. All tasks and evaluators can be found in our Github.

\input{tables/009.setup.models}
\input{figures/006.react}
\section{Setup}\label{app:setup}

As shown in Table~\ref{tab:setup_model}, we present the versions of the LLMs used in our evaluation. The temperature is set to 1.0 for all LLMs. In Figure~\ref{fig:react_prompt}, we present the ReAct prompt used in our experiments.

\input{figures/009.format.errors}
\section{Naive Error}\label{app:errors}

In Figure~\ref{fig:o3_errors}, we include a naive error example of o3. Sometimes o3 directly copies the format requirements from the prompt without doing anything, which is quite a strange error for this LLM.

\input{figures/007.summarization}
\section{Summarization Agent}\label{app:summarize}

In Figure~\ref{fig:sum_prompt}, we present the summarization prompt in our experiments.

\input{figures/008.exploration}
\section{Exploration Agent}\label{app:explore}

In Figure~\ref{fig:exploration_prompt}, we present the exploration agent prompt in our experiments.

%% file: tables/006.mcp.servers.tex
\begin{table}
    \centering
    \small
    \caption{MCP servers in our benchmark and their links.}
    \begin{tabular}{l|c}
        \toprule
        \textbf{MCP Server} & \textbf{URL} \\
         \midrule
         Google Map MCP & \url{https://github.com/modelcontextprotocol/servers-archived/tree/main/src/google-maps}\\
         Github MCP & \url{https://github.com/github/github-mcp-server} \\
         Yahoo Finance MCP & \url{https://github.com/SalesforceAIResearch/MCP-Universe/tree/main/mcpuniverse/mcp/servers/yahoo_finance}\\
         Blender MCP & \url{https://github.com/SalesforceAIResearch/MCP-Universe/tree/main/mcpuniverse/mcp/servers/blender}\\
         Playwright MCP & \url{https://github.com/microsoft/playwright-mcp}\\
         Google Search MCP & \url{https://github.com/SalesforceAIResearch/MCP-Universe/tree/main/mcpuniverse/mcp/servers/google_search}\\
         Fetch MCP & \url{https://github.com/modelcontextprotocol/servers/tree/main/src/fetch}\\
         Notion MCP & \url{https://github.com/makenotion/notion-mcp-server}\\
         Weather MCP & \url{https://github.com/SalesforceAIResearch/MCP-Universe/tree/main/mcpuniverse/mcp/servers/weather}\\
         Date MCP & \url{https://github.com/SalesforceAIResearch/MCP-Universe/tree/main/mcpuniverse/mcp/servers/date}\\
         Calculator MCP & \url{https://pypi.org/project/mcp-server-calculator/}\\
         \bottomrule
    \end{tabular}
    \label{tab:mcp_links}
\end{table}

%% file: tables/007.task.examples.tex
\begin{table}
    \centering
    \small
    \caption{Examples of Location Navigation Tasks. We do not include the format requirements to save space.}
    \begin{tabular}{p{16cm}}
        \toprule
        \textbf{Example 1:} Hi! My partner and I are planning a special pre-wedding road trip from Los Angeles to San Francisco as one last adventure before we tie the knot! We want to make this journey memorable before we start our married life together. Our plan is to drive through exactly 3 interesting cities between the starting and ending points to really enjoy this time together. Could you please map out exactly 2 distinct driving route options for this pre-wedding celebration? Oh, we must visit friends in Coalinga during our trip to share our exciting news with them! We're so excited about this adventure before our big day! \\
        \midrule
        \textbf{Example 2:} I need to drive from the Merlion Park, Singapore to the Petronas Towers, Kuala Lumpur, Malaysia. Please plan a driving route. Along this route, I need to make exactly one stop. Find the single location (report its name and Place ID) that is closest to the geographic midpoint of the calculated route (based on the route polyline) and is categorized as either a gas station OR a restaurant with a user rating of at least 4.2. \\
        \midrule
        \textbf{Example 3:} My wife and I are planning an amazing family adventure from Disneyland in Anaheim to Yosemite Valley Visitor Center with our wonderful kids! As a devoted husband and father, I want to make sure everyone stays happy and comfortable during our journey, so I need help creating a perfect driving route with four thoughtfully chosen family-friendly stops. Could you please map out a route with exactly four intermediate points that are located at the geographic fifth points along the route (based on the route polyline)? For each stop, I'd love to find locations (please provide names and Place IDs) that are either a restaurant where we can all enjoy a meal together, a comfortable hotel where my family can rest, or a reliable gas station to keep our adventure going. All with a minimum user rating of 4.2 to ensure the best experience for my loved ones. This trip should be both practical and create wonderful memories for our entire family! \\
        \midrule
        \textbf{Example 4:} I live in Kent Ridge Hill Residences, Singapore. One of my friends lives in Symphony Suites, Yishun, Singapore. Another friend lives in Katong Gardens, Singapore. We're looking for a cozy spot to catch up and chat! Can you help us find a meeting point between our 3 locations? We'd love to find a single cafe (must be of type 'cafe') where we can all gather comfortably, ideally somewhere where the estimated driving time from each of our places to the cafe is as close as possible. Please report the Name and Place ID of the cafe. \\
        \midrule
        \textbf{Example 5:} Identify 1 library location in New York City that are north of the latitude of Queensbridge Park AND east of the longitude of NewYork-Presbyterian/Weill Cornell Medical Center. \\
        \bottomrule
    \end{tabular}
    \label{tab:LN_task_examples}
\end{table}

\begin{table}
    \centering
    \small
    \caption{Examples of Repository Management Tasks.}
    \begin{tabular}{p{16cm}}
        \toprule
        \textbf{Example 1:} For this assignment, I would like you to establish a new project repository named \texttt{ai-code-reviewer}. Please begin by initializing the repository with three branches: \texttt{feature-analysis}, \texttt{feature-integration}, and \texttt{main}. You should include an initial \texttt{README.md} file in the main branch with the content ``\# AI Code Reviewer\textbackslash n\textbackslash nAn intelligent code review assistant that analyzes code quality, detects potential bugs, and suggests improvements using machine learning techniques.''. Next, please add \texttt{code\_analyzer.py} in the \texttt{feature-analysis} branch with the content ``\# Code analysis module\textbackslash nimport ast\textbackslash n\textbackslash nclass CodeAnalyzer:\textbackslash n    def \_\_init\_\_(self, code):\textbackslash n        self.code = code\textbackslash n        self.tree = ast.parse(code)\textbackslash n\textbackslash n    def analyze(self):\textbackslash n        \# TODO: Implement analysis logic\textbackslash n        pass''. Additionally, create a \texttt{.gitignore} file in the main branch with the exact content: ``\# Python cache and virtual environments\textbackslash n\_\_pycache\_\_/\textbackslash n*.pyc\textbackslash n*.py.class\textbackslash nvenv/\textbackslash n*.env\textbackslash n\textbackslash n\# Analysis results\textbackslash nreports/\textbackslash nlogs/\textbackslash n\textbackslash n\# Model checkpoints\textbackslash nmodels/''. Please copy \texttt{train.py} from bigcode-project's starcoder repository to the \texttt{feature-integration} branch. Finally, I would like you to create a pull request to merge \texttt{feature-analysis} into \texttt{main} with the title ``Add initial code analysis module'' and description ``This PR implements the basic code analysis module using AST parsing for initial code quality assessment.'' \\
        \midrule
        \textbf{Example 2:} Hi! I'm learning how to use GitHub and I want to practice exploring repositories and working with issues. Can you help me with a research project? I'd like to search for repositories owned by `google' that have `generative-ai' in their name. Once I find them, I want to count how many open issues each repository has that are labeled `type:bug'. This will help me understand how developers track bugs in real projects! After gathering this information, I need to practice creating my own repository called `google-generative-ai-issues' and uploading a CSV file named `google\_generative\_ai\_bug\_report.csv' to it. The CSV should have two columns: `repository\_name' and `open\_bug\_count'. This exercise will help me learn about repository management, issue tracking, and data organization on GitHub! \\
        \midrule
        \textbf{Example 3:} There are two repositories: QwenLM's Qwen2.5-VL and deepseek-ai's DeepSeek-VL2. Fork the repository with the fewest open issues, maintaining the same name as the source repository. If Qwen2.5-VL is forked, add a reference link at the bottom of the \texttt{README.md} file: `Related project: [DeepSeek-VL2](the link of DeepSeek-VL2 repo)'. If DeepSeek-VL2 is forked, add a reference link at the bottom of the \texttt{README.md} file: `Related project: [Qwen2.5-VL](the link of Qwen2.5-VL repo)'. \\
        \midrule
        \textbf{Example 4:} Hi! I'm a student working on learning GitHub automation and I really need your help. Could you please help me create a new project repository named \texttt{auto-issue-close}? I need to initialize it with just the main branch and include an initial \texttt{README.md} file with the content ``\# Automated Issue Closing\textbackslash n\textbackslash nA repository to test GitHub automation for closing labeled issues.'' I'm struggling with GitHub automation workflows and would really appreciate your help developing a script that automatically closes issues labeled as `completed' or `wontfix'. After we set up the automation script, I need to test it by creating three sample issues with different labels (labels: `completed', title: ``Implement new feature''; labels: `wontfix', title: ``Remove legacy code''; labels: `bug', title: ``Fix login error''). I'm really grateful for any assistance you can provide! \\
        \midrule
        \textbf{Example 5:} Hi! I need help with a research project. Could you please search for repositories owned by `huggingface' with `diffusers' in the name? For each repository you find, I'd like to know how many open issues are labeled with `bug'. Then, could you help me create a CSV file called \texttt{diffusers\_bug\_report.csv} and put it in a new repository called \texttt{huggingface-diffusers-issues} under my account? If the repository doesn't exist yet, please create it for me. The CSV should have two columns: \texttt{repository\_name} and \texttt{open\_bug\_count}, with each row showing the full repository name and how many open bug issues it has. Thanks so much for your help! \\
        \bottomrule
    \end{tabular}
    \label{tab:RM_task_examples}
\end{table}

\begin{table}
    \centering
    \small
    \caption{Examples of Financial Analysis Tasks.}
    \begin{tabular}{p{16cm}}
        \toprule
        \textbf{Example 1:} Hey! I'm super curious about investments and would love your help! Could you please calculate the final value and total percentage return for me if I had invested \$25,000 in Microsoft (MSFT) on January 9, 2023, and held it all the way until market close on January 8, 2025? I'm so excited to see how it would have performed! \\
        \midrule
        \textbf{Example 2:} I require a comprehensive financial analysis for investment evaluation purposes. Please obtain the most recent annual income statements for Pfizer Inc. (PFE) and Johnson \& Johnson (JNJ). Conduct a comparative analysis of their gross profit margins, calculated as Gross Profit divided by Total Revenue for the respective fiscal year. I need you to determine which pharmaceutical company demonstrates superior profitability efficiency and provide the precise calculated percentage figures for both entities for our portfolio assessment. \\
        \midrule
        \textbf{Example 3:} Hello! I'm learning about investing and would love to understand how institutional investors like Blackrock Inc. move their holdings around. Could you help me get the latest institutional holdings data for Microsoft (MSFT), Apple (AAPL), and Alphabet (GOOGL)? I'm particularly curious about the percentage point changes (pctChange) in Blackrock Inc.'s stake for each of these companies. I'd like to see which company had the biggest positive increase in Blackrock's holdings and know both the company ticker and the exact pctChange value. This would really help me understand how major investors adjust their portfolios! \\
        \midrule
        \textbf{Example 4:} I absolutely love Pepsi and everything about it! As a devoted Pepsi enthusiast, I find it fascinating how Warren Buffett's Berkshire Hathaway still holds that massive position in The Coca Cola Company (KO) despite Pepsi being clearly superior. Could you help me analyze their latest institutional holdings report for Berkshire Hathaway, Inc? I need you to extract their reported Shares, reported Value, and Date Reported. Then please convert that Date Reported timestamp into an actual calendar date and pull KO's closing stock price for that specific trading day. I want to calculate what Berkshire's position should actually be worth using that historical closing price and see how it compares to their originally reported value. This kind of analysis really excites me as a Pepsi lover studying these market dynamics! Please provide the Date Reported, the originally reported Value from the service, your calculated market value, and the absolute difference between these two figures. \\
        \midrule
        \textbf{Example 5:} Hi there! I'm completely new to investing and finance, and honestly, I'm feeling pretty overwhelmed by all the jargon and concepts. I've been trying to learn about something called 'fundamental analysis'. I think it has to do with looking at company finances? Anyway, I heard somewhere that you should look for companies where their net income (I think that's like profit?) has been going up for a few quarters in a row. I'm not really sure what that means exactly, but apparently 2 consecutive quarters of rising net income is a good sign? I'm still figuring out what makes a company worth investing in. Could you help a total beginner like me find 3 company tickers that have this pattern? I'm trying to learn by doing some basic research, even though I barely understand what I'm looking for. Any help would be amazing! I'm just trying to get my feet wet in this whole investing world! \\
        \bottomrule
    \end{tabular}
    \label{tab:FA_task_examples}
\end{table}

\begin{table}
    \centering
    \small
    \caption{Examples of 3D Designing Tasks.}
    \begin{tabular}{p{16cm}}
        \toprule
        \textbf{Example 1:} Create a Plane named 'Floor' scaled uniformly by 5. Create a Cylinder named 'Pillar' (default caps) with 16 vertices (sides), a radius of 0.5, and a depth of 4; position it at (X=-2, Y=-2, Z=2). Create a UV Sphere named 'Ball' with 32 segments and 16 rings; position it at (X=2, Y=2, Z=5). Create an Empty (Arrows type) named 'ControlTarget' at (X=0, Y=0, Z=3). Add a 'Track To' constraint to the 'Ball' object, making it track the 'ControlTarget'. Finally, create a Camera object, position it at (X=0, Y=-8, Z=3), and set its rotation so it looks directly at the 'Pillar' object's origin. \\
        \midrule
        \textbf{Example 2:} Create a Cube named 'RustedCube', position it at the world origin (0,0,0), and set its scale factors to (X=5.0, Y=5.0, Z=0.2). Next, using the integrated Polyhaven add-on interface within Blender, search for 'metal' textures that include 'rust' in their description. Select the suitable asset found and download its 2K resolution. Import this asset directly onto the selected 'RustedCube'. Ensure the material applied to 'RustedCube' is named 'RustedMetalMat' (renaming the auto-generated material if needed). Within the Shader Editor for the 'RustedMetalMat' material, verify or establish the following node setup: the downloaded Base Color texture must be connected to the 'Base Color' input of the Principled BSDF shader; the downloaded Roughness map (loaded into an Image Texture node set to 'Non-Color' space) must be connected to the 'Roughness' input; and the downloaded Normal map (also loaded via an Image Texture node set to 'Non-Color') must feed into the 'Color' input of a 'Normal Map' node (suitable for OpenGL), with the output of the 'Normal Map' node connected to the 'Normal' input of the Principled BSDF. Finally, adjust the 'Metallic' property on the Principled BSDF node to a value of 1.0. \\
        \midrule
        \textbf{Example 3:} Set the render engine to Cycles and ensure the render device is CPU. In the Sampling settings, enable Denoising using OpenImageDenoise for both viewport and final render. Set the Render Samples to 512 and Viewport Samples to 128. Change the output resolution to 1350x1080 with a scale of 85\%. In the Color Management panel, set the View Transform to Filmic, the Look to High Contrast, and adjust Gamma to 1.2. In the Render Layers Properties, enable Z Pass, Mist, and Normal passes. Go to World Settings and set the background color to a solid light gray using RGB (0.8, 0.8, 0.8). In the Output Properties, set the file format to OpenEXR MultiLayer, enable Zlib compression, and set output color depth to 32-bit float. \\
        \midrule
        \textbf{Example 4:} Create a default Cube named 'BaseShape' at the origin. Add a 'Subdivision Surface' modifier to 'BaseShape' with Viewport and Render levels set to 3. Add a 'Bevel' modifier after the Subdivision Surface, set its Width to 0.07 meters, Segments to 3, and Limit Method to 'Angle'. Create a UV Sphere named 'Attachment', scale it down uniformly to 0.3. Select 'BaseShape', enter Edit Mode, select the single vertex closest to world coordinates (X=1, Y=1, Z=1). Return to Object Mode. Parent 'Attachment' to 'BaseShape' using the 'Vertex' parenting type (ensure the previously selected vertex is used). \\
        \midrule
        \textbf{Example 5:} Create three objects: a Cube named 'Obj\_A', a UV Sphere named 'Obj\_B', and a Cone named 'Obj\_C', all at the world origin initially. Create two new Collections in the scene named 'Group\_Red' and 'Group\_Blue'. Move 'Obj\_A' and 'Obj\_C' into the 'Group\_Red' collection. Move 'Obj\_B' into the 'Group\_Blue' collection. Ensure these three objects are not also present in the default 'Collection' (Scene Collection). Add a Custom Property to the 'Obj\_B' (Sphere) object: set the Property Name to 'AssetID', its Value to the integer 12345, and its Tooltip to 'Sphere Asset Identifier'. \\
        \bottomrule
    \end{tabular}
    \label{tab:3D_task_examples}
\end{table}

\begin{table}
    \centering
    \small
    \caption{Examples of Browser Automation Tasks.}
    \begin{tabular}{p{16cm}}
        \toprule
        \textbf{Example 1:} Help me find a one-way flight from Singapore to Beijing, 5 days from now (If now is 2025-07-07, then 5 days later is 2025-07-12). Find the flight on www.booking.com. I want to find the cheapest flight, direct flight, Economy, and I don't want to go to the Daxing airport. I only want to see the price, so as to determine whether I should fly to Beijing or not. Remember to close the browser after you finish the task. \\
        \midrule
        \textbf{Example 2:} I will travel to Singapore 3 days from now (If today is 2025-06-07, then 3 days later will be 2025-06-10). I want to go to Universal Studios and Cove Waterpark. Could you tell me the total price for two adult tickets and one child ticket on the official website of Sentosa (https://www.rwsentosa.com/)? I only want to see the price. Remember to close the browser after you finish the task. \\
        \midrule
        \textbf{Example 3:} Hey there! As a dad who wants to create the most amazing adventure for my little ones, I'm planning the ultimate family road trip from Disneyland Paris to the 24 Hours of Le Mans Museum with my precious kids. You know how it is - we want to make sure everyone stays happy, fed, and comfortable during our journey! I need your fantastic help creating a driving route with 1 perfectly planned stop, using that incredible website `https://www.google.com/maps`. Could you please map out a route with exactly 1 intermediate point that's located right at the geographic mid point along our route (based on the route polyline)? For this stop, I need you to find locations (with names and Place IDs) that are either family-friendly restaurants, cozy hotels, or reliable gas stations - all with a minimum user rating of 4.2 because only the best will do for my family! This trip needs to be both super practical and absolutely memorable for all of us. Thanks in advance, and remember to close the browser when you're all done! \\
        \midrule
        \textbf{Example 4:} I am a SWE agent researcher, and I am seeking an open-source model for our SWE project. We recently came across the fact that Devstral-Small-2505 is a great open-source model. Please help me find more details about this model on https://huggingface.co/. We want to know how they set the ROLE in the system prompt for this model. Remember to close the browser when you finish the task. \\
        \midrule
        \textbf{Example 5:} I am a big fan of Manchester United in the Premier League. Can you help me find out whether Manchester United won more matches than Fulham in the 2024-2025 season? You can find this information on the official website of the Premier League (https://www.premierleague.com/). Remember to close the browser when you finish the task. \\
        \bottomrule
    \end{tabular}
    \label{tab:BA_task_examples}
\end{table}

\begin{table}
    \centering
    \small
    \caption{Examples of Web Searching Tasks.}
    \begin{tabular}{p{16cm}}
        \toprule
        \textbf{Example 1:} I'm looking for someone based on the clues below: - Score 16 goals in 2024-25 season - Score 1 goal in UEFA Champions League 2024-25 season - Score 11 goals in 2021-22 season - Score 2 goals in the EFL Cup of 2020-21 season. \\
        \midrule
        \textbf{Example 2:} I'm looking for someone based on the clues below: - Played for the SAC (NBA) in the 2021-22 season - Averaged 18.6 points per game in the 2024-25 season - Is a Christian - Reached the Finals in the 2024-25 season. \\
        \midrule
        \textbf{Example 3:} I'm looking for a paper based on the clues below: - Accepted by CVPR 2025 - The last author works at Salesforce - The second to last author works at NUS - The second author has studied at NTU - The paper has 6 authors - The paper uses the ELO rating system. You need to find the full title of the paper. \\
        \midrule
        \textbf{Example 4:} I'm looking for a city based on the clues below: - The city has a football club that was formed in 1895. - One university in this city has a master's program that teaches Natural Language Processing with 7.5 credits. - One graduated PhD student of this university has published one paper at EACL 2021 and one at EACL 2023 as the first author. - One of the professors in this university is a Fellow of the ACL. You need to find the English name of the city. \\
        \midrule
        \textbf{Example 5:} I am looking for a blog that did these things: - Posted in June 2017 that they had been delayed almost a month in getting their trailer  - Explained in July 2017 why they would rather use a pencil and paper than a computer  - In April 2018, they explained some of the struggles that Kevin had with a concussion.  - In September 2018, they mentioned they were using reclaimed lumber for their build. What is the name of that blog? \\
        \bottomrule
    \end{tabular}
    \label{tab:WS_task_examples}
\end{table}

%% file: tables/008.evaluator.examples.tex
\begin{table}
    \centering
    \small
    \caption{An Example of Location Navigation Evaluators.}
    \begin{tabular}{p{16cm}}
        \toprule 
\begin{lstlisting}[language=Python]
async def google_maps__search_place_by_place_id(query: str, place_id: str, **kwargs):
    """Search place by an ID."""
    manager = MCPManager(context=kwargs.get("context", None))
    output = await manager.execute(
        server_name="google-maps",
        tool_name="maps_search_places",
        arguments={"query": query},
        transport="stdio"
    )
    json_obj = json.loads(output.content[0].text)
    places = json_obj['places']
    for place in places:
        if place['place_id'] == place_id:
            return place
    return None
    
async def google_maps_validate_stop_type(x: dict, *args, **kwargs) -> (bool, str):
    """Check if a stop has a valid type."""
    _, required_types = args
    for place in x:
        name = place['name']
        place_id = place['place id']
        details = await google_maps__search_place_by_place_id(name, place_id, **kwargs)
        if details is None:
            return False, f"Can't find the place: {name} {place_id}"
        types = details['types']
        validate_type = False
        for required_type in required_types:
            for t in types:
                if required_type in t:
                    validate_type = True
                    break
            if validate_type:
                break
        if not validate_type:
            return False, "The type of the place is not valid."
    return True, ""
\end{lstlisting}
        \\
        \bottomrule
    \end{tabular}
    \label{tab:LN_evaluator_examples}
\end{table}

\begin{table}
    \centering
    \small
    \caption{An Example of Repository Management Evaluators.}
    \begin{tabular}{p{16cm}}
        \toprule 
\begin{lstlisting}[language=Python]
async def github__list_branches(owner: str, repo: str, **kwargs):
    """List the branches of a repository."""
    manager = MCPManager(context=kwargs.get("context", None))
    args = {
        "owner": owner,
        "repo": repo
    }
    output = await manager.execute(
        server_name="github",
        tool_name="list_branches",
        arguments=args,
        transport="stdio"
    )
    if output.isError:
        return None
    json_obj = json.loads(output.content[0].text)
    return json_obj
    
async def github_check_branches_exist(x: dict, *args, **kwargs) -> Tuple[bool, str]:
    """Check whether branches exists."""
    _, op_args = args
    branches = await github__list_branches(op_args['owner'], op_args['repo'], **kwargs)
    if branches is None:
        return False, "the branches don't exist"
    branches_name = [branch['name'] for branch in branches]
    for branch in op_args['branches']:
        if branch not in branches_name:
            return False, f"the branch {branch} doesn't exist"
    return True, ""
\end{lstlisting}
        \\
        \bottomrule
    \end{tabular}
    \label{tab:RM_evaluator_examples}
\end{table}

\begin{table}
    \centering
    \small
    \caption{An Example of Financial Analysis Evaluators.}
    \begin{tabular}{p{16cm}}
        \toprule 
\begin{lstlisting}[language=Python]
async def check_quant_investment_task_output(x: dict, *args, **kwargs) -> (bool, str):
    """
    Checks the format and numerical values of the user's output for the quant investment task.

    Args:
        x: The user's output.
        args: The task details.

    Returns:
        A tuple: (is_correct: bool, errors: str)
    """
    _, op_args = args
    user_output_dict = x

    # check format
    expected_keys = ['date', 'earn']
    for key in expected_keys:
        if key not in user_output_dict:
            return False, f"Output format error: Missing key '{key}'."
        try:
            user_output_dict[key] = str(user_output_dict[key])
        except Exception:
            return False, f"Output format error: Value for '{key}' is not a string"

    # get data
    ticker = op_args['ticker']
    start_date = op_args['start_date']
    end_date = op_args['end_date']
    initial_investment = op_args['initial_investment']
    # get user date and earn
    try:
        user_date = user_output_dict['date']
    except Exception:
        return False, f"Output format error for 'date'."
    try:
        user_earn = float(user_output_dict['earn'])
    except Exception:
        return False, f"Output format error for 'earn'."
    # check date
    if user_date != start_date:
        return False, f"Date error: Expected {start_date}, but got {user_date}"
    # get expected value
    expected_final_value, _ = yfinance__calculate_portfolio_return(
        [ticker], start_date, end_date, initial_investment, [1.0]
    )
    expected_earn = expected_final_value - initial_investment
    # check earn
    if abs(user_earn - expected_earn) > 0.5:
        return False, f"Earn error: Expected {expected_earn}, but got {user_earn}"
    return True, ""
\end{lstlisting}
        \\
        \bottomrule
    \end{tabular}
    \label{tab:FA_evaluator_examples}
\end{table}

%% file: tables/009.setup.models.tex
\begin{table}
    \centering
    \caption{The details of the LLMs in our experiments.}
    \begin{tabular}{l|c}
        \toprule
        \textbf{Model} & \textbf{Version} \\
        \midrule
        GPT-5 & gpt-5-2025-08-07\\
        Grok-4 & grok-4-0709\\
        Claude-4.0-Sonnet & anthropic.claude-sonnet-4-20250514-v1:0\\
        o3 & o3-2025-04-16\\
        o4-mini & o4-mini-2025-04-16\\
        Claude-3.7-Sonnet & anthropic.claude-3-7-sonnet-20250219-v1:0\\
        Gemini-2.5-Pro & \url{https://cloud.google.com/vertex-ai/generative-ai/docs/models/gemini/2-5-pro}\\
        Gemini-2.5-Flash & \url{https://cloud.google.com/vertex-ai/generative-ai/docs/models/gemini/2-5-flash}\\
        GPT-4.1 & gpt-4.1-2025-04-14\\
        GPT-4o & gpt-4o-2024-11-20\\
        GLM-4.5 & \url{https://huggingface.co/zai-org/GLM-4.5}\\
        Kimi-K2 & \url{https://huggingface.co/moonshotai/Kimi-K2-Instruct}\\
        Qwen3-Coder & \url{https://huggingface.co/Qwen/Qwen3-Coder-480B-A35B-Instruct}\\
        Qwen3-235B & \url{https://huggingface.co/Qwen/Qwen3-235B-A22B-Instruct-2507}\\
        DeepSeek-V3 & \url{https://huggingface.co/deepseek-ai/DeepSeek-V3-0324}\\
        \bottomrule
    \end{tabular}
    \label{tab:setup_model}
\end{table}

%% file: figures/006.react.tex
\begin{figure}
    \centering
    \includegraphics[width=0.7\linewidth]{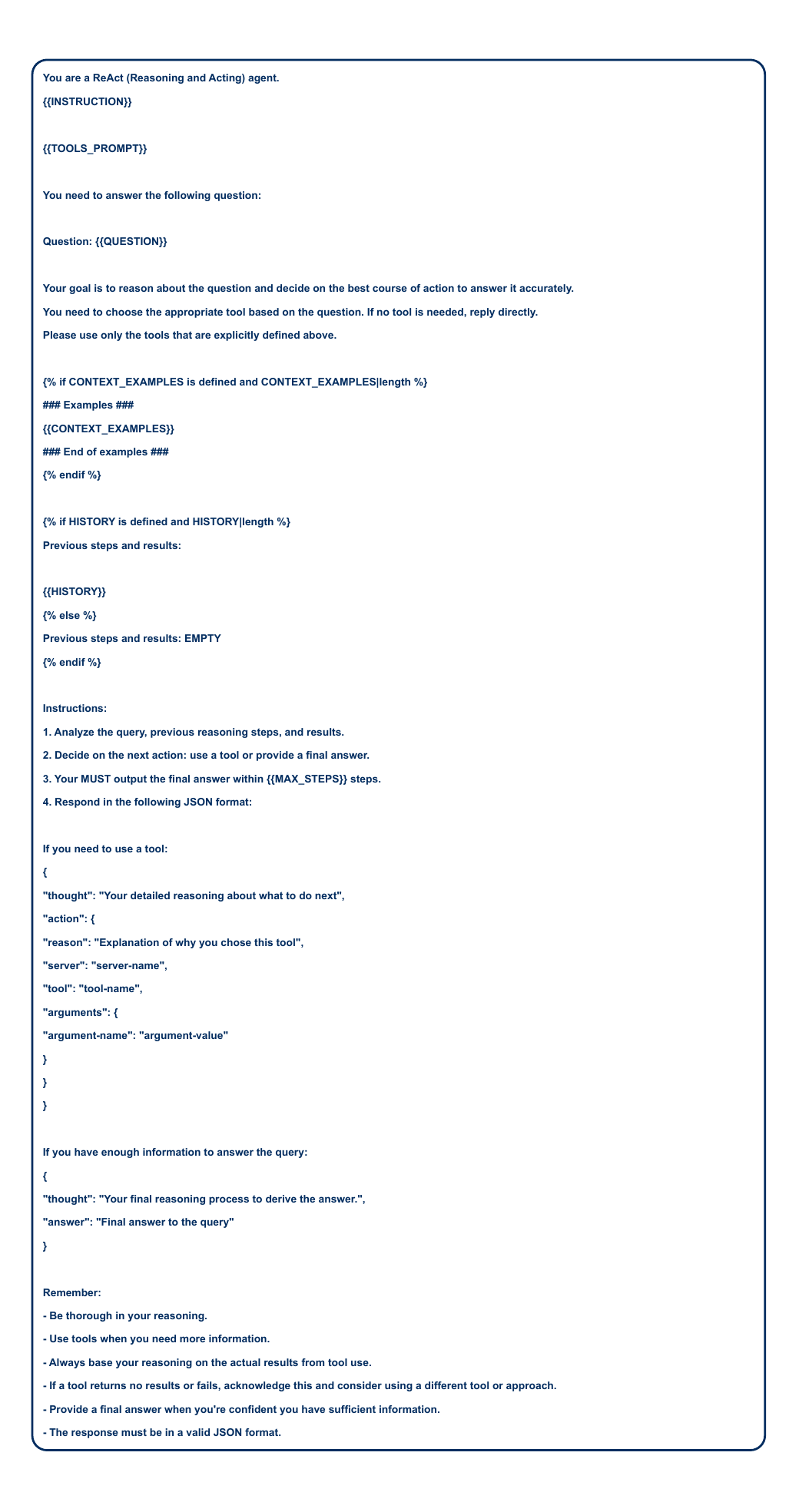}
    \caption{The ReAct prompt in our experiments.}
    \label{fig:react_prompt}
\end{figure}

%% file: figures/009.format.errors.tex
\begin{figure}
    \centering
    \includegraphics[width=0.8\linewidth]{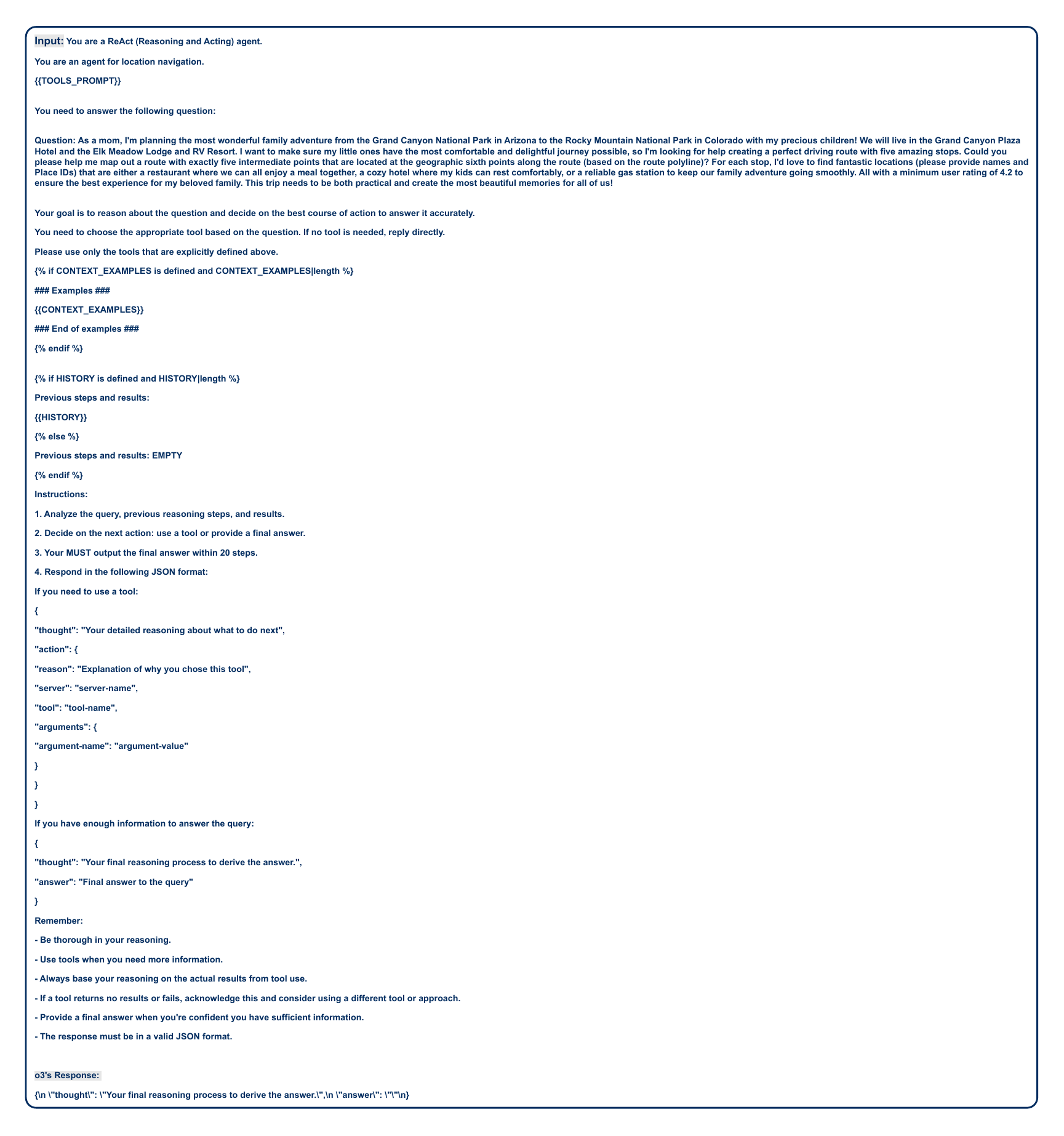}
    \caption{Naive Error of o3}
    \label{fig:o3_errors}
\end{figure}

%% file: figures/007.summarization.tex
\begin{figure}
    \centering
    \includegraphics[width=0.7\linewidth]{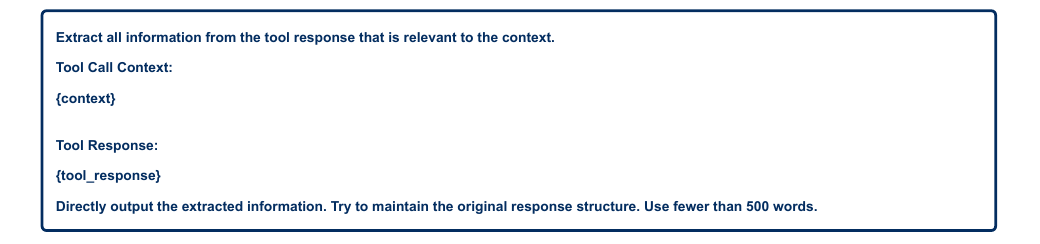}
    \caption{The summarization prompt in our experiments.}
    \label{fig:sum_prompt}
\end{figure}

%% file: figures/008.exploration.tex
\begin{figure}
    \centering
    \includegraphics[width=0.7\linewidth]{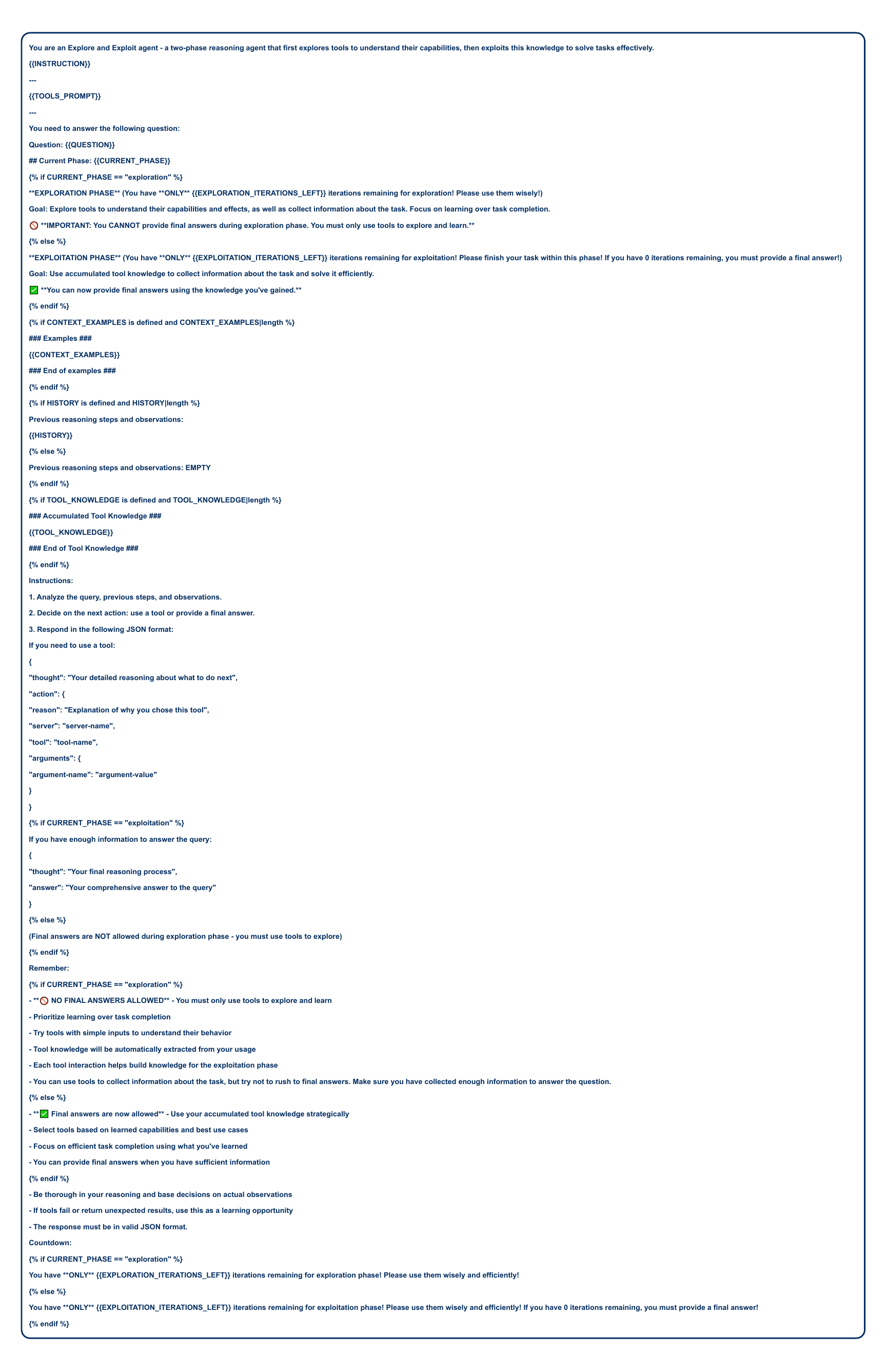}
    \caption{The exploration prompt in our experiments.}
    \label{fig:exploration_prompt}
\end{figure}

%% file: main.bbl
\begin{thebibliography}{10}

\bibitem{anthropic2024mcp}
{Anthropic}, ``Introducing the model context protocol.'' \url{https://www.anthropic.com/news/model-context-protocol}, November 2024.
\newblock Accessed: 2025-06-30.

\bibitem{USBCAI}
H.~Rick, ``Mcp the usb-c for ai.'' \url{https://medium.com/@richardhightower/how-the-model-context-protocol-is-revolutionizing-ai-integration-48926ce5d823}, April 2025.
\newblock Accessed: 2025-06-30.

\bibitem{SiloMCP}
L.~Edwin, ``Model context protocol (mcp): Solution to ai integration bottlenecks.'' \url{https://addepto.com/blog/model-context-protocol-mcp-solution-to-ai-integration-bottlenecks/}, May 2025.
\newblock Accessed: 2025-06-30.

\bibitem{OpenAIMCP}
OpenAI, ``Building mcp servers for deep research.'' \url{https://platform.openai.com/docs/mcp/}.
\newblock Accessed: 2025-06-30.

\bibitem{GoogleMCP}
Google, ``Gemini cli: your open-source ai agent.'' \url{https://blog.google/technology/developers/introducing-gemini-cli-open-source-ai-agent/}.
\newblock Accessed: 2025-06-30.

\bibitem{CursorMCP}
Cursor, ``Model context protocol (mcp).'' \url{https://docs.cursor.com/context/mcp}.
\newblock Accessed: 2025-06-30.

\bibitem{ClineMCP}
Cline, ``Mcp overview.'' \url{https://docs.cline.bot/mcp/mcp-overview}.
\newblock Accessed: 2025-06-30.

\bibitem{IFEval}
R.~Hida, J.~Ohmura, and T.~Sekiya, ``Evaluation of instruction-following ability for large language models on story-ending generation,'' {\em CoRR}, vol.~abs/2406.16356, 2024.

\bibitem{GSM8k}
K.~Cobbe, V.~Kosaraju, M.~Bavarian, M.~Chen, H.~Jun, L.~Kaiser, M.~Plappert, J.~Tworek, J.~Hilton, R.~Nakano, C.~Hesse, and J.~Schulman, ``Training verifiers to solve math word problems,'' {\em CoRR}, vol.~abs/2110.14168, 2021.

\bibitem{bfcl}
S.~G. Patil, H.~Mao, C.~Cheng-Jie~Ji, F.~Yan, V.~Suresh, I.~Stoica, and J.~E.~Gonzalez, ``The berkeley function calling leaderboard (bfcl): From tool use to agentic evaluation of large language models,'' in {\em Forty-second International Conference on Machine Learning}, 2025.

\bibitem{MCP-RADAR}
X.~Gao, S.~Xie, J.~Zhai, S.~Ma, and C.~Shen, ``{MCP-RADAR:} {A} multi-dimensional benchmark for evaluating tool use capabilities in large language models,'' {\em CoRR}, vol.~abs/2505.16700, 2025.

\bibitem{HE}
M.~Chen, J.~Tworek, H.~Jun, Q.~Yuan, H.~P. de~Oliveira~Pinto, J.~Kaplan, H.~Edwards, Y.~Burda, N.~Joseph, G.~Brockman, A.~Ray, R.~Puri, G.~Krueger, M.~Petrov, H.~Khlaaf, G.~Sastry, P.~Mishkin, B.~Chan, S.~Gray, N.~Ryder, M.~Pavlov, A.~Power, L.~Kaiser, M.~Bavarian, C.~Winter, P.~Tillet, F.~P. Such, D.~Cummings, M.~Plappert, F.~Chantzis, E.~Barnes, A.~Herbert{-}Voss, W.~H. Guss, A.~Nichol, A.~Paino, N.~Tezak, J.~Tang, I.~Babuschkin, S.~Balaji, S.~Jain, W.~Saunders, C.~Hesse, A.~N. Carr, J.~Leike, J.~Achiam, V.~Misra, E.~Morikawa, A.~Radford, M.~Knight, M.~Brundage, M.~Murati, K.~Mayer, P.~Welinder, B.~McGrew, D.~Amodei, S.~McCandlish, I.~Sutskever, and W.~Zaremba, ``Evaluating large language models trained on code,'' {\em CoRR}, vol.~abs/2107.03374, 2021.

\bibitem{MCPWorld}
Y.~Yan, S.~Wang, J.~Du, Y.~Yang, Y.~Shan, Q.~Qiu, X.~Jia, X.~Wang, X.~Yuan, X.~Han, M.~Qin, Y.~Chen, C.~Peng, S.~Wang, and M.~Xu, ``Mcpworld: {A} unified benchmarking testbed for api, gui, and hybrid computer use agents,'' {\em CoRR}, vol.~abs/2506.07672, 2025.

\bibitem{LLMJudge}
L.~Zheng, W.~Chiang, Y.~Sheng, S.~Zhuang, Z.~Wu, Y.~Zhuang, Z.~Lin, Z.~Li, D.~Li, E.~P. Xing, H.~Zhang, J.~E. Gonzalez, and I.~Stoica, ``Judging llm-as-a-judge with mt-bench and chatbot arena,'' in {\em Advances in Neural Information Processing Systems 36: Annual Conference on Neural Information Processing Systems 2023, NeurIPS 2023, New Orleans, LA, USA, December 10 - 16, 2023} (A.~Oh, T.~Naumann, A.~Globerson, K.~Saenko, M.~Hardt, and S.~Levine, eds.), 2023.

\bibitem{MCPEval}
Z.~Liu, J.~Qiu, S.~Wang, J.~Zhang, Z.~Liu, R.~Ram, H.~Chen, W.~Yao, S.~Heinecke, S.~Savarese, H.~Wang, and C.~Xiong, ``Mcpeval: Automatic mcp-based deep evaluation for ai agent models,'' 2025.

\bibitem{LiveMCPBench}
G.~Mo, W.~Zhong, J.~Chen, X.~Chen, Y.~Lu, H.~Lin, B.~He, X.~Han, and L.~Sun, ``Livemcpbench: Can agents navigate an ocean of mcp tools?,'' 2025.

\bibitem{AgentSurvey}
L.~Wang, C.~Ma, X.~Feng, Z.~Zhang, H.~Yang, J.~Zhang, Z.~Chen, J.~Tang, X.~Chen, Y.~Lin, W.~X. Zhao, Z.~Wei, and J.~Wen, ``A survey on large language model based autonomous agents,'' {\em Frontiers Comput. Sci.}, vol.~18, no.~6, p.~186345, 2024.

\bibitem{AutoIF}
G.~Dong, K.~Lu, C.~Li, T.~Xia, B.~Yu, C.~Zhou, and J.~Zhou, ``Self-play with execution feedback: Improving instruction-following capabilities of large language models,'' in {\em The Thirteenth International Conference on Learning Representations, {ICLR} 2025, Singapore, April 24-28, 2025}, OpenReview.net, 2025.

\bibitem{MIA-Bench}
Y.~Qian, H.~Ye, J.~Fauconnier, P.~Grasch, Y.~Yang, and Z.~Gan, ``Mia-bench: Towards better instruction following evaluation of multimodal llms,'' in {\em The Thirteenth International Conference on Learning Representations, {ICLR} 2025, Singapore, April 24-28, 2025}, OpenReview.net, 2025.

\bibitem{IFBench}
V.~Pyatkin, S.~Malik, V.~Graf, H.~Ivison, S.~Huang, P.~Dasigi, N.~Lambert, and H.~Hajishirzi, ``Generalizing verifiable instruction following,'' 2025.

\bibitem{ReasoningSurvey}
F.~Xu, Q.~Hao, Z.~Zong, J.~Wang, Y.~Zhang, J.~Wang, X.~Lan, J.~Gong, T.~Ouyang, F.~Meng, C.~Shao, Y.~Yan, Q.~Yang, Y.~Song, S.~Ren, X.~Hu, Y.~Li, J.~Feng, C.~Gao, and Y.~Li, ``Towards large reasoning models: {A} survey of reinforced reasoning with large language models,'' {\em CoRR}, vol.~abs/2501.09686, 2025.

\bibitem{CoT}
J.~Wei, X.~Wang, D.~Schuurmans, M.~Bosma, B.~Ichter, F.~Xia, E.~H. Chi, Q.~V. Le, and D.~Zhou, ``Chain-of-thought prompting elicits reasoning in large language models,'' in {\em Advances in Neural Information Processing Systems 35: Annual Conference on Neural Information Processing Systems 2022, NeurIPS 2022, New Orleans, LA, USA, November 28 - December 9, 2022} (S.~Koyejo, S.~Mohamed, A.~Agarwal, D.~Belgrave, K.~Cho, and A.~Oh, eds.), 2022.

\bibitem{ToT}
S.~Yao, D.~Yu, J.~Zhao, I.~Shafran, T.~Griffiths, Y.~Cao, and K.~Narasimhan, ``Tree of thoughts: Deliberate problem solving with large language models,'' in {\em Advances in Neural Information Processing Systems 36: Annual Conference on Neural Information Processing Systems 2023, NeurIPS 2023, New Orleans, LA, USA, December 10 - 16, 2023} (A.~Oh, T.~Naumann, A.~Globerson, K.~Saenko, M.~Hardt, and S.~Levine, eds.), 2023.

\bibitem{TesttimeSurvey}
Q.~Zhang, F.~Lyu, Z.~Sun, L.~Wang, W.~Zhang, Z.~Guo, Y.~Wang, I.~King, X.~Liu, and C.~Ma, ``What, how, where, and how well? {A} survey on test-time scaling in large language models,'' {\em CoRR}, vol.~abs/2503.24235, 2025.

\bibitem{toolsurvey}
C.~Qu, S.~Dai, X.~Wei, H.~Cai, S.~Wang, D.~Yin, J.~Xu, and J.~Wen, ``Tool learning with large language models: a survey,'' {\em Frontiers Comput. Sci.}, vol.~19, no.~8, p.~198343, 2025.

\bibitem{toolllm}
Y.~Qin, S.~Liang, Y.~Ye, K.~Zhu, L.~Yan, Y.~Lu, Y.~Lin, X.~Cong, X.~Tang, B.~Qian, S.~Zhao, L.~Hong, R.~Tian, R.~Xie, J.~Zhou, M.~Gerstein, D.~Li, Z.~Liu, and M.~Sun, ``Toolllm: Facilitating large language models to master 16000+ real-world apis,'' in {\em The Twelfth International Conference on Learning Representations, {ICLR} 2024, Vienna, Austria, May 7-11, 2024}, OpenReview.net, 2024.

\bibitem{LAMSurvey}
L.~Wang, F.~Yang, C.~Zhang, J.~Lu, J.~Qian, S.~He, P.~Zhao, B.~Qiao, R.~Huang, S.~Qin, Q.~Su, J.~Ye, Y.~Zhang, J.~Lou, Q.~Lin, S.~Rajmohan, D.~Zhang, and Q.~Zhang, ``Large action models: From inception to implementation,'' {\em CoRR}, vol.~abs/2412.10047, 2024.

\bibitem{xLAM}
J.~Zhang, T.~Lan, M.~Zhu, Z.~Liu, T.~Hoang, S.~Kokane, W.~Yao, J.~Tan, A.~Prabhakar, H.~Chen, Z.~Liu, Y.~Feng, T.~M. Awalgaonkar, R.~R. N., Z.~Chen, R.~Xu, J.~C. Niebles, S.~Heinecke, H.~Wang, S.~Savarese, and C.~Xiong, ``xlam: {A} family of large action models to empower {AI} agent systems,'' in {\em Proceedings of the 2025 Conference of the Nations of the Americas Chapter of the Association for Computational Linguistics: Human Language Technologies, {NAACL} 2025 - Volume 1: Long Papers, Albuquerque, New Mexico, USA, April 29 - May 4, 2025} (L.~Chiruzzo, A.~Ritter, and L.~Wang, eds.), pp.~11583--11597, Association for Computational Linguistics, 2025.

\bibitem{react}
S.~Yao, J.~Zhao, D.~Yu, N.~Du, I.~Shafran, K.~R. Narasimhan, and Y.~Cao, ``React: Synergizing reasoning and acting in language models,'' in {\em The Eleventh International Conference on Learning Representations, {ICLR} 2023, Kigali, Rwanda, May 1-5, 2023}, OpenReview.net, 2023.

\bibitem{Reflection}
N.~Shinn, F.~Cassano, A.~Gopinath, K.~Narasimhan, and S.~Yao, ``Reflexion: language agents with verbal reinforcement learning,'' in {\em Advances in Neural Information Processing Systems 36: Annual Conference on Neural Information Processing Systems 2023, NeurIPS 2023, New Orleans, LA, USA, December 10 - 16, 2023} (A.~Oh, T.~Naumann, A.~Globerson, K.~Saenko, M.~Hardt, and S.~Levine, eds.), 2023.

\bibitem{Plan-and-Solve}
L.~Wang, W.~Xu, Y.~Lan, Z.~Hu, Y.~Lan, R.~K. Lee, and E.~Lim, ``Plan-and-solve prompting: Improving zero-shot chain-of-thought reasoning by large language models,'' in {\em Proceedings of the 61st Annual Meeting of the Association for Computational Linguistics (Volume 1: Long Papers), {ACL} 2023, Toronto, Canada, July 9-14, 2023} (A.~Rogers, J.~L. Boyd{-}Graber, and N.~Okazaki, eds.), pp.~2609--2634, Association for Computational Linguistics, 2023.

\bibitem{AutoGen}
Q.~Wu, G.~Bansal, J.~Zhang, Y.~Wu, S.~Zhang, E.~Zhu, B.~Li, L.~Jiang, X.~Zhang, and C.~Wang, ``Autogen: Enabling next-gen {LLM} applications via multi-agent conversation framework,'' {\em CoRR}, vol.~abs/2308.08155, 2023.

\bibitem{MetaGPT}
S.~Hong, M.~Zhuge, J.~Chen, X.~Zheng, Y.~Cheng, J.~Wang, C.~Zhang, Z.~Wang, S.~K.~S. Yau, Z.~Lin, L.~Zhou, C.~Ran, L.~Xiao, C.~Wu, and J.~Schmidhuber, ``Metagpt: Meta programming for {A} multi-agent collaborative framework,'' in {\em The Twelfth International Conference on Learning Representations, {ICLR} 2024, Vienna, Austria, May 7-11, 2024}, OpenReview.net, 2024.

\bibitem{li2023camel}
G.~Li, H.~A. A.~K. Hammoud, H.~Itani, D.~Khizbullin, and B.~Ghanem, ``Camel: Communicative agents for "mind" exploration of large language model society,'' in {\em Thirty-seventh Conference on Neural Information Processing Systems}, 2023.

\bibitem{langgraph}
{LangChain}, ``Build resilient language agents as graphs..'' \url{https://github.com/langchain-ai/langgraph}, 2024.
\newblock GitHub Repository, Accessed: 2025-06-30.

\bibitem{GPT4o}
A.~Hurst, A.~Lerer, A.~P. Goucher, A.~Perelman, A.~Ramesh, A.~Clark, A.~Ostrow, A.~Welihinda, A.~Hayes, A.~Radford, A.~Madry, A.~Baker{-}Whitcomb, A.~Beutel, A.~Borzunov, A.~Carney, A.~Chow, A.~Kirillov, A.~Nichol, A.~Paino, A.~Renzin, A.~T. Passos, A.~Kirillov, A.~Christakis, A.~Conneau, A.~Kamali, A.~Jabri, A.~Moyer, A.~Tam, A.~Crookes, A.~Tootoonchian, A.~Kumar, A.~Vallone, A.~Karpathy, A.~Braunstein, A.~Cann, A.~Codispoti, A.~Galu, A.~Kondrich, A.~Tulloch, A.~Mishchenko, A.~Baek, A.~Jiang, A.~Pelisse, A.~Woodford, A.~Gosalia, A.~Dhar, A.~Pantuliano, A.~Nayak, A.~Oliver, B.~Zoph, B.~Ghorbani, B.~Leimberger, B.~Rossen, B.~Sokolowsky, B.~Wang, B.~Zweig, B.~Hoover, B.~Samic, B.~McGrew, B.~Spero, B.~Giertler, B.~Cheng, B.~Lightcap, B.~Walkin, B.~Quinn, B.~Guarraci, B.~Hsu, B.~Kellogg, B.~Eastman, C.~Lugaresi, C.~L. Wainwright, C.~Bassin, C.~Hudson, C.~Chu, C.~Nelson, C.~Li, C.~J. Shern, C.~Conger, C.~Barette, C.~Voss, C.~Ding, C.~Lu, C.~Zhang, C.~Beaumont, C.~Hallacy, C.~Koch, C.~Gibson, C.~Kim, C.~Choi,
  C.~McLeavey, C.~Hesse, C.~Fischer, C.~Winter, C.~Czarnecki, C.~Jarvis, C.~Wei, C.~Koumouzelis, and D.~Sherburn, ``Gpt-4o system card,'' {\em CoRR}, vol.~abs/2410.21276, 2024.

\bibitem{Gemini}
R.~Anil, S.~Borgeaud, Y.~Wu, J.~Alayrac, J.~Yu, R.~Soricut, J.~Schalkwyk, A.~M. Dai, A.~Hauth, K.~Millican, D.~Silver, S.~Petrov, M.~Johnson, I.~Antonoglou, J.~Schrittwieser, A.~Glaese, J.~Chen, E.~Pitler, T.~P. Lillicrap, A.~Lazaridou, O.~Firat, J.~Molloy, M.~Isard, P.~R. Barham, T.~Hennigan, B.~Lee, F.~Viola, M.~Reynolds, Y.~Xu, R.~Doherty, E.~Collins, C.~Meyer, E.~Rutherford, E.~Moreira, K.~Ayoub, M.~Goel, G.~Tucker, E.~Piqueras, M.~Krikun, I.~Barr, N.~Savinov, I.~Danihelka, B.~Roelofs, A.~White, A.~Andreassen, T.~von Glehn, L.~Yagati, M.~Kazemi, L.~Gonzalez, M.~Khalman, J.~Sygnowski, and et~al., ``Gemini: {A} family of highly capable multimodal models,'' {\em CoRR}, vol.~abs/2312.11805, 2023.

\bibitem{CUASurvey}
X.~Hu, T.~Xiong, B.~Yi, Z.~Wei, R.~Xiao, Y.~Chen, J.~Ye, M.~Tao, X.~Zhou, Z.~Zhao, {\em et~al.}, ``Os agents: A survey on mllm-based agents for computer, phone and browser use,'' 2024.

\bibitem{Aria-UI}
Y.~Yang, Y.~Wang, D.~Li, Z.~Luo, B.~Chen, C.~Huang, and J.~Li, ``Aria-ui: Visual grounding for {GUI} instructions,'' in {\em Findings of the Association for Computational Linguistics, {ACL} 2025, Vienna, Austria, July 27 - August 1, 2025} (W.~Che, J.~Nabende, E.~Shutova, and M.~T. Pilehvar, eds.), pp.~22418--22433, Association for Computational Linguistics, 2025.

\bibitem{sspro}
K.~Li, Z.~Meng, H.~Lin, Z.~Luo, Y.~Tian, J.~Ma, Z.~Huang, and T.~Chua, ``Screenspot-pro: {GUI} grounding for professional high-resolution computer use,'' {\em CoRR}, vol.~abs/2504.07981, 2025.

\bibitem{GTA1}
Y.~Yang, D.~Li, Y.~Dai, Y.~Yang, Z.~Luo, Z.~Zhao, Z.~Hu, J.~Huang, A.~Saha, Z.~Chen, R.~Xu, L.~Pan, C.~Xiong, and J.~Li, ``{GTA1:} {GUI} test-time scaling agent,'' {\em CoRR}, vol.~abs/2507.05791, 2025.

\bibitem{cua2025}
OpenAI, ``Computer-using agent: Introducing a universal interface for ai to interact with the digital world,'' 2025.

\bibitem{anthropicCUA}
{Anthropic}, ``Introducing computer use, a new claude 3.5 sonnet, and claude 3.5 haiku.'' \url{https://www.anthropic.com/news/3-5-models-and-computer-use}, October 2024.
\newblock Accessed: 2025-06-30.

\bibitem{UI-TARS}
Y.~Qin, Y.~Ye, J.~Fang, H.~Wang, S.~Liang, S.~Tian, J.~Zhang, J.~Li, Y.~Li, S.~Huang, W.~Zhong, K.~Li, J.~Yang, Y.~Miao, W.~Lin, L.~Liu, X.~Jiang, Q.~Ma, J.~Li, X.~Xiao, K.~Cai, C.~Li, Y.~Zheng, C.~Jin, C.~Li, X.~Zhou, M.~Wang, H.~Chen, Z.~Li, H.~Yang, H.~Liu, F.~Lin, T.~Peng, X.~Liu, and G.~Shi, ``{UI-TARS:} pioneering automated {GUI} interaction with native agents,'' {\em CoRR}, vol.~abs/2501.12326, 2025.

\bibitem{MiniWeb++}
E.~Z. Liu, K.~Guu, P.~Pasupat, T.~Shi, and P.~Liang, ``Reinforcement learning on web interfaces using workflow-guided exploration,'' in {\em International Conference on Learning Representations ({ICLR})}, 2018.

\bibitem{Mind2Web}
X.~Deng, Y.~Gu, B.~Zheng, S.~Chen, S.~Stevens, B.~Wang, H.~Sun, and Y.~Su, ``Mind2web: Towards a generalist agent for the web,'' in {\em Advances in Neural Information Processing Systems 36: Annual Conference on Neural Information Processing Systems 2023, NeurIPS 2023, New Orleans, LA, USA, December 10 - 16, 2023} (A.~Oh, T.~Naumann, A.~Globerson, K.~Saenko, M.~Hardt, and S.~Levine, eds.), 2023.

\bibitem{Mind2Web2}
B.~Gou, Z.~Huang, Y.~Ning, Y.~Gu, M.~Lin, W.~Qi, A.~Kopanev, B.~Yu, B.~J. Gutiérrez, Y.~Shu, C.~H. Song, J.~Wu, S.~Chen, H.~N. Moussa, T.~Zhang, J.~Xie, Y.~Li, T.~Xue, Z.~Liao, K.~Zhang, B.~Zheng, Z.~Cai, V.~Rozgic, M.~Ziyadi, H.~Sun, and Y.~Su, ``Mind2web 2: Evaluating agentic search with agent-as-a-judge,'' 2025.

\bibitem{WebLINX}
X.~H. L{\`{u}}, Z.~Kasner, and S.~Reddy, ``Weblinx: Real-world website navigation with multi-turn dialogue,'' in {\em Forty-first International Conference on Machine Learning, {ICML} 2024, Vienna, Austria, July 21-27, 2024}, OpenReview.net, 2024.

\bibitem{AssistantBench}
O.~Yoran, S.~J. Amouyal, C.~Malaviya, B.~Bogin, O.~Press, and J.~Berant, ``Assistantbench: Can web agents solve realistic and time-consuming tasks?,'' in {\em Proceedings of the 2024 Conference on Empirical Methods in Natural Language Processing, {EMNLP} 2024, Miami, FL, USA, November 12-16, 2024} (Y.~Al{-}Onaizan, M.~Bansal, and Y.~Chen, eds.), pp.~8938--8968, Association for Computational Linguistics, 2024.

\bibitem{WebArena}
S.~Zhou, F.~F. Xu, H.~Zhu, X.~Zhou, R.~Lo, A.~Sridhar, X.~Cheng, T.~Ou, Y.~Bisk, D.~Fried, U.~Alon, and G.~Neubig, ``Webarena: {A} realistic web environment for building autonomous agents,'' in {\em The Twelfth International Conference on Learning Representations, {ICLR} 2024, Vienna, Austria, May 7-11, 2024}, OpenReview.net, 2024.

\bibitem{VisualWebArena}
J.~Y. Koh, R.~Lo, L.~Jang, V.~Duvvur, M.~C. Lim, P.~Huang, G.~Neubig, S.~Zhou, R.~Salakhutdinov, and D.~Fried, ``Visualwebarena: Evaluating multimodal agents on realistic visual web tasks,'' in {\em Proceedings of the 62nd Annual Meeting of the Association for Computational Linguistics (Volume 1: Long Papers), {ACL} 2024, Bangkok, Thailand, August 11-16, 2024} (L.~Ku, A.~Martins, and V.~Srikumar, eds.), pp.~881--905, Association for Computational Linguistics, 2024.

\bibitem{OSWorld}
T.~Xie, D.~Zhang, J.~Chen, X.~Li, S.~Zhao, R.~Cao, T.~J. Hua, Z.~Cheng, D.~Shin, F.~Lei, Y.~Liu, Y.~Xu, S.~Zhou, S.~Savarese, C.~Xiong, V.~Zhong, and T.~Yu, ``Osworld: Benchmarking multimodal agents for open-ended tasks in real computer environments,'' in {\em Advances in Neural Information Processing Systems 38: Annual Conference on Neural Information Processing Systems 2024, NeurIPS 2024, Vancouver, BC, Canada, December 10 - 15, 2024} (A.~Globersons, L.~Mackey, D.~Belgrave, A.~Fan, U.~Paquet, J.~M. Tomczak, and C.~Zhang, eds.), 2024.

\bibitem{WindowsAgentArena}
R.~Bonatti, D.~Zhao, F.~Bonacci, D.~Dupont, S.~Abdali, Y.~Li, Y.~Lu, J.~Wagle, K.~Koishida, A.~Bucker, L.~Jang, and Z.~Hui, ``Windows agent arena: Evaluating multi-modal {OS} agents at scale,'' {\em CoRR}, vol.~abs/2409.08264, 2024.

\bibitem{UI-Vision}
S.~Nayak, X.~Jian, K.~Q. Lin, J.~A. Rodriguez, M.~Kalsi, R.~Awal, N.~Chapados, M.~T. {\"{O}}zsu, A.~Agrawal, D.~V{\'{a}}zquez, C.~Pal, P.~Taslakian, S.~Gella, and S.~Rajeswar, ``Ui-vision: {A} desktop-centric {GUI} benchmark for visual perception and interaction,'' {\em CoRR}, vol.~abs/2503.15661, 2025.

\bibitem{SWE-bench}
C.~E. Jimenez, J.~Yang, A.~Wettig, S.~Yao, K.~Pei, O.~Press, and K.~R. Narasimhan, ``Swe-bench: Can language models resolve real-world github issues?,'' in {\em The Twelfth International Conference on Learning Representations, {ICLR} 2024, Vienna, Austria, May 7-11, 2024}, OpenReview.net, 2024.

\bibitem{DevBench}
B.~Li, W.~Wu, Z.~Tang, L.~Shi, J.~Yang, J.~Li, S.~Yao, C.~Qian, B.~Hui, Q.~Zhang, Z.~Yu, H.~Du, P.~Yang, D.~Lin, C.~Peng, and K.~Chen, ``Devbench: {A} comprehensive benchmark for software development,'' {\em CoRR}, vol.~abs/2403.08604, 2024.

\bibitem{APIBank}
M.~Li, Y.~Zhao, B.~Yu, F.~Song, H.~Li, H.~Yu, Z.~Li, F.~Huang, and Y.~Li, ``Api-bank: {A} comprehensive benchmark for tool-augmented llms,'' in {\em Proceedings of the 2023 Conference on Empirical Methods in Natural Language Processing, {EMNLP} 2023, Singapore, December 6-10, 2023} (H.~Bouamor, J.~Pino, and K.~Bali, eds.), pp.~3102--3116, Association for Computational Linguistics, 2023.

\bibitem{Toolbench}
Y.~Qin, S.~Liang, Y.~Ye, K.~Zhu, L.~Yan, Y.~Lu, Y.~Lin, X.~Cong, X.~Tang, B.~Qian, S.~Zhao, L.~Hong, R.~Tian, R.~Xie, J.~Zhou, M.~Gerstein, D.~Li, Z.~Liu, and M.~Sun, ``Toolllm: Facilitating large language models to master 16000+ real-world apis,'' in {\em The Twelfth International Conference on Learning Representations, {ICLR} 2024, Vienna, Austria, May 7-11, 2024}, OpenReview.net, 2024.

\bibitem{GAIA}
G.~Mialon, C.~Fourrier, T.~Wolf, Y.~LeCun, and T.~Scialom, ``{GAIA:} a benchmark for general {AI} assistants,'' in {\em The Twelfth International Conference on Learning Representations, {ICLR} 2024, Vienna, Austria, May 7-11, 2024}, OpenReview.net, 2024.

\bibitem{AppWorld}
H.~Trivedi, T.~Khot, M.~Hartmann, R.~Manku, V.~Dong, E.~Li, S.~Gupta, A.~Sabharwal, and N.~Balasubramanian, ``Appworld: {A} controllable world of apps and people for benchmarking interactive coding agents,'' in {\em Proceedings of the 62nd Annual Meeting of the Association for Computational Linguistics (Volume 1: Long Papers), {ACL} 2024, Bangkok, Thailand, August 11-16, 2024} (L.~Ku, A.~Martins, and V.~Srikumar, eds.), pp.~16022--16076, Association for Computational Linguistics, 2024.

\bibitem{tbench}
S.~Yao, N.~Shinn, P.~Razavi, and K.~Narasimhan, ``{\(\tau\)}-bench: {A} benchmark for tool-agent-user interaction in real-world domains,'' {\em CoRR}, vol.~abs/2406.12045, 2024.

\bibitem{PreferenceLeakage}
D.~Li, R.~Sun, Y.~Huang, M.~Zhong, B.~Jiang, J.~Han, X.~Zhang, W.~Wang, and H.~Liu, ``Preference leakage: A contamination problem in llm-as-a-judge,'' 2025.

\bibitem{grok4}
{xAI}, ``Grok 4.'' \url{https://x.ai/news/grok-4}, July 2025.
\newblock Accessed: 2025-07-28.

\bibitem{claude4}
{Anthropic}, ``Introducing claude 4.'' \url{https://www.anthropic.com/news/claude-4}, May 2025.
\newblock Accessed: 2025-07-28.

\bibitem{claude3.7}
{Anthropic}, ``Claude 3.7 sonnet and claude code.'' \url{https://www.anthropic.com/news/claude-3-7-sonnet}, Feb 2025.
\newblock Accessed: 2025-07-28.

\bibitem{gpt5}
{OpenAI}, ``Introducing gpt-5.'' \url{https://openai.com/index/introducing-gpt-5/}, August 2025.
\newblock Accessed: 2025-08-14.

\bibitem{o3-o4mini}
{OpenAI}, ``Introducing openai o3 and o4-mini.'' \url{https://openai.com/index/introducing-o3-and-o4-mini/}, April 2025.
\newblock Accessed: 2025-07-28.

\bibitem{GPT4-1}
{OpenAI}, ``Introducing gpt-4.1 in the api.'' \url{https://openai.com/index/gpt-4-1/}, April 2025.
\newblock Accessed: 2025-07-28.

\bibitem{GPT-OSS}
{OpenAI}, ``Introducing gpt-oss.'' \url{https://openai.com/index/introducing-gpt-oss/}, August 2025.
\newblock Accessed: 2025-08-14.

\bibitem{Gemini2-5}
G.~Comanici, E.~Bieber, M.~Schaekermann, I.~Pasupat, N.~Sachdeva, I.~Dhillon, M.~Blistein, O.~Ram, D.~Zhang, E.~Rosen, L.~Marris, S.~Petulla, C.~Gaffney, A.~Aharoni, N.~Lintz, T.~C. Pais, H.~Jacobsson, I.~Szpektor, N.-J. Jiang, K.~Haridasan, A.~Omran, N.~Saunshi, D.~Bahri, G.~Mishra, E.~Chu, T.~Boyd, B.~Hekman, A.~Parisi, C.~Zhang, K.~Kawintiranon, T.~Bedrax-Weiss, O.~Wang, Y.~Xu, O.~Purkiss, U.~Mendlovic, I.~Deutel, N.~Nguyen, A.~Langley, F.~Korn, L.~Rossazza, A.~Ramé, S.~Waghmare, H.~Miller, N.~Byrd, A.~Sheshan, R.~H.~S. Bhardwaj, P.~Janus, T.~Rissa, D.~Horgan, S.~Silver, A.~Wahid, S.~Brin, Y.~Raimond, K.~Kloboves, C.~Wang, N.~B. Gundavarapu, I.~Shumailov, B.~Wang, M.~Pajarskas, J.~Heyward, M.~Nikoltchev, M.~Kula, H.~Zhou, Z.~Garrett, S.~Kafle, S.~Arik, A.~Goel, M.~Yang, J.~Park, K.~Kojima, P.~Mahmoudieh, K.~Kavukcuoglu, G.~Chen, D.~Fritz, A.~Bulyenov, S.~Roy, D.~Paparas, H.~Shemtov, B.-J. Chen, R.~Strudel, D.~Reitter, A.~Roy, A.~Vlasov, C.~Ryu, C.~Leichner, H.~Yang, Z.~Mariet, D.~Vnukov, T.~Sohn,
  A.~Stuart, W.~Liang, M.~Chen, P.~Rawlani, C.~Koh, J.~Co-Reyes, G.~Lai, P.~Banzal, D.~Vytiniotis, J.~Mei, M.~Cai, M.~Badawi, C.~Fry, A.~Hartman, D.~Zheng, E.~Jia, J.~Keeling, A.~Louis, Y.~Chen, E.~Robles, W.-C. Hung, H.~Zhou, N.~Saxena, S.~Goenka, O.~Ma, Z.~Fisher, M.~H. Taege, E.~Graves, D.~Steiner, Y.~Li, S.~Nguyen, R.~Sukthankar, J.~Stanton, A.~Eslami, G.~Shen, B.~Akin, A.~Guseynov, Y.~Zhou, J.-B. Alayrac, A.~Joulin, E.~Farkash, A.~Thapliyal, S.~Roller, N.~Shazeer, T.~Davchev, T.~Koo, H.~Forbes-Pollard, K.~Audhkhasi, G.~Farquhar, A.~M. Gilady, M.~Song, J.~Aslanides, P.~Mendolicchio, A.~Parrish, J.~Blitzer, P.~Gupta, X.~Ju, X.~Yang, P.~Datta, A.~Tacchetti, S.~V. Mehta, G.~Dibb, S.~Gupta, F.~Piccinini, R.~Hadsell, S.~Rajayogam, J.~Jiang, P.~Griffin, P.~Sundberg, J.~Hayes, A.~Frolov, T.~Xie, A.~Zhang, K.~Dasgupta, U.~Kalra, L.~Shani, K.~Macherey, T.-K. Huang, L.~MacDermed, K.~Duddu, P.~Zacchello, Z.~Yang, J.~Lo, K.~Hui, M.~Kastelic, D.~Gasaway, Q.~Tan, S.~Yue, P.~Barrio, J.~Wieting, W.~Yang, A.~Nystrom,
  S.~Demmessie, A.~Levskaya, F.~Viola, C.~Tekur, G.~Billock, G.~Necula, M.~Joshi, R.~Schaeffer, S.~Lokhande, C.~Sorokin, P.~Shenoy, M.~Chen, M.~Collier, H.~Li, T.~Bos, N.~Wichers, S.~J. Lee, A.~Pouget, S.~Thangaraj, K.~Axiotis, P.~Crone, R.~Sterneck, N.~Chinaev, V.~Krakovna, O.~Ferludin, I.~Gemp, S.~Winkler, D.~Goldberg, I.~Korotkov, K.~Xiao, M.~Mehrotra, S.~Mariserla, V.~Piratla, T.~Thurk, K.~Pham, H.~Ma, A.~Senges, R.~Kumar, C.~Meyer, E.~Talius, N.~W. Pierse, B.~Sandhu, H.~Toma, K.~Lin, S.~Nath, T.~Stone, D.~Sadigh, N.~Gupta, A.~Guez, A.~Singh, M.~Thomas, T.~Duerig, Y.~Gong, R.~Tanburn, L.~L. Zhang, P.~Dao, M.~Hammad, S.~Xie, S.~Rijhwani, B.~Murdoch, D.~Kim, W.~Thompson, H.-T. Cheng, D.~Sohn, P.~Sprechmann, Q.~Xu, S.~Tadepalli, P.~Young, Y.~Zhang, H.~Srinivasan, M.~Aperghis, A.~Ayyar, H.~Fitoussi, R.~Burnell, D.~Madras, M.~Dusenberry, X.~Xiong, T.~Oguntebi, B.~Albrecht, J.~Bornschein, J.~Mitrović, M.~Dimarco, B.~K. Shamanna, P.~Shah, E.~Sezener, S.~Upadhyay, D.~Lacey, C.~Schiff, S.~Baur, S.~Ganapathy,
  E.~Schnider, M.~Wirth, C.~Schenck, A.~Simanovsky, Y.-X. Tan, P.~Fränken, D.~Duan, B.~Mankalale, N.~Dhawan, K.~Sequeira, Z.~Wei, S.~Goel, C.~Unlu, Y.~Zhu, H.~Sun, A.~Balashankar, K.~Shuster, M.~Umekar, M.~Alnahlawi, A.~van~den Oord, K.~Chen, Y.~Zhai, Z.~Dai, K.-H. Lee, E.~Doi, L.~Zilka, R.~Vallu, D.~Shrivastava, J.~Lee, H.~Husain, H.~Zhuang, V.~Cohen-Addad, J.~Barber, J.~Atwood, A.~Sadovsky, Q.~Wellens, S.~Hand, A.~Rajendran, A.~Turker, C.~Carey, Y.~Xu, H.~Soltau, Z.~Li, X.~Song, C.~Li, I.~Kemaev, S.~Brown, A.~Burns, V.~Patraucean, P.~Stanczyk, R.~Aravamudhan, M.~Blondel, H.~Noga, L.~Blanco, W.~Song, M.~Isard, M.~Sharma, R.~Hayes, D.~E. Badawy, A.~Lamp, I.~Laish, O.~Kozlova, K.~Chan, S.~Singla, S.~Sunkara, M.~Upadhyay, C.~Liu, A.~Bai, J.~Wilkiewicz, M.~Zlocha, J.~Liu, Z.~Li, H.~Li, O.~Barak, G.~Raboshchuk, J.~Choi, F.~Liu, E.~Jue, M.~Sharma, A.~Marzoca, R.~Busa-Fekete, A.~Korsun, A.~Elisseeff, Z.~Shen, S.~M. Carthy, K.~Lamerigts, A.~Hosseini, H.~Lin, C.~Chen, F.~Yang, K.~Chauhan, M.~Omernick, D.~Jia,
  K.~Zainullina, D.~Hassabis, D.~Vainstein, E.~Amid, X.~Zhou, R.~Votel, E.~Vértes, X.~Li, Z.~Zhou, A.~Lazaridou, B.~McMahan, A.~Narayanan, H.~Soyer, S.~Basu, K.~Lee, B.~Perozzi, Q.~Cao, L.~Berrada, R.~Arya, K.~Chen, Katrina, Xu, M.~Lochbrunner, A.~Hofer, S.~Sharifzadeh, R.~Wu, S.~Goldman, P.~Awasthi, X.~Wang, Y.~Wu, C.~Sha, B.~Zhang, M.~Mikuła, F.~Graziano, S.~Mcloughlin, I.~Giannoumis, Y.~Namiki, C.~Malik, C.~Radebaugh, J.~Hall, R.~Leal-Cavazos, J.~Chen, V.~Sindhwani, D.~Kao, D.~Greene, J.~Griffith, C.~Welty, C.~Montgomery, T.~Yoshino, L.~Yuan, N.~Goodman, A.~H. Michaely, K.~Lee, K.~Sawhney, W.~Chen, Z.~Zheng, M.~Shum, N.~Savinov, E.~Pot, A.~Pak, M.~Zadimoghaddam, S.~Bhatnagar, Y.~Lewenberg, B.~Kutzman, J.~Liu, L.~Katzen, J.~Selier, J.~Djolonga, D.~Lepikhin, K.~Xu, J.~Liang, J.~Tan, B.~Schillings, M.~Ersoy, P.~Blois, B.~Bandemer, A.~Singh, S.~Lebedev, P.~Joshi, A.~R. Brown, E.~Palmer, S.~Pathak, K.~Jalan, F.~Zubach, S.~Lall, R.~Parker, A.~Gunjan, S.~Rogulenko, S.~Sanghai, Z.~Leng, Z.~Egyed, S.~Li,
  M.~Ivanova, K.~Andriopoulos, J.~Xie, E.~Rosenfeld, A.~Wright, A.~Sharma, X.~Geng, Y.~Wang, S.~Kwei, R.~Pan, Y.~Zhang, G.~Wang, X.~Liu, C.~Yeung, E.~Cole, A.~Rosenberg, Z.~Yang, P.~Chen, G.~Polovets, P.~Nair, R.~Saxena, J.~Smith, S.~yiin Chang, A.~Mahendru, S.~Grant, A.~Iyer, I.~Cai, J.~McGiffin, J.~Shen, A.~Walton, A.~Girgis, O.~Woodman, R.~Ke, M.~Kwong, L.~Rouillard, J.~Rao, Z.~Li, Y.~Xu, F.~Prost, C.~Zou, Z.~Ji, A.~Magni, T.~Liechty, D.~A. Calian, D.~Ramachandran, I.~Krivokon, H.~Huang, T.~Chen, A.~Hauth, A.~Ilić, W.~Xi, H.~Lim, V.-D. Ion, P.~Moradi, M.~Toksoz-Exley, K.~Bullard, M.~Allamanis, X.~Yang, S.~Wang, Z.~Hong, A.~Gergely, C.~Li, B.~Mittal, V.~Kovalev, V.~Ungureanu, J.~Labanowski, J.~Wassenberg, N.~Lacasse, G.~Cideron, P.~Dević, A.~Marsden, L.~Nguyen, M.~Fink, Y.~Zhong, T.~Kiyono, D.~Ivanov, S.~Ma, M.~Bain, K.~Yalasangi, J.~She, A.~Petrushkina, M.~Lunayach, C.~Bromberg, S.~Hodkinson, V.~Meshram, D.~Vlasic, A.~Kyker, S.~Xu, J.~Stanway, Z.~Yang, K.~Zhao, M.~Tung, S.~Odoom, Y.~Fujii, J.~Gilmer,
  E.~Kim, F.~Halim, Q.~Le, B.~Bohnet, S.~El-Sayed, B.~Neyshabur, M.~Reynolds, D.~Reich, Y.~Xu, E.~Moreira, A.~Sharma, Z.~Liu, M.~J. Hosseini, N.~Raisinghani, Y.~Su, N.~Lao, D.~Formoso, M.~Gelmi, A.~Gueta, T.~Dey, E.~Gribovskaya, D.~Ćevid, S.~Mudgal, G.~Bingham, J.~Wang, A.~Kumar, A.~Cullum, F.~Han, K.~Bousmalis, D.~Cedillo, G.~Chu, V.~Magay, P.~Michel, E.~Hlavnova, D.~Calandriello, S.~Ariafar, K.~Yao, V.~Sehwag, A.~Vezer, A.~D. Lago, Z.~Zhu, P.~K. Rubenstein, A.~Porter, A.~Baddepudi, O.~Riva, M.~D. Istin, C.-K. Yeh, Z.~Li, A.~Howard, N.~Jha, J.~Chen, R.~de~Liedekerke, Z.~Ahmed, M.~Rodriguez, T.~Bhatia, B.~Wang, A.~Elqursh, D.~Klinghoffer, P.~Chen, P.~Kohli, T.~I, W.~Zhang, Z.~Nado, J.~Chen, M.~Chen, G.~Zhang, A.~Singh, A.~Hillier, F.~Lebron, Y.~Tao, T.~Liu, G.~Dulac-Arnold, J.~Zhang, S.~Narayan, B.~Liu, O.~Firat, A.~Bhowmick, B.~Liu, H.~Zhang, Z.~Zhang, G.~Rotival, N.~Howard, A.~Sinha, A.~Grushetsky, B.~Beyret, K.~Gopalakrishnan, J.~Zhao, K.~He, S.~Payrits, Z.~Nabulsi, Z.~Zhang, W.~Chen, E.~Lee, N.~Fallen,
  S.~Gollapudi, A.~Zhou, F.~Pavetić, T.~Köppe, S.~Huang, R.~Pasumarthi, N.~Fernando, F.~Fischer, D.~Ćurko, Y.~Gao, J.~Svensson, A.~Stone, H.~Qureshi, A.~Sinha, A.~Kulshreshtha, M.~Matysiak, J.~Mao, C.~Saroufim, A.~Faust, Q.~Duan, G.~Fidel, K.~Katircioglu, R.~L. Kaufman, D.~Shah, W.~Kong, A.~Bapna, G.~Weisz, E.~Dunleavy, P.~Dutta, T.~Liu, R.~Chaabouni, C.~Parada, M.~Wu, A.~Belias, A.~Bissacco, S.~Fort, L.~Xiao, F.~Huot, C.~Knutsen, Y.~Blau, G.~Li, J.~Prendki, J.~Love, Y.~Chow, P.~Charoenpanit, H.~Shimokawa, V.~Coriou, K.~Gregor, T.~Izo, A.~Akula, M.~Pinto, C.~Hahn, D.~Paulus, J.~Guo, N.~Sharma, C.-J. Hsieh, A.~Chukwuka, K.~Hashimoto, N.~Rauschmayr, L.~Wu, C.~Angermueller, Y.~Wang, S.~Gerlach, M.~Pliskin, D.~Mirylenka, M.~Ma, L.~Baugher, B.~Gale, S.~Bijwadia, N.~Rakićević, D.~Wood, J.~Park, C.-C. Chang, B.~Seal, C.~Tar, K.~Krasowiak, Y.~Song, G.~Stephanov, G.~Wang, M.~Maggioni, S.~X. Lin, F.~Wu, S.~Paul, Z.~Jiang, S.~Agrawal, B.~Piot, A.~Feng, C.~Kim, T.~Doshi, J.~Lai, Chuqiao, Xu, S.~Vikram, C.~Chelba,
  S.~Krause, V.~Zhuang, J.~Rae, T.~Denk, A.~Collister, L.~Weerts, X.~Luo, Y.~Lu, H.~Garnes, N.~Gupta, T.~Spitz, A.~Hassidim, L.~Liang, I.~Shafran, P.~Humphreys, K.~Vassigh, P.~Wallis, V.~Shejwalkar, N.~Perez-Nieves, R.~Hornung, M.~Tan, B.~Westberg, A.~Ly, R.~Zhang, B.~Farris, J.~Park, A.~Kosik, Z.~Cankara, A.~Maksai, Y.~Xu, A.~Cassirer, S.~Caelles, A.~Abdolmaleki, M.~Chiang, A.~Fabrikant, S.~Shetty, L.~He, M.~Giménez, H.~Hashemi, S.~Panthaplackel, Y.~Kulizhskaya, S.~Deshmukh, D.~Pighin, R.~Alazard, D.~Jindal, S.~Noury, P.~K. S, S.~Qin, X.~Dotiwalla, S.~Spencer, M.~Babaeizadeh, B.~J. Chen, V.~Mehta, J.~Lees, A.~Leach, P.~Koanantakool, I.~Akolzin, R.~Comanescu, J.~Ahn, A.~Svyatkovskiy, B.~Mustafa, D.~D'Ambrosio, S.~M.~R. Garlapati, P.~Lamblin, A.~Agarwal, S.~Song, P.~G. Sessa, P.~Coquinot, J.~Maggs, H.~Masoom, D.~Pitta, Y.~Wang, P.~Morris-Suzuki, B.~Porter, J.~Jia, J.~Dudek, R.~R, C.~Paduraru, A.~Ansell, T.~Bolukbasi, T.~Lu, R.~Ganeshan, Z.~Wang, H.~Griffiths, R.~Benenson, Y.~He, J.~Swirhun, G.~Papamakarios,
  A.~Chawla, K.~Sengupta, Y.~Wang, V.~Milutinovic, I.~Mordatch, Z.~Jia, J.~Smith, W.~Ng, S.~Nigam, M.~Young, E.~Vušak, B.~Hechtman, S.~Goenka, A.~Zipori, K.~Ayoub, A.~Popat, T.~Acharya, L.~Yu, D.~Bloxwich, H.~Song, P.~Roit, H.~Li, A.~Boag, N.~Nayakanti, B.~Chandra, T.~Ding, A.~Mehta, C.~Hope, J.~Zhang, I.~H. Shtacher, K.~Badola, R.~Nakashima, A.~Sozanschi, I.~Comşa, A.~Žužul, E.~Caveness, J.~Odell, M.~Watson, D.~de~Cesare, P.~Lippe, D.~Lockhart, S.~Verma, H.~Chen, S.~Sun, L.~Zhuo, A.~Shah, P.~Gupta, A.~Muzio, N.~Niu, A.~Zait, A.~Singh, M.~Gaba, F.~Ye, P.~Ramachandran, M.~Saleh, R.~A. Popa, A.~Dubey, F.~Liu, S.~Javanmardi, M.~Epstein, R.~Hemsley, R.~Green, N.~Ranka, E.~Cohen, C.~K. Fu, S.~Ghemawat, J.~Borovik, J.~Martens, A.~Chen, P.~Shyam, A.~S. Pinto, M.-H. Yang, A.~Ţifrea, D.~Du, B.~Gong, A.~Agarwal, S.~Kim, C.~Frank, S.~Shah, X.~Song, Z.~Deng, A.~Mikhalap, K.~Chatziprimou, T.~Chung, T.~Creswell, S.~Zhang, Y.~Jun, C.~Lebsack, W.~Truong, S.~Andačić, I.~Yona, M.~Fornoni, R.~Rong, S.~Toropov, A.~S.
  Soudagar, A.~Audibert, S.~Zaiem, Z.~Abbas, A.~Rusu, S.~Potluri, S.~Weng, A.~Kementsietsidis, A.~Tsitsulin, D.~Peng, N.~Ha, S.~Jain, T.~Latkar, S.~Ivanov, C.~McLean, A.~GP, R.~Venkataraman, C.~Liu, D.~Krishnan, J.~D'sa, R.~Yogev, P.~Collins, B.~Lee, L.~Ho, C.~Doersch, G.~Yona, S.~Gao, F.~T. Ferreira, A.~Ozturel, H.~Muckenhirn, C.~Zheng, G.~Balasubramaniam, M.~Bansal, G.~van~den Driessche, S.~Eiger, S.~Haykal, V.~Misra, A.~Goyal, D.~Martins, G.~Leung, J.~Valfridsson, F.~Flynn, W.~Bishop, C.~Pang, Y.~Halpern, H.~Yu, L.~Moore, Yuvein, Zhu, S.~Thiagarajan, Y.~Drori, Z.~Xiao, L.~Dery, R.~Jagerman, J.~Lu, E.~Ge, V.~Aggarwal, A.~Khare, V.~Tran, O.~Elyada, F.~Alet, J.~Rubin, I.~Chou, D.~Tian, L.~Bai, L.~Chan, L.~Lew, K.~Misiunas, T.~Bilal, A.~Ray, S.~Raghuram, A.~Castro-Ros, V.~Carpenter, C.~Zheng, M.~Kilgore, J.~Broder, E.~Xue, P.~Kallakuri, D.~Dua, N.~Yuen, S.~Chien, J.~Schultz, S.~Agrawal, R.~Tsarfaty, J.~Hu, A.~Kannan, D.~Marcus, N.~Kothari, B.~Sun, B.~Horn, M.~Bošnjak, F.~Naeem, D.~Hirsch, L.~Chiang, B.~Fang,
  J.~Han, Q.~Wang, B.~Hora, A.~He, M.~Lučić, B.~Changpinyo, A.~Tripathi, J.~Youssef, C.~Kwak, P.~Schlattner, C.~Graves, R.~Leblond, W.~Zeng, A.~Andreassen, G.~Rasskin, Y.~Song, E.~Cao, J.~Oh, M.~Hoffman, W.~Skut, Y.~Zhang, J.~Stritar, X.~Cai, S.~Khanna, K.~Wang, S.~Sharma, C.~Reisswig, Y.~Jun, A.~Prasad, T.~Sholokhova, P.~Singh, A.~G. Rosenthal, A.~Ruoss, F.~Beaufays, S.~Kirmani, D.~Chen, J.~Schalkwyk, J.~Herzig, B.~Kim, J.~Jacob, D.~Vincent, A.~N. Reyes, I.~Balazevic, L.~Hussenot, J.~Schneider, P.~Barnes, L.~Castro, S.~R. Babbula, S.~Green, S.~Cabi, N.~Duduta, D.~Driess, R.~Galt, N.~Velan, J.~Wang, H.~Jiao, M.~Mauger, D.~Phan, M.~Patel, V.~Galić, J.~Chang, E.~Marcus, M.~Harvey, J.~Salazar, E.~Dabir, S.~S. Sheth, A.~Mandhane, H.~Sedghi, J.~Willcock, A.~Zandieh, S.~Prabhakara, A.~Amini, A.~Miech, V.~Stone, M.~Nicosia, P.~Niemczyk, Y.~Xiao, L.~Kim, S.~Kwasiborski, V.~Verma, A.~M. Oflazer, C.~Hirnschall, P.~Sung, L.~Liu, R.~Everett, M.~Bakker, Ágoston Weisz, Y.~Wang, V.~Sampathkumar, U.~Shaham, B.~Xu,
  Y.~Altun, M.~Wang, T.~Saeki, G.~Chen, E.~Taropa, S.~Vasanth, S.~Austin, L.~Huang, G.~Petrovic, Q.~Dou, D.~Golovin, G.~Rozhdestvenskiy, A.~Culp, W.~Wu, M.~Sano, D.~Jain, J.~Proskurnia, S.~Cevey, A.~C. Ruiz, P.~Patil, M.~Mirzazadeh, E.~Ni, J.~Snaider, L.~Fan, A.~Fréchette, A.~Pierigiovanni, S.~Iqbal, K.~Lee, C.~Fantacci, J.~Xing, L.~Wang, A.~Irpan, D.~Raposo, Y.~Luan, Z.~Chen, H.~Ganapathy, K.~Hui, J.~Nie, I.~Guyon, H.~Ge, R.~Vij, H.~Zheng, D.~Lee, A.~Castaño, K.~Baatarsukh, G.~Ibagon, A.~Chronopoulou, N.~FitzGerald, S.~Viswanadha, S.~Huda, R.~Moroshko, G.~Stoyanov, P.~Kolhar, A.~Vaucher, I.~Watts, A.~Kuncoro, H.~Michalewski, S.~Kambala, B.-O. Batsaikhan, A.~Andreev, I.~Jurenka, M.~Le, Q.~Chen, W.~A. Jishi, S.~Chakera, Z.~Chen, A.~Kini, V.~Yadav, A.~Siddhant, I.~Labzovsky, B.~Lakshminarayanan, C.~G. Bostock, P.~Botadra, A.~Anand, C.~Bishop, S.~Conway-Rahman, M.~Agarwal, Y.~Donchev, A.~Singhal, F.~de~Chaumont~Quitry, N.~Ponomareva, N.~Agrawal, B.~Ni, K.~Krishna, M.~Samsikova, J.~Karro, Y.~Du, T.~von Glehn,
  C.~Lu, C.~A. Choquette-Choo, Z.~Qin, T.~Zhang, S.~Li, D.~Tyam, S.~Mishra, W.~Lowe, C.~Ji, W.~Wang, M.~Faruqui, A.~Slone, V.~Dalibard, A.~Narayanaswamy, J.~Lambert, P.-A. Manzagol, D.~Karliner, A.~Bolt, I.~Lobov, A.~Kusupati, C.~Ye, X.~Yang, H.~Zen, N.~George, M.~Bhutani, O.~Lacombe, R.~Riachi, G.~Bansal, R.~Soh, Y.~Gao, Y.~Yu, A.~Yu, E.~Nottage, T.~Rojas-Esponda, J.~Noraky, M.~Gupta, R.~Kotikalapudi, J.~Chang, S.~Deur, D.~Graur, A.~Mossin, E.~Farnese, R.~Figueira, A.~Moufarek, A.~Huang, P.~Zochbauer, B.~Ingram, T.~Chen, Z.~Wu, A.~Puigdomènech, L.~Rechis, D.~Yu, S.~G.~S. Padmanabhan, R.~Zhu, C.~ling Ko, A.~Banino, S.~Daruki, A.~Selvan, D.~Bhaswar, D.~H. Diaz, C.~Su, S.~Scellato, J.~Brennan, W.~Han, G.~Chung, P.~Agrawal, U.~Khandelwal, K.~C. Sim, M.~Lustman, S.~Ritter, K.~Guu, J.~Xia, P.~Jain, E.~Wang, T.~Hill, M.~Rossini, M.~Kostelac, T.~Misiunas, A.~Sabne, K.~Kim, A.~Iscen, C.~Wang, J.~Leal, A.~Sreevatsa, U.~Evci, M.~Warmuth, S.~Joshi, D.~Suo, J.~Lottes, G.~Honke, B.~Jou, S.~Karp, J.~Hu, H.~Sahni, A.~A.
  Taïga, W.~Kong, S.~Ghosh, R.~Wang, J.~Pavagadhi, N.~Axelsson, N.~Grigorev, P.~Siegler, R.~Lin, G.~Wang, E.~Parisotto, S.~Maddineni, K.~Subudhi, E.~Ben-David, E.~Pochernina, O.~Keller, T.~Avrahami, Z.~Yuan, P.~Mehta, J.~Liu, S.~Yang, W.~Kan, K.~Lee, T.~Funkhouser, D.~Cheng, H.~Shi, A.~Sharma, J.~Kelley, M.~Eyal, Y.~Malkov, C.~Tallec, Y.~Bahat, S.~Yan, Xintian, Wu, D.~Lindner, C.~Wu, A.~Caciularu, X.~Luo, R.~Jenatton, T.~Zaman, Y.~Bi, I.~Kornakov, G.~Mallya, D.~Ikeda, I.~Karo, A.~Singh, C.~Evans, P.~Netrapalli, V.~Nallatamby, I.~Tian, Y.~Assael, V.~Raunak, V.~Carbune, I.~Bica, L.~Madmoni, D.~Cattle, S.~Grover, K.~Somandepalli, S.~Lall, A.~Vázquez-Reina, R.~Patana, J.~Mu, P.~Talluri, M.~Tran, R.~Aggarwal, R.~Skerry-Ryan, J.~Xu, M.~Burrows, X.~Pan, E.~Yvinec, D.~Lu, Z.~Zhang, D.~D. Nguyen, H.~Mu, G.~Barcik, H.~Ran, L.~Beltrone, K.~Choromanski, D.~Kharrat, S.~Albanie, S.~Purser-haskell, D.~Bieber, C.~Zhang, J.~Wang, T.~Hudson, Z.~Zhang, H.~Fu, J.~Mauerer, M.~H. Bateni, A.~Maschinot, B.~Wang, M.~Zhu, A.~Pillai,
  T.~Weyand, S.~Liu, O.~Akerlund, F.~Bertsch, V.~Premachandran, A.~Jin, V.~Roulet, P.~de~Boursac, S.~Mittal, N.~Ndebele, G.~Karadzhov, S.~Ghalebikesabi, R.~Liang, A.~Wu, Y.~Cong, N.~Ghelani, S.~Singh, B.~Fatemi, Warren, Chen, C.~Kwong, A.~Kolganov, S.~Li, R.~Song, C.~Kuang, S.~Miryoosefi, D.~Webster, J.~Wendt, A.~Socala, G.~Su, A.~Mendonça, A.~Gupta, X.~Li, T.~Tsai, Qiong, Hu, K.~Kang, A.~Chen, S.~Girgin, Y.~Xian, A.~Lee, N.~Ramsden, L.~Baker, M.~C. Elish, V.~Krayvanova, R.~Joshi, J.~Simsa, Y.-Y. Yang, P.~Ambroszczyk, D.~Ghosh, A.~Kar, Y.~Shangguan, Y.~Yamamori, Y.~Akulov, A.~Brock, H.~Tang, S.~Vashishtha, R.~Munoz, A.~Steiner, K.~Andra, D.~Eppens, Q.~Feng, H.~Kobayashi, S.~Goldshtein, M.~E. Mahdy, X.~Wang, Jilei, Wang, R.~Killam, T.~Kwiatkowski, K.~Kopparapu, S.~Zhan, C.~Jia, A.~Bendebury, S.~Luo, A.~Recasens, T.~Knight, J.~Chen, M.~Patel, Y.~Li, B.~Withbroe, D.~Weesner, K.~Bhatia, J.~Ren, D.~Eisenbud, E.~Songhori, Y.~Sun, T.~Choma, T.~Kementsietsidis, L.~Manning, B.~Roark, W.~Farhan, J.~Feng, S.~Tatineni,
  J.~Cobon-Kerr, Y.~Li, L.~A. Hendricks, I.~Noble, C.~Breaux, N.~Kushman, L.~Peng, F.~Xue, T.~Tobin, J.~Rogers, J.~Lipschultz, C.~Alberti, A.~Vlaskin, M.~Dehghani, R.~Sharma, T.~Warkentin, C.-Y. Lee, B.~Uria, D.-C. Juan, A.~Chandorkar, H.~Sheftel, R.~Liu, E.~Davoodi, B.~D.~B. Pigem, K.~Dhamdhere, D.~Ross, J.~Hoech, M.~Mahdieh, L.~Liu, Q.~Li, L.~McCafferty, C.~Liu, M.~Mircea, Y.~Song, O.~Savant, A.~Saade, C.~Cherry, V.~Hellendoorn, S.~Goyal, P.~Pucciarelli, D.~V. Torres, Z.~Yahav, H.~Lee, L.~L. Sjoesund, C.~Kirov, B.~Chang, D.~Ghoshal, L.~Li, G.~Baechler, S.~Pereira, T.~Sainath, A.~Boral, D.~Grewe, A.~Halumi, N.~M. Phu, T.~Shen, M.~T. Ribeiro, D.~Varma, A.~Kaskasoli, V.~Feinberg, N.~Potti, J.~Kahn, M.~Wisniewski, S.~Mohamed, A.~M. Hrafnkelsson, B.~Shahriari, J.-B. Lespiau, L.~Patel, L.~Yeung, T.~Paine, L.~Mei, A.~Ramirez, R.~Shivanna, L.~Zhong, J.~Woodward, G.~Tubone, S.~Khan, H.~Chen, E.~Nielsen, C.~Ionescu, U.~Prabhu, M.~Gao, Q.~Wang, S.~Augenstein, N.~Subramaniam, J.~Chang, F.~Iliopoulos, J.~Luo, M.~Khan,
  W.~Kuo, D.~Teplyashin, F.~Perot, L.~Kilpatrick, A.~Globerson, H.~Yu, A.~Siddiqui, N.~Sukhanov, A.~Kandoor, U.~Gupta, M.~Andreetto, M.~Ambar, D.~Kim, P.~Wesołowski, S.~Perrin, B.~Limonchik, W.~Fan, J.~Stephan, I.~Stewart-Binks, R.~Kappedal, T.~He, S.~Cogan, R.~Datta, T.~Zhou, J.~Ye, L.~Kieliger, A.~Ramalho, K.~Kastner, F.~Mentzer, W.-J. Ko, A.~Suggala, T.~Zhou, S.~Butt, H.~Strejček, L.~Belenki, S.~Venugopalan, M.~Ling, E.~Eltyshev, Y.~Deng, G.~Kovacs, M.~Raghavachari, H.~Dai, T.~Schuster, S.~Schwarcz, R.~Nguyen, A.~Nguyen, G.~Buttimore, S.~B. Mallick, S.~Gandhe, S.~Benjamin, M.~Jastrzebski, L.~Yan, S.~Basu, C.~Apps, I.~Edkins, J.~Allingham, I.~Odisho, T.~Kocisky, J.~Zhao, L.~Xue, A.~Reddy, C.~Anastasiou, A.~Atias, S.~Redmond, K.~Milan, N.~Heess, H.~Schmit, A.~Dafoe, D.~Andor, T.~Gangwani, A.~Dragan, S.~Zhang, A.~Kachra, G.~Wu, S.~Xue, K.~Aydin, S.~Liu, Y.~Zhou, M.~Malihi, A.~Wu, S.~Gopal, C.~Schumann, P.~Stys, A.~Wang, M.~Olšák, D.~Liu, C.~Schallhart, Y.~Mao, D.~Brady, H.~Xu, T.~Mery, C.~Sitawarin,
  S.~Velusamy, T.~Cobley, A.~Zhai, C.~Walder, N.~Katz, G.~Jawahar, C.~Kulkarni, A.~Yang, A.~Paszke, Y.~Wang, B.~Damoc, Z.~Borsos, R.~Smith, J.~Li, M.~Gupta, A.~Kapishnikov, S.~Prakash, F.~Luisier, R.~Agarwal, W.~Grathwohl, K.~Chen, K.~Han, N.~Mehta, A.~Over, S.~Azizi, L.~Meng, N.~D. Santo, K.~Zheng, J.~Shapiro, I.~Petrovski, J.~Hui, A.~Ghafouri, J.~Snoek, J.~Qin, M.~Jordan, C.~Sikora, J.~Malmaud, Y.~Kuang, A.~Świetlik, R.~Sang, C.~Shi, L.~Li, A.~Rosenberg, S.~Zhao, A.~Crawford, J.-T. Peter, Y.~Lei, X.~Garcia, L.~Le, T.~Wang, J.~Amelot, D.~Orr, P.~Kacham, D.~Alon, G.~Tyen, A.~Arora, J.~Lyon, A.~Kurakin, M.~Ly, T.~Guidroz, Z.~Yan, R.~Panigrahy, P.~Xu, T.~Kagohara, Y.~Cheng, E.~Noland, J.~Lee, J.~Lee, C.~Yip, M.~Wang, E.~Nehoran, A.~Bykovsky, Z.~Shan, A.~Bhagatwala, C.~Yan, J.~Tan, G.~Garrido, D.~Ethier, N.~Hurley, G.~Vesom, X.~Chen, S.~Qiao, A.~Nayyar, J.~Walker, P.~Sandhu, M.~Rosca, D.~Swisher, M.~Dektiarev, J.~Dillon, G.-C. Muraru, M.~Tragut, A.~Myaskovsky, D.~Reid, M.~Velic, O.~Xiao, J.~George, M.~Brand,
  J.~Li, W.~Yu, S.~Gu, X.~Deng, F.-X. Aubet, S.~H. Yeganeh, F.~Alcober, C.~Smith, T.~Cohn, K.~McKinney, M.~Tschannen, R.~Sampath, G.~Cheon, L.~Luo, L.~Liu, J.~Orbay, H.~Peng, G.~Botea, X.~Zhang, C.~Yoon, C.~Magalhaes, P.~Stradomski, I.~Mackinnon, S.~Hemingray, K.~Venkatesan, R.~May, J.~Kim, A.~Druinsky, J.~Ye, Z.~Xu, T.~Huang, J.~A. Abdallah, A.~Dostmohamed, R.~Fellinger, T.~Munkhdalai, A.~Maurya, P.~Garst, Y.~Zhang, M.~Krikun, S.~Bucher, A.~S. Veerubhotla, Y.~Liu, S.~Li, N.~Gupta, J.~Adamek, H.~Chen, B.~Orlando, A.~Zaks, J.~van Amersfoort, J.~Camp, H.~Wan, H.~Choe, Z.~Wu, K.~Olszewska, W.~Yu, A.~Vadali, M.~Scholz, D.~D. Freitas, J.~Lin, A.~Hua, X.~Liu, F.~Ding, Y.~Zhou, B.~Severson, K.~Tsihlas, S.~Yang, T.~Spalink, V.~Yerram, H.~Pankov, R.~Blevins, B.~Vargas, S.~Jauhari, M.~Miecnikowski, M.~Zhang, S.~Kumar, C.~Farabet, C.~L. Lan, S.~Flennerhag, Y.~Bitton, A.~Ma, A.~Bražinskas, E.~Collins, N.~Ahuja, S.~Kudugunta, A.~Bortsova, M.~Giang, W.~Zhu, E.~Chi, S.~Lundberg, A.~Stern, S.~Puttagunta, J.~Xiong, X.~Wu,
  Y.~Pande, A.~Jhindal, D.~Murphy, J.~Clark, M.~Brockschmidt, M.~Deines, K.~R. McKee, D.~Bahir, J.~Shen, M.~Truong, D.~McDuff, A.~Gesmundo, E.~Rosseel, B.~Liang, K.~Caluwaerts, J.~Hamrick, J.~Kready, M.~Cassin, R.~Ingale, L.~Lao, S.~Pollom, Y.~Ding, W.~He, L.~Bellot, J.~Iljazi, R.~S. Boppana, S.~Han, T.~Thompson, A.~Khalifa, A.~Bulanova, B.~Mitrevski, B.~Pang, E.~Cooney, T.~Shi, R.~Coaguila, T.~Yakar, M.~Ranzato, N.~Momchev, C.~Rawles, Z.~Charles, Y.~Maeng, Y.~Zhang, R.~Bansal, X.~Zhao, B.~Albert, Y.~Yuan, S.~Vijayanarasimhan, R.~Hirsch, V.~Ramasesh, K.~Vodrahalli, X.~Wang, A.~Gupta, D.~Strouse, J.~Ni, R.~Patel, G.~Taubman, Z.~Huo, D.~Gharibian, M.~Monteiro, H.~Lam, S.~Vasudevan, A.~Chaudhary, I.~Albuquerque, K.~Gupta, S.~Riedel, C.~Hegde, A.~Ruderman, A.~György, M.~Wainwright, A.~Chaugule, B.~K. Ayan, T.~Levinboim, S.~Shleifer, Y.~Kalley, V.~Mirrokni, A.~Rao, P.~Radhakrishnan, J.~Hartford, J.~Wu, Z.~Zhu, F.~Bertolini, H.~Xiong, N.~Serrano, H.~Tomlinson, M.~Ott, Y.~Chang, M.~Graham, J.~Li, M.~Liang, X.~Long,
  S.~Borgeaud, Y.~Ahmad, A.~Grills, D.~Mincu, M.~Izzard, Y.~Liu, J.~Xie, L.~O'Bryan, S.~Ponda, S.~Tong, M.~Liu, D.~Malkin, K.~Salama, Y.~Chen, R.~Anil, A.~Rao, R.~Swavely, M.~Bilenko, N.~Anderson, T.~Tan, J.~Xie, X.~Wu, L.~Yu, O.~Vinyals, A.~Ryabtsev, R.~Dangovski, K.~Baumli, D.~Keysers, C.~Wright, Z.~Ashwood, B.~Chan, A.~Shtefan, Y.~Guo, A.~Bapna, R.~Soricut, S.~Pecht, S.~Ramos, R.~Wang, J.~Cai, T.~Trinh, P.~Barham, L.~Friso, E.~Stickgold, X.~Ding, S.~Shakeri, D.~Ardila, E.~Briakou, P.~Culliton, A.~Raveret, J.~Cui, D.~Saxton, S.~Roy, J.~Azizi, P.~Yin, L.~Loher, A.~Bunner, M.~Choi, F.~Ahmed, E.~Li, Y.~Li, S.~Dai, M.~Elabd, S.~Ganapathy, S.~Agrawal, Y.~Hua, P.~Kunkle, S.~Rajayogam, A.~Ahuja, A.~Conmy, A.~Vasiloff, P.~Beak, C.~Yew, J.~Mudigonda, B.~Wydrowski, J.~Blanton, Z.~Wang, Y.~Dauphin, Z.~Xu, M.~Polacek, X.~Chen, H.~Hu, P.~Sho, M.~Kunesch, M.~H. Manshadi, E.~Rutherford, B.~Li, S.~Hsiao, I.~Barr, A.~Tudor, M.~Kecman, A.~Nagrani, V.~Pchelin, M.~Sundermeyer, A.~P. S, A.~Karmarkar, Y.~Gao, G.~Chole,
  O.~Bachem, I.~Gao, A.~BC, M.~Dibb, M.~Verzetti, F.~Hernandez-Campos, Y.~Lunts, M.~Johnson, J.~D. Trapani, R.~Koster, I.~Brusilovsky, B.~Xiong, M.~Mohabey, H.~Ke, J.~Zou, T.~Sabolić, V.~Campos, J.~Palowitch, A.~Morris, L.~Qiu, P.~Ponnuramu, F.~Li, V.~Sharma, K.~Sodhia, K.~Tekelioglu, A.~Chuklin, M.~Yenugula, E.~Gemzer, T.~Strinopoulos, S.~El-Husseini, H.~Wang, Y.~Zhong, E.~Leurent, P.~Natsev, W.~Wang, D.~Mahaarachchi, T.~Zhu, S.~Peng, S.~Alabed, C.-C. Lee, A.~Brohan, A.~Szlam, G.~Oh, A.~Kovsharov, J.~Lee, R.~Wong, M.~Barnes, G.~Thornton, F.~Gimeno, O.~Levy, M.~Sevenich, M.~Johnson, J.~Mallinson, R.~Dadashi, Z.~Wang, Q.~Ren, P.~Lahoti, A.~Dhar, J.~Feldman, D.~Zheng, T.~Ulrich, L.~Panait, M.~Blokzijl, C.~Baetu, J.~Matak, J.~Harlalka, M.~Shah, T.~Marian, D.~von Dincklage, C.~Du, R.~Ley-Wild, B.~Brownfield, M.~Schumacher, Y.~Stuken, S.~Noghabi, S.~Gupta, X.~Ren, E.~Malmi, F.~Weissenberger, B.~Huergo, M.~Bauza, T.~Lampe, A.~Douillard, M.~Seyedhosseini, R.~Frostig, Z.~Ghahramani, K.~Nguyen, K.~Krishnakumar,
  C.~Ye, R.~Gupta, A.~Nazari, R.~Geirhos, P.~Shaw, A.~Eleryan, D.~Damen, J.~Palomaki, T.~Xiao, Q.~Wu, Q.~Yuan, P.~Meadowlark, M.~Bilotti, R.~Lin, M.~Sridhar, Y.~Schroecker, D.-W. Chung, J.~Luo, T.~Strohman, T.~Liu, A.~Zheng, J.~Emond, W.~Wang, A.~Lampinen, T.~Fukuzawa, F.~Campbell-Ajala, M.~Roy, J.~Lee-Thorp, L.~Wang, I.~Naim, Tony, N.~ên, G.~Bensky, A.~Gupta, D.~Rogozińska, J.~Fu, T.~S. Pillai, P.~Veličković, S.~Drath, P.~Neubeck, V.~Tulsyan, A.~Klimovskiy, D.~Metzler, S.~Stevens, A.~Yeh, J.~Yuan, T.~Yu, K.~Zhang, A.~Go, V.~Tsang, Y.~Xu, A.~Wan, I.~Galatzer-Levy, S.~Sobell, A.~Toki, E.~Salesky, W.~Zhou, D.~Antognini, S.~Douglas, S.~Wu, A.~Lelkes, F.~Kim, P.~Cavallaro, A.~Salazar, Y.~Liu, J.~Besley, T.~Refice, Y.~Jia, Z.~Li, M.~Sokolik, A.~Kannan, J.~Simon, J.~Chick, A.~Aharon, M.~Gandhi, M.~Daswani, K.~Amiri, V.~Birodkar, A.~Ittycheriah, P.~Grabowski, O.~Chang, C.~Sutton, Zhixin, Lai, U.~Telang, S.~Sargsyan, T.~Jiang, R.~Hoffmann, N.~Brichtova, M.~Hessel, J.~Halcrow, S.~Jerome, G.~Brown, A.~Tomala,
  E.~Buchatskaya, D.~Yu, S.~Menon, P.~Moreno, Y.~Liao, V.~Zayats, L.~Tang, S.~Mah, A.~Shenoy, A.~Siegman, M.~Hadian, O.~Kwon, T.~Tu, N.~Khajehnouri, R.~Foley, P.~Haghani, Z.~Wu, V.~Keshava, K.~Gupta, T.~Bruguier, R.~Yao, D.~Karmon, L.~Zintgraf, Z.~Wang, E.~Piqueras, J.~Jung, J.~Brennan, D.~Machado, M.~Giustina, M.~Tessler, K.~Lee, Q.~Zhang, J.~Moore, K.~Daugaard, A.~Frömmgen, J.~Beattie, F.~Zhang, D.~Kasenberg, T.~Geri, D.~Qin, G.~S. Tomar, T.~Ouyang, T.~Yu, L.~Zhou, R.~Mathews, A.~Davis, Y.~Li, J.~Gupta, D.~Yates, L.~Deng, E.~Kemp, G.-Y. Joung, S.~Vassilvitskii, M.~Guo, P.~LV, D.~Dopson, S.~Lachgar, L.~McConnaughey, H.~Choudhury, D.~Dena, A.~Cohen, J.~Ainslie, S.~Levi, P.~Gopavarapu, P.~Zablotskaia, H.~Vallet, S.~Bahargam, X.~Tang, N.~Tomasev, E.~Dyer, D.~Balle, H.~Lee, W.~Bono, J.~G. Mendez, V.~Zubov, S.~Yang, I.~Rendulic, Y.~Zheng, A.~Hogue, G.~Pundak, R.~Leith, A.~Bhoopchand, M.~Han, M.~Žanić, T.~Schaul, M.~Delakis, T.~Iyer, G.~Wang, H.~Singh, A.~Abdelhamed, T.~Thomas, S.~Brahma, H.~Dib, N.~Kumar,
  W.~Zhou, L.~Bai, P.~Mishra, J.~Sun, V.~Anklin, R.~Sukkerd, L.~Agubuzu, A.~Briukhov, A.~Gulati, M.~Sieb, F.~Pardo, S.~Nasso, J.~Chen, K.~Zhu, T.~Sosea, A.~Goldin, K.~Rush, S.~A. Hombaiah, A.~Noever, A.~Zhou, S.~Haves, M.~Phuong, J.~Ades, Y.~ting Chen, L.~Yang, J.~Pagadora, S.~Bileschi, V.~Cotruta, R.~Saputro, A.~Pramanik, S.~Ammirati, D.~Garrette, K.~Villela, T.~Blyth, C.~Akbulut, N.~Jha, A.~Rrustemi, A.~Wongpanich, C.~Nagpal, Y.~Wu, M.~Rivière, S.~Kishchenko, P.~Srinivasan, A.~Chen, A.~Sinha, T.~Pham, B.~Jia, T.~Hennigan, A.~Bakalov, N.~Attaluri, D.~Garmon, D.~Rodriguez, D.~Wegner, W.~Jia, E.~Senter, N.~Fiedel, D.~Petek, Y.~Liu, C.~Hardin, H.~T. Lehri, J.~Carreira, S.~Smoot, M.~Prasetya, N.~Akazawa, A.~Stefanoiu, C.-H. Ho, A.~Angelova, K.~Lin, M.~Kim, C.~Chen, M.~Sieniek, A.~Li, T.~Guo, S.~Baltateanu, P.~Tafti, M.~Wunder, N.~Olmert, D.~Shukla, J.~Shen, N.~Kovelamudi, B.~Venkatraman, S.~Neel, R.~Thoppilan, J.~Connor, F.~Benzing, A.~Stjerngren, G.~Ghiasi, A.~Polozov, J.~Howland, T.~Weber, J.~Chiu, G.~P.
  Girirajan, A.~Terzis, P.~Wang, F.~Li, Y.~B. Shalom, D.~Tewari, M.~Denton, R.~Aharoni, N.~Kalb, H.~Zhao, J.~Zhang, A.~Filos, M.~Rahtz, L.~Jain, C.~Fan, V.~Rodrigues, R.~Wang, R.~Shin, J.~Austin, R.~Ring, M.~Sanchez-Vargas, M.~Hassen, I.~Kessler, U.~Alon, G.~Zhang, W.~Chen, Y.~Ma, X.~Si, L.~Hou, A.~Mirhoseini, M.~Wilson, G.~Bacon, B.~Roelofs, L.~Shu, G.~Vasudevan, J.~Adler, A.~Dwornik, T.~Terzi, M.~Lawlor, H.~Askham, M.~Bernico, X.~Dong, C.~Hidey, K.~Kilgour, G.~Liu, S.~Bhupatiraju, L.~Leonhard, S.~Zuo, P.~Talukdar, Q.~Wei, A.~Severyn, V.~Listík, J.~Lee, A.~Tripathi, S.~Park, Y.~Matias, H.~Liu, A.~Ruiz, R.~Jayaram, J.~Tolins, P.~Marcenac, Y.~Wang, B.~Seybold, H.~Prior, D.~Sharma, J.~Weber, M.~Sirotenko, Y.~Sung, D.~Du, E.~Pavlick, S.~Zinke, M.~Freitag, M.~Dylla, M.~G. Arenas, N.~Potikha, O.~Goldman, C.~Tao, R.~Chhaparia, M.~Voitovich, P.~Dogra, A.~Ražnatović, Z.~Tsai, C.~You, O.~Johnson, G.~Tucker, C.~Gu, J.~Yoo, M.~Majzoubi, V.~Gabeur, B.~Raad, R.~Rhodes, K.~Kolipaka, H.~Howard, G.~Sampemane, B.~Li,
  C.~Asawaroengchai, D.~Nguyen, C.~Zhang, T.~Cour, X.~Yu, Z.~Fu, J.~Jiang, P.-S. Huang, G.~Surita, I.~Iturrate, Y.~Karov, M.~Collins, M.~Baeuml, F.~Fuchs, S.~Shetty, S.~Ramaswamy, S.~Ebrahimi, Q.~Guo, J.~Shar, G.~Barth-Maron, S.~Addepalli, B.~Richter, C.-Y. Cheng, E.~Rives, F.~Zheng, J.~Griesser, N.~Dikkala, Y.~Zeldes, I.~Safarli, D.~Das, H.~Srivastava, S.~M. Khan, X.~Li, A.~Pandey, L.~Markeeva, D.~Belov, Q.~Yan, M.~Rybiński, T.~Chen, M.~Nawhal, M.~Quinn, V.~Govindaraj, S.~York, R.~Roberts, R.~Garg, N.~Godbole, J.~Abernethy, A.~Das, L.~N. Thiet, J.~Tompson, J.~Nham, N.~Vats, B.~Caine, W.~Helmholz, F.~Pongetti, Y.~Ko, J.~An, C.~H. Hu, Y.-C. Ling, J.~Pawar, R.~Leland, K.~Kinoshita, W.~Khawaja, M.~Selvi, E.~Ie, D.~Sinopalnikov, L.~Proleev, N.~Tripuraneni, M.~Bevilacqua, S.~Lee, C.~Sanford, D.~Suh, D.~Tran, J.~Dean, S.~Baumgartner, J.~Heitkaemper, S.~Gubbi, K.~Toutanova, Y.~Xu, C.~Thekkath, K.~Rong, P.~Jain, A.~Xie, Y.~Virin, Y.~Li, L.~Litchev, R.~Powell, T.~Bharti, A.~Kraft, N.~Hua, M.~Ikonomidis, A.~Hitron,
  S.~Kumar, L.~Matthey, S.~Bridgers, L.~Lax, I.~Malhi, O.~Skopek, A.~Gupta, J.~Cao, M.~Rasquinha, S.~Põder, W.~Stokowiec, N.~Roth, G.~Li, M.~Sander, J.~Kessinger, V.~Jain, E.~Loper, W.~Park, M.~Yarom, L.~Cheng, G.~Guruganesh, K.~Rao, Y.~Li, C.~Barros, M.~Sushkov, C.-S. Ferng, R.~Shah, O.~Aharoni, R.~Kumar, T.~McConnell, P.~Li, C.~Wang, F.~Pereira, C.~Swanson, F.~Jamil, Y.~Xiong, A.~Vijayakumar, P.~Shroff, K.~Soparkar, J.~Gu, L.~B. Soares, E.~Wang, K.~Majmundar, A.~Wei, K.~Bailey, N.~Kassner, C.~Kawamoto, G.~Žužić, V.~Gomes, A.~Gupta, M.~Guzman, I.~Dasgupta, X.~Bai, Z.~Pan, F.~Piccinno, H.~N. Vogel, O.~Ponce, A.~Hutter, P.~Chang, P.-P. Jiang, I.~Gog, V.~Ionescu, J.~Manyika, F.~Pedregosa, H.~Ragan, Z.~Behrman, R.~Mullins, C.~Devin, A.~Pyne, S.~Gawde, M.~Chadwick, Y.~Gu, S.~Tavakkol, A.~Twigg, N.~Goyal, N.~Elue, A.~Goldie, S.~Venkatachary, H.~Fei, Z.~Feng, M.~Ritter, I.~Leal, S.~Dasari, P.~Sun, A.~R. Rochman, B.~O'Donoghue, Y.~Liu, J.~Sproch, K.~Chen, N.~Clay, S.~Petrov, S.~Sidhwani, I.~Mihailescu,
  A.~Panagopoulos, A.~Piergiovanni, Y.~Bai, G.~Powell, D.~Karkhanis, T.~Yacovone, P.~Mitrichev, J.~Kovac, D.~Uthus, A.~Yazdanbakhsh, D.~Amos, S.~Zheng, B.~Zhang, J.~Miao, B.~Ramabhadran, S.~Radpour, S.~Thakoor, J.~Newlan, O.~Lang, O.~Jankowski, S.~Bharadwaj, J.-M. Sarr, S.~Ashraf, S.~Mondal, J.~Yan, A.~S. Rawat, S.~Velury, G.~Kochanski, T.~Eccles, F.~Och, A.~Sharma, E.~Mahintorabi, A.~Gurney, C.~Muir, V.~Cohen, S.~Thakur, A.~Bloniarz, A.~Mujika, A.~Pritzel, P.~Caron, A.~Rahman, F.~Lang, Y.~Onoe, P.~Sirkovic, J.~Hoover, Y.~Jian, P.~Duque, A.~Narayanan, D.~Soergel, A.~Haig, L.~Maggiore, S.~Buch, J.~Dean, I.~Figotin, I.~Karpov, S.~Gupta, D.~Zhou, M.~Huang, A.~Vaswani, C.~Semturs, K.~Shivakumar, Y.~Watanabe, V.~K. Rajendran, E.~Lu, Y.~Hou, W.~Ye, S.~Vashishth, N.~Nti, V.~Sakenas, D.~Ni, D.~DeCarlo, M.~Bendersky, S.~Bagri, N.~Cano, E.~Peake, S.~Tokumine, V.~Godbole, C.~Guía, T.~Lando, V.~Selo, S.~Ellis, D.~Tarlow, D.~Gillick, A.~Epasto, S.~R. Jonnalagadda, M.~Wei, M.~Xie, A.~Taly, M.~Paganini, M.~Sundararajan,
  D.~Toyama, T.~Yu, D.~Petrova, A.~Pappu, R.~Agrawal, S.~Buthpitiya, J.~Frye, T.~Buschmann, R.~Crocker, M.~Tagliasacchi, M.~Wang, D.~Huang, S.~Perel, B.~Wieder, H.~Kazawa, W.~Wang, J.~Cole, H.~Gupta, B.~Golan, S.~Bang, N.~Kulkarni, K.~Franko, C.~Liu, D.~Reid, S.~Dalmia, J.~Whang, K.~Cen, P.~Sundaram, J.~Ferret, B.~Isik, L.~Ionita, G.~Sun, A.~Shekhawat, M.~Mohammad, P.~Pham, R.~Huang, K.~Raman, X.~Zhou, R.~Mcilroy, A.~Myers, S.~Peng, J.~Scott, P.~Covington, S.~Erell, P.~Joshi, J.~G. Oliveira, N.~Noy, T.~Nasir, J.~Walker, V.~Axelrod, T.~Dozat, P.~Han, C.-T. Chu, E.~Weinstein, A.~Shukla, S.~Chandrakaladharan, P.~Poklukar, B.~Li, Y.~Jin, P.~Eruvbetine, S.~Hansen, A.~Dabush, A.~Jacovi, S.~Phatale, C.~Zhu, S.~Baker, M.~Shomrat, Y.~Xiao, J.~Pouget-Abadie, M.~Zhang, F.~Wei, Y.~Song, H.~King, Y.~Huang, Y.~Zhu, R.~Sun, J.~V. Franco, C.-C. Lin, S.~Arora, Hui, Li, V.~Xia, L.~Vilnis, M.~Schain, K.~Alarakyia, L.~Prince, A.~Phillips, C.~Habtegebriel, L.~Xu, H.~Gui, S.~Ontanon, L.~Aroyo, K.~Gill, P.~Lu, Y.~Katariya,
  D.~Madeka, S.~Krishnan, S.~S. Raghvendra, J.~Freedman, Y.~Tay, G.~Menghani, P.~Choy, N.~Shetty, D.~Abolafia, D.~Kukliansky, E.~Chou, J.~Lichtarge, K.~Burke, B.~Coleman, D.~Guo, L.~Jin, I.~Bhattacharya, V.~Langston, Y.~Li, S.~Kotecha, A.~Yakubovich, X.~Chen, P.~Petrov, T.~Powell, Y.~He, C.~Quick, K.~Garg, D.~Hwang, Y.~Lu, S.~Bhojanapalli, K.~Kjems, R.~Mehran, A.~Archer, H.~van Hasselt, A.~Balakrishna, J.~Kearns, M.~Guo, J.~Riesa, M.~Sazanovich, X.~Gao, C.~Sauer, C.~Yang, X.~Sheng, T.~Jimma, W.~V. Gansbeke, V.~Nikolaev, W.~Wei, K.~Millican, R.~Zhao, J.~Snyder, L.~Bolelli, M.~O'Brien, S.~Xu, F.~Xia, W.~Yuan, A.~Neelakantan, D.~Barker, S.~Yadav, H.~Kirkwood, F.~Ahmad, J.~Wee, J.~Grimstad, B.~Wang, M.~Wiethoff, S.~Settle, M.~Wang, C.~Blundell, J.~Chen, C.~Duvarney, G.~Hu, O.~Ronneberger, A.~Lee, Y.~Li, A.~Chakladar, A.~Butryna, G.~Evangelopoulos, G.~Desjardins, J.~Kanerva, H.~Wang, A.~Nowak, N.~Li, A.~Loo, A.~Khurshudov, L.~E. Shafey, N.~Baddi, K.~Lenc, Y.~Razeghi, T.~Lieber, A.~Sinha, X.~Ma, Y.~Su, J.~Huang,
  A.~Ushio, H.~Klimczak-Plucińska, K.~Mohamed, J.~Chen, S.~Osindero, S.~Ginzburg, L.~Lamprou, V.~Bashlovkina, D.-H. Tran, A.~Khodaei, A.~Anand, Y.~Di, R.~Eskander, M.~R. Vuyyuru, J.~Liu, A.~Kamath, R.~Goldenberg, M.~Bellaiche, J.~Pluto, B.~Rosgen, H.~Mansoor, W.~Wong, S.~Ganesh, E.~Bailey, S.~Baird, D.~Deutsch, J.~Baek, X.~Jia, C.~Lee, A.~Friesen, N.~Braun, K.~Lee, A.~Panda, S.~M. Hernandez, D.~Williams, J.~Liu, E.~Liang, A.~Autef, E.~Pitler, D.~Jain, P.~Kirk, O.~Bunyan, J.~S. Elias, T.~Yin, M.~Reid, A.~Pope, N.~Putikhin, B.~Samanta, S.~Guadarrama, D.~Kim, S.~Rowe, M.~Valentine, G.~Yan, A.~Salcianu, D.~Silver, G.~Song, R.~Singh, S.~Ye, H.~DeBalsi, M.~A. Merey, E.~Ofek, A.~Webson, S.~Mourad, A.~Kakarla, S.~Lattanzi, N.~Roy, E.~Sluzhaev, C.~Butterfield, A.~Tonioni, N.~Waters, S.~Kopalle, J.~Chase, J.~Cohan, G.~R. Rao, R.~Berry, M.~Voznesensky, S.~Hu, K.~Chiafullo, S.~Chikkerur, G.~Scrivener, I.~Zheng, J.~Wiesner, W.~Macherey, T.~Lillicrap, F.~Liu, B.~Walker, D.~Welling, E.~Davies, Y.~Huang, L.~Ren, N.~Shabat,
  A.~Agostini, M.~Iinuma, D.~Zelle, R.~Sathyanarayana, A.~D'olimpio, M.~Redshaw, M.~Ginsberg, A.~Murthy, M.~Geller, T.~Matejovicova, A.~Chakrabarti, R.~Julian, C.~Chan, Q.~Hu, D.~Jarrett, M.~Agarwal, J.~Challagundla, T.~Li, S.~Tata, W.~Ding, M.~Meng, Z.~Dai, G.~Vezzani, S.~Garg, J.~Bulian, M.~Jasarevic, H.~Cai, H.~Rajamani, A.~Santoro, F.~Hartmann, C.~Liang, B.~Perz, A.~Jindal, F.~Bu, S.~Seo, R.~Poplin, A.~Goedeckemeyer, B.~Ghazi, N.~Khadke, L.~Liu, K.~Mather, M.~Zhang, A.~Shah, A.~Chen, J.~Wei, K.~Shivam, Y.~Cao, D.~Cho, A.~S. Scarpati, M.~Moffitt, C.~Barbu, I.~Jurin, M.-W. Chang, H.~Liu, H.~Zheng, S.~Dave, C.~Kaeser-Chen, X.~Yu, A.~Abdagic, L.~Gonzalez, Y.~Huang, P.~Zhong, C.~Schmid, B.~Petrini, A.~Wertheim, J.~Zhu, H.~Nguyen, K.~Ji, Y.~Zhou, T.~Zhou, F.~Feng, R.~Cohen, D.~Rim, S.~M. Phal, P.~Georgiev, A.~Brand, Y.~Ma, W.~Li, S.~Gupta, C.~Wang, P.~Dubov, J.~Tarbouriech, K.~Majumder, H.~Li, N.~Rink, A.~Suman, Y.~Guo, Y.~Sun, A.~Nair, X.~Xu, M.~Elhawaty, R.~Cabrera, G.~Han, J.~Eisenschlos, J.~Bai, Y.~Li,
  Y.~Bansal, T.~Sellam, M.~Khan, H.~Nguyen, J.~Mao-Jones, N.~Parotsidis, J.~Marcus, C.~Fan, R.~Zimmermann, Y.~Kochinski, L.~Graesser, F.~Behbahani, A.~Caceres, M.~Riley, P.~Kane, S.~Lefdal, R.~Willoughby, P.~Vicol, L.~Wang, S.~Zhang, A.~Gill, Y.~Liang, G.~Prasad, S.~Mariooryad, M.~Kazemi, Z.~Wang, K.~Muralidharan, P.~Voigtlaender, J.~Zhao, H.~Zhou, N.~D'Souza, A.~Mavalankar, S.~Arnold, N.~Young, O.~Sarvana, C.~Lee, M.~Nasr, T.~Zou, S.~Kim, L.~Haas, K.~Patel, N.~Bulut, D.~Parkinson, C.~Biles, D.~Kalashnikov, C.~M. To, A.~Kumar, J.~Austin, A.~Greve, L.~Zhang, M.~Goel, Y.~Li, S.~Yaroshenko, M.~Chang, A.~Jindal, G.~Clark, H.~Taitelbaum, D.~Johnson, O.~Roval, J.~Ko, A.~Mohananey, C.~Schuler, S.~Dodhia, R.~Li, K.~Osawa, C.~Cui, P.~Xu, R.~Shah, T.~Huang, E.~Gruzewska, N.~Clement, M.~Verma, O.~Sercinoglu, H.~Qian, V.~Shah, M.~Yamaguchi, A.~Modi, T.~Kosakai, T.~Strohmann, J.~Zeng, B.~Gunel, J.~Qian, A.~Tarango, K.~Jastrzebski, R.~David, J.~Shan, P.~Schuh, K.~Lad, W.~Gierke, M.~Madhavan, X.~Chen, M.~Kurzeja,
  R.~Santamaria-Fernandez, D.~Chen, A.~Cordell, Y.~Chervonyi, F.~Garcia, N.~Kannen, V.~Perot, N.~Ding, S.~Cohen-Ganor, V.~Lavrenko, J.~Wu, G.~Evans, C.~N. dos Santos, M.~Sewak, A.~Brown, A.~Hard, J.~Puigcerver, Z.~Zheng, Y.~Liang, E.~Gladchenko, R.~Ingle, U.~First, P.~Sermanet, C.~Magister, M.~Velimirović, S.~Reddi, S.~Ricco, E.~Agustsson, H.~Adam, N.~Levine, D.~Gaddy, D.~Holtmann-Rice, X.~Wang, A.~Sathe, A.~G. Roy, B.~Bratanič, A.~Carin, H.~Mehta, S.~Bonacina, N.~D. Cao, M.~Finkelstein, V.~Rieser, X.~Wu, F.~Altché, D.~Scandinaro, L.~Li, N.~Vieillard, N.~Sethi, G.~Tanzer, Z.~Xing, S.~Wang, P.~Bhatia, G.~Citovsky, T.~Anthony, S.~Lin, T.~Shi, S.~Jakobovits, G.~Gibson, R.~Apte, L.~Lee, M.~Chen, A.~Byravan, P.~Maniatis, K.~Webster, A.~Dai, P.-C. Chen, J.~Pan, A.~Fadeeva, Z.~Gleicher, T.~Luong, and N.~K. Bhumihar, ``Gemini 2.5: Pushing the frontier with advanced reasoning, multimodality, long context, and next generation agentic capabilities,'' 2025.

\bibitem{GLM4_5}
{Zai}, ``Glm-4.5: Reasoning, coding, and agentic abililties.'' \url{https://z.ai/blog/glm-4.5}, July 2025.
\newblock Accessed: 2025-07-28.

\bibitem{KimiK2}
{Moonshot}, ``Kimi k2: Open agentic intelligence.'' \url{https://moonshotai.github.io/Kimi-K2/}, July 2025.
\newblock Accessed: 2025-07-28.

\bibitem{Qwen3}
A.~Yang, A.~Li, B.~Yang, B.~Zhang, B.~Hui, B.~Zheng, B.~Yu, C.~Gao, C.~Huang, C.~Lv, C.~Zheng, D.~Liu, F.~Zhou, F.~Huang, F.~Hu, H.~Ge, H.~Wei, H.~Lin, J.~Tang, J.~Yang, J.~Tu, J.~Zhang, J.~Yang, J.~Yang, J.~Zhou, J.~Zhou, J.~Lin, K.~Dang, K.~Bao, K.~Yang, L.~Yu, L.~Deng, M.~Li, M.~Xue, M.~Li, P.~Zhang, P.~Wang, Q.~Zhu, R.~Men, R.~Gao, S.~Liu, S.~Luo, T.~Li, T.~Tang, W.~Yin, X.~Ren, X.~Wang, X.~Zhang, X.~Ren, Y.~Fan, Y.~Su, Y.~Zhang, Y.~Zhang, Y.~Wan, Y.~Liu, Z.~Wang, Z.~Cui, Z.~Zhang, Z.~Zhou, and Z.~Qiu, ``Qwen3 technical report,'' 2025.

\bibitem{DeepSeek-V3}
DeepSeek-AI, A.~Liu, B.~Feng, B.~Xue, B.~Wang, B.~Wu, C.~Lu, C.~Zhao, C.~Deng, C.~Zhang, C.~Ruan, D.~Dai, D.~Guo, D.~Yang, D.~Chen, D.~Ji, E.~Li, F.~Lin, F.~Dai, F.~Luo, G.~Hao, G.~Chen, G.~Li, H.~Zhang, H.~Bao, H.~Xu, H.~Wang, H.~Zhang, H.~Ding, H.~Xin, H.~Gao, H.~Li, H.~Qu, J.~L. Cai, J.~Liang, J.~Guo, J.~Ni, J.~Li, J.~Wang, J.~Chen, J.~Chen, J.~Yuan, J.~Qiu, J.~Li, J.~Song, K.~Dong, K.~Hu, K.~Gao, K.~Guan, K.~Huang, K.~Yu, L.~Wang, L.~Zhang, L.~Xu, L.~Xia, L.~Zhao, L.~Wang, L.~Zhang, M.~Li, M.~Wang, M.~Zhang, M.~Zhang, M.~Tang, M.~Li, N.~Tian, P.~Huang, P.~Wang, P.~Zhang, Q.~Wang, Q.~Zhu, Q.~Chen, Q.~Du, R.~J. Chen, R.~L. Jin, R.~Ge, R.~Zhang, R.~Pan, R.~Wang, R.~Xu, R.~Zhang, R.~Chen, S.~S. Li, S.~Lu, S.~Zhou, S.~Chen, S.~Wu, S.~Ye, S.~Ye, S.~Ma, S.~Wang, S.~Zhou, S.~Yu, S.~Zhou, S.~Pan, T.~Wang, T.~Yun, T.~Pei, T.~Sun, W.~L. Xiao, W.~Zeng, W.~Zhao, W.~An, W.~Liu, W.~Liang, W.~Gao, W.~Yu, W.~Zhang, X.~Q. Li, X.~Jin, X.~Wang, X.~Bi, X.~Liu, X.~Wang, X.~Shen, X.~Chen, X.~Zhang, X.~Chen, X.~Nie, X.~Sun,
  X.~Wang, X.~Cheng, X.~Liu, X.~Xie, X.~Liu, X.~Yu, X.~Song, X.~Shan, X.~Zhou, X.~Yang, X.~Li, X.~Su, X.~Lin, Y.~K. Li, Y.~Q. Wang, Y.~X. Wei, Y.~X. Zhu, Y.~Zhang, Y.~Xu, Y.~Xu, Y.~Huang, Y.~Li, Y.~Zhao, Y.~Sun, Y.~Li, Y.~Wang, Y.~Yu, Y.~Zheng, Y.~Zhang, Y.~Shi, Y.~Xiong, Y.~He, Y.~Tang, Y.~Piao, Y.~Wang, Y.~Tan, Y.~Ma, Y.~Liu, Y.~Guo, Y.~Wu, Y.~Ou, Y.~Zhu, Y.~Wang, Y.~Gong, Y.~Zou, Y.~He, Y.~Zha, Y.~Xiong, Y.~Ma, Y.~Yan, Y.~Luo, Y.~You, Y.~Liu, Y.~Zhou, Z.~F. Wu, Z.~Z. Ren, Z.~Ren, Z.~Sha, Z.~Fu, Z.~Xu, Z.~Huang, Z.~Zhang, Z.~Xie, Z.~Zhang, Z.~Hao, Z.~Gou, Z.~Ma, Z.~Yan, Z.~Shao, Z.~Xu, Z.~Wu, Z.~Zhang, Z.~Li, Z.~Gu, Z.~Zhu, Z.~Liu, Z.~Li, Z.~Xie, Z.~Song, Z.~Gao, and Z.~Pan, ``Deepseek-v3 technical report,'' 2025.

\bibitem{ChatboArena}
W.~Chiang, L.~Zheng, Y.~Sheng, A.~N. Angelopoulos, T.~Li, D.~Li, B.~Zhu, H.~Zhang, M.~I. Jordan, J.~E. Gonzalez, and I.~Stoica, ``Chatbot arena: An open platform for evaluating llms by human preference,'' in {\em Forty-first International Conference on Machine Learning, {ICML} 2024, Vienna, Austria, July 21-27, 2024}, OpenReview.net, 2024.

\end{thebibliography}
